\documentclass[sigconf]{aamas} 

\usepackage{balance} % for balancing columns on the final page
\usepackage{graphicx}           % For including images
\usepackage[space]{grffile}     % For spaces in image names
\usepackage{url}                % For displaying urls
\usepackage{hyperref}
\usepackage{subfigure}
\usepackage{graphbox}
\usepackage[capitalize,noabbrev]{cleveref}

\usepackage{algorithm}
\usepackage{algorithmic}
\usepackage{booktabs}
\usepackage{multirow}
\usepackage{makecell}
\usepackage{hhline}
\usepackage{wrapfig}
\usepackage{xcolor}
\usepackage{amsmath}
\usepackage{mathtools}
\usepackage{amsthm}
\usepackage{bbm}
\usepackage{array}
\usepackage{arydshln}
\usepackage{enumitem}

\makeatletter
\gdef\@copyrightpermission{
  \begin{minipage}{0.2\columnwidth}
   \href{https://creativecommons.org/licenses/by/4.0/}{\includegraphics[width=0.90\textwidth]{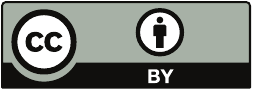}}
  \end{minipage}\hfill
  \begin{minipage}{0.8\columnwidth}
   \href{https://creativecommons.org/licenses/by/4.0/}{This work is licensed under a Creative Commons Attribution International 4.0 License.}
  \end{minipage}
  \vspace{5pt}
}
\makeatother

\setcopyright{ifaamas}
\acmConference[AAMAS '25]{Proc.\@ of the 24th International Conference
on Autonomous Agents and Multiagent Systems (AAMAS 2025)}{May 19 -- 23, 2025}
{Detroit, Michigan, USA}{Y.~Vorobeychik, S.~Das, A.~Nowé  (eds.)}
\copyrightyear{2025}
\acmYear{2025}
\acmDOI{}
\acmPrice{}
\acmISBN{}

\acmSubmissionID{<<OpenReview submission id>>}

\title{\texorpdfstring{$\beta$}{beta}-DQN: Improving Deep Q-Learning By Evolving the Behavior}

\author{Hongming Zhang}
\affiliation{
  \institution{University of Alberta and Amii}
  \city{Edmonton}
  \country{Canada}}
\email{hongmin2@ualberta.ca}

\author{Fengshuo Bai}
\affiliation{
  \institution{Shanghai Jiao Tong University}
  \institution{Zhongguancun Academy}
  \city{Shanghai}
  \country{China}}
\email{fengshuobai@sjtu.edu.cn}

\author{Chenjun Xiao}
\affiliation{
  \institution{CUHK-Shenzhen}
  \city{Shenzhen}
  \country{China}}
\email{chenjunx@cuhk.edu.cn}

\author{Chao Gao}
\affiliation{
  \institution{Edmonton Research Center, Huawei}
  \city{Edmonton}
  \country{Canada}}
\email{chao.gao4@huawei.com}

\author{Bo Xu}
\affiliation{
  \institution{CASIA}
  \city{Beijing}
  \country{China}}
\email{boxu@ia.ac.cn}

\author{Martin M{\"u}ller}
\affiliation{
  \institution{University of Alberta and Amii}
  \city{Edmonton}
  \country{Canada}}
\email{mmueller@ualberta.ca}

\begin{abstract} 

While many sophisticated exploration methods have been proposed, their lack of generality and high computational cost often lead researchers to favor simpler methods like $\epsilon$-greedy. Motivated by this, we introduce $\beta$-DQN, a simple and efficient exploration method that augments the standard DQN with a behavior function $\beta$. This function estimates the probability that each action has been taken at each state. By leveraging $\beta$, we generate a population of diverse policies that balance exploration between state-action coverage and overestimation bias correction. An adaptive meta-controller is designed to select an effective policy for each episode, enabling flexible and explainable exploration. $\beta$-DQN is straightforward to implement and adds minimal computational overhead to the standard DQN. Experiments on both simple and challenging exploration domains show that $\beta$-DQN outperforms existing baseline methods across a wide range of tasks, providing an effective solution for improving exploration in deep reinforcement learning.
\end{abstract}


\newcommand{\BibTeX}{\rm B\kern-.05em{\sc i\kern-.025em b}\kern-.08em\TeX}

\begin{document}

%%% The following commands remove the headers in your paper. For final 
%%% papers, these will be inserted during the pagination process.

\pagestyle{fancy}
\fancyhead{}

%%% The next command prints the information defined in the preamble.

\maketitle 

%%%%%%%%%%%%%%%%%%%%%%%%%%%%%%%%%%%%%%%%%%%%%%%%%%%%%%%%%%%%%%%%%%%%%%%%

\section{Introduction}
Exploration is considered as a major challenge in deep reinforcement learning (DRL)~\citep{sutton2018reinforcement,yang2021exploration}. The agent needs to trade off between exploiting current knowledge for known rewards and exploring the environment for potential better rewards. 
While many complex methods have been proposed for efficient exploration, the most commonly used ones are still simple methods such as $\epsilon$-greedy and entropy regularization \citep{mnih2015human,schulman2017proximal,Bai_Zhang_Tao_Wu_Wang_Xu_2023,zhang2020alphazero}. 
We identify the possible reasons. First, more advanced methods require meticulous hyper-parameter tuning and much computational cost \citep{badia2020agent57,Badia2020Never,fan2023learnable}.
Second, these methods adopt specialized inductive biases, which may achieve high performance in specific hard exploration environments but tend to underperform compared to simpler methods across a broad range of domains, highlighting their lack of generality~\citep{burda2018exploration,Taiga2020On,bai2024rat}.

We improve exploration while considering the following aspects:
(1) \textbf{Simplicity}. We aim to achieve clear improvement while keeping the method simple. This ensure the method is straightforward to implement and minimizes the burden of hyper-parameters tuning.
(2) \textbf{Mild Increase in Computational Cost}. While prioritizing sample efficiency in RL tasks, we aim to strike a balance that avoids substantial increase in training time. Our goal is to develop a method that is both effective and efficient.
(3) \textbf{Generality Across Tasks}. The method should maintain generality, 
% and be applicable to a wide range of tasks, 
rather than being tailored to specific hard exploration environments.

Motivated by these considerations, we propose $\beta$-DQN, a simple and efficient exploration method that augments the standard DQN with a behavior function $\beta$. The function $\beta$ represents the behavior policy that collects data in the replay memory,
% \footnote{The behavior policy could be one policy, or an average of multiple policies.}, 
estimating the probability that each action has been taken at each state. Combined with the $Q$ function in DQN, we use $\beta$ for three purposes:
(1) \textbf{Exploration for state-action coverage}. Taking actions with low probabilities based on $\beta$ encourages the agent to explore the state-action space for better coverage;
(2) \textbf{Exploration for overestimation bias correction}. Exploring overestimated actions to get feedback and correct the overestimation bias in the $Q$ function;
(3) \textbf{Pure exploitation}. Using $\beta$ to mask the $Q$ function at unseen actions derives a greedy policy that represents pure exploitation. Interpolating among them allows us to construct a population of temporally-extended policies that interleave exploration and exploitation at intra-episodic level with clear purposes~\citep{pislar2022when}. We then design a meta-controller to adaptively select an effective policy for each episode, providing flexibility without an accompanying hyperparameter-tuning burden.

% Taking actions with low probabilities encourages the agent to explore the state space for better \emph{coverage}. Consider the $Q$ function learned by DQN may overestimate the action values due to the maximum operator in the Bellman update, taking actions with the highest $Q$ values is a natural choice to try these actions and \emph{correct the overestimation bias}~\citep{simmons2021reward,schaul2022the}. 
% Finally, if we reduce $Q$ values at unseen actions determined by behavior function $\beta$, denoted as $Q_{\textit{mask}}$, we can derive a greedy policy that represents \emph{pure exploitation} based on current experience in the memory.
% In addition, we use $\beta$ to constrain TD learning to bootstrap from in-sample state-action pairs as shown in \cref{eq:in distribution td learning}. The $Q$ function learns in-sample estimation from the replay memory and generalizes at missing data. 
% This share the same purpose as offline RL that tries to maximize the cumulative rewards limited to a dataset~\citep{lange2012batch,levine2020offline}, which leads to a conservative policy with best possible performance.
% With the set of diverse policies, we consider the policy selection as a non-stationary multi-armed bandit problem (MAB)~\citep{garivier2008upper}. 

Our method have the following advantages:
(1) We only additionally learn a behavior function, which is straightforward to implement and computationally efficient compared to previous methods~\citep{badia2020agent57,kim2023lesson}. 
(2) When constructing diverse polices, we do not inject inductive biases specialized for one specific task, making the method general and applicable across a wide range of domains. 
(3) Our method interleaves exploitation and exploration at the intra-episodic level, carries out temporal-extended exploration, and is state-dependent, which is considered the most effective approach~\citep{osband2016deep,pislar2022when,dabney2021temporallyextended}.
We report promising results on dense-reward MinAtar~\citep{young19minatar} and sparse-reward MiniGrid~\citep{MinigridMiniworld23}, demonstrating that our method significantly enhances performance and exhibits broad applicability in both easy and hard exploration domains.

\begin{figure*}[t]
\begin{center}
    \vspace{-0.15in}
    \subfigure{\Description{}\includegraphics[width=0.95\textwidth,height=0.3\textwidth]{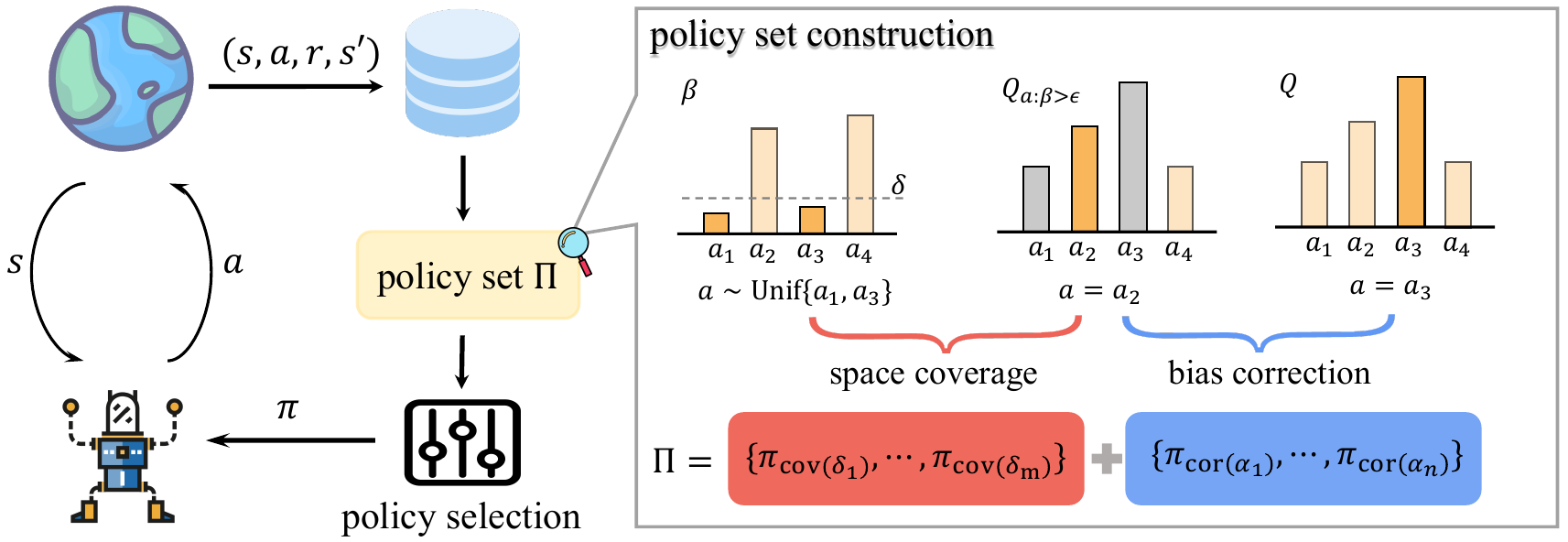}}
    \vskip -0.15in
    \caption{Method overview. We learn a behavior function $\beta$ from the replay memory and augment it with the $Q$ function for three purposes: state-action coverage (left), pure exploitation (middle), and overestimation correction (right). By interpolating between these strategies, we construct a policy set that balances exploration and exploitation at the intra-episodic level. A meta-controller adaptively selects a policy for each episode.}
    % \vskip -0.15in
    \label{fig:overview}
\end{center}
\end{figure*}

\section{Related Work}
\label{sec:related work}
Reinforcement learning (RL) is known for learning through trial and error~\citep{sutton2018reinforcement,dong2020deep}. If a state-action pair has not been encountered, it cannot be learned~\citep{pislar2022when}, making exploration a central challenge in RL.
The most commonly used exploration strategies are simple dithering methods like $\epsilon$-greedy and entropy regularization~\citep{zhang2020taxonomy,mnih2015human,schulman2017proximal,haarnoja2018soft,zhang2024provable}.
While these methods are general, they are often inefficient because they are state-independent and lack temporal persistence~\citep{pislar2022when,yang2021exploration}. Therefore, inducing a consistent, state-dependent exploration policy over multiple time steps has been a key pursuit in the field~\citep{osband2016deep,sekar2020planning,ecoffet2021first,dabney2021temporallyextended,simmons2021reward,pislar2022when,ji2023seizing}.

\textbf{Temporally-Extended Exploration.}
Bootstrapped DQN~\citep{osband2016deep} induces temporally-extended exploration by building $K$ bootstrapped estimates of the Q-value function in parallel and sampling a single $Q$ function for the duration of one episode. The computational cost increases linearly with the number of heads $K$. Temporally-extended $\epsilon$-Greedy ($\epsilon$z-greedy)~\citep{dabney2021temporallyextended} simply repeats the sampled random action for a random duration, but its exploration remains state-independent. Our method, based on the behavior function $\beta$, induces temporally-extended exploration by selecting actions according to state-dependent probabilities.

Besides adding temporal persistence to the exploration policy explicitly, another line of work involves adding exploration bonuses to the environment reward~\citep{bellemare2016unifying,tang2017exploration,ostrovski2017count,lobel2023flipping,sun2022optimistic,guo2022byol,jarrett2022curiosity,zhang2020bebold,NEURIPS2022_8be9c134}. These bonuses are designed to encourage the agent to visit states that are novel. The count-based exploration bonus encourages the agent to visit states with low visit counts~\citep{bellemare2016unifying,tang2017exploration,ostrovski2017count}. Prediction error-based methods, such as the Intrinsic Curiosity Module (ICM)~\citep{pathak2017curiosity} and Random Network Distillation (RND)~\citep{burda2018exploration,Badia2020Never}, operate on the intuition that the prediction error will be low on states that are previously visited and high on newly visited states. These methods are designed to tackle difficult exploration problems and usually perform well in hard exploration environments. However, they often underperform compared to simple methods like $\epsilon$-greedy in easy exploration environments~\citep{Taiga2020On}, highlighting their lack of generality. In contrast, our method is designed to be general and applicable across a wide range of tasks.

% Count-based bonuses encourage agents to visit states with low visit counts, and various methods have been proposed to estimate counts in high-dimensional states spaces~\citep{bellemare2016unifying,tang2017exploration,ostrovski2017count}. 

\textbf{Population-based Exploration.}
Recent promising works tried to handle the exploration problem using population-based
methods, which collect samples with diverse behaviors derived from a population of different exploratory policies~\citep{badia2020agent57,fan2022generalized,kapturowski2023humanlevel,fan2023learnable,kim2023lesson,sun2020novel,anschel2017averaged}. These methods have demonstrated powerful performance, outperforming the standard human benchmark on all 57 Atari games~\citep{bellemare2013arcade}.
They maintain a group of diverse actors with independent parameters, build distributed systems, and interact with the environment over billions of frames. While these methods achieve significant performance gains, their computational cost is substantial and often unaffordable in practice. This widens the gap between researchers with ample access to computational resources and those without~\citep{ceron2021revisiting}. $\beta$-DQN introduces a population of diverse policies with minimal computational overhead, making it much more accessible. Our goal is to absorb the strengths of existing population-based methods and design an effective approach with a low computational cost.

\section{Background}
\label{sec:background}

\textbf{Markov Decision Process (MDP).} Reinforcement learning (RL)~\citep{sutton2018reinforcement} is a paradigm of agent learning via interaction. It can be modeled as a Markov Decision Process~(MDP) $\mathcal{M}=(\mathcal{S}, \mathcal{A}, R, P,\rho_0, \gamma)$. $\mathcal{S}$ is the state space, $\mathcal{A}$ is the action space, 
$P:\mathcal{S} \times \mathcal{A} \times \mathcal{S} \rightarrow [0, 1]$ is the environment transition dynamics, $R: \mathcal{S} \times \mathcal{A} \times \mathcal{S} \rightarrow \mathbb{R}$ is the reward function, $\rho_0: \mathcal{S}\rightarrow \mathbb{R}$ is the initial state distribution and $\gamma \in (0,1)$ is the discount factor.
The goal of the agent is to learn an optimal policy that maximizes the expected discounted cumulative rewards $\mathbb{E}[\sum_{t=0}^{\infty}\gamma^t r_t]$.

\textbf{Deep Q-Network (DQN).}
Q-learning is a classic algorithm to learn the optimal policy. It learns the $Q$ function with Bellman optimality equation~\citep{Bellman+2021}, $Q^*(s,a)=\mathbb{E}[r+\gamma\max_{a^\prime}Q^*(s^\prime,a^\prime)]$. An optimal policy is then derived by taking an action with maximum $Q$ value at each state.
DQN~\citep{mnih2015human} scales up Q-learning by using deep neural networks and experience replay~\citep{lin1992self}.
It stores transitions in a replay memory and samples batches of data uniformly to estimate an action-value function $Q_\theta$ with temporal-difference (TD) learning:
\begin{equation}
Q_\theta(s,a) \leftarrow r(s,a)+\gamma\max_{a^\prime}Q_\theta(s^\prime,a^\prime). 
\label{eq:bellman}
\end{equation}
A target network with parameters $\theta^{\bar ~}$ copies the parameters from $\theta$ only every $C$ steps to stabilize the computation of learning target $y=r(s,a)+\gamma \max_{a^\prime}Q_{\theta^{\bar ~}}(s^\prime,a^\prime)$.

\section{Method}
Drawing from insights introduced in \cref{sec:related work}, promising exploration strategies should be state-dependent, temporally-extended, and consist of a set of diverse policies. Keeping simplicity and generality in mind, we design an exploration method for DQN that performs well across a wide range of domains and is computationally affordable for the research community. We additionally learn a behavior function $\beta$ and construct a set of policies that balance exploration between state coverage and bias correction. A meta-controller is then designed to adaptively select a policy for each episode.
In~\cref{sec:basic polices}, we introduce how to learn the behavior function $\beta$ and augment it with the $Q$ function for three purposes: exploration for state-action coverage, exploration for overestimation bias correction, and pure exploitation. In \cref{sec:policy set}, we derive a set of policies by interpolating exploration and exploitation at the intra-episodic level. Based on this policy set, we design an adaptive meta-controller in \cref{sec:policy choosen} to choose an effective policy for interacting with the environment in each episode. \cref{fig:overview} provides an overview of our method.

\subsection{\texorpdfstring{Behavior Function $\beta$}{Behavior Function beta}}
\label{sec:basic polices}

% We construct three basic functions: two for exploration and one for exploitation.
% The behavior function $\beta$ is used to explore underexplored actions. The $Q$ function is used to explore overestimated actions. 
% The $Q_{\textit{mask}}$ which reduces unseen state-action values is used to go to the boundary of current experience, which is exploitation. 
% We get three basic functions with clear purposes, and the only extra computation comes from the learning of the behavior function $\beta$ comparing with DQN.

Learning the behavior function $\beta$ is straightforward. We sample a batch of data $B$ from the replay memory and train a network using supervised learning with cross entropy loss:
\begin{equation}
    \label{eq:behavior policy learning}
    \mathcal{L}_{\beta}=-\frac{1}{|B|}\sum_{(s,a) \in B}\log\beta (s,a).
\end{equation}
$\beta$ represents an average of the policies that collect data in the replay memory. It estimates the probability of each action that has been taken at each state. We use the same data batch to learn $\beta$ and $Q$, thus incurring no additional computational cost for sampling.

\textbf{Exploration for State-Action Coverage.} $\beta$ differentiates between actions that are frequently taken and those that are rarely taken. We sample actions with probabilities lower than a threshold $\delta$ to explore the state-action space for better coverage:
\begin{equation}
    a \sim \text{Unif}\{a: \beta(a|s) \le \delta \}
\label{eq:explore cov}
\end{equation}
Here, $\text{Unif}(\cdot)$ denotes selecting an action randomly from a given set, and $\delta \in [0,1]$ is a parameter. This policy is purely exploratory, focusing on better state-action coverage without considering the rewards obtained.

\textbf{Exploration for Overestimation Bias Correction.} Valued-based methods such as DQN estimate an action-value function $Q$ with temporal-difference (TD) learning with~\cref{eq:bellman}. The maximum operator in the Bellman update may lead to overestimation of action values~\citep{van2016deep,fujimoto2018addressing}. The greedy policy based on $Q$,
\begin{equation}
    a = \arg\max_a Q(s,a),
\label{eq:q greedy}
\end{equation}
is an optimistic policy that may take erroneously overestimated actions. 
This can be one kind of exploration that induces corrective feedback to mitigate biased estimation in $Q$ function~\citep{kumar2020discor,schaul2022the}.

An alternative way to benefit from $\beta$ is constraining the learning to in-sample state-action pairs~\citep{xiao2023the,bai2024efficient,zhang2023replay,zhang2024exploiting}:
\begin{equation}
    Q(s,a) \leftarrow r+\gamma\max_{a^\prime:\beta(a^\prime|s^\prime)>\epsilon}Q(s^\prime,a^\prime).
\label{eq:in distribution td learning}
\end{equation}
The max operator only bootstraps from actions well-supported in the replay memory, determined by $\beta(s,a)>\epsilon$, where $\epsilon$ is a small number.
% and we use $0.05$ across our experiments.
Because the data coverage in the replay memory is limited to a tiny subset of the entire environment space, when combing with deep neural networks, \cref{eq:in distribution td learning} learns an in-sample estimation at existing state-action pairs and generalizes to missing data.
While it may still overestimate out-of-distribution state-action pairs, leading to unrealistic values and inducing exploration to correct the bias, an additional benefit is that the learning process becomes more stable and exhibits better convergence properties.

\begin{proposition}
    \label{prop:in-sample td convergence}
        In the tabular case with finite state action space $\mathcal{S}\times \mathcal{A}$, the temporal difference learning masked by $\beta$ given in \cref{eq:in distribution td learning} uniquely converges to the optimal in-sample value $\widehat{Q}^*$ on explored state-action pairs. When $\beta(a|s)>\epsilon$ for all $a \in \mathcal{A}$, $\widehat{Q}^*$ equals to $Q^*$, which recovers the original temporal difference learning without action mask in \cref{eq:bellman}.
\end{proposition}

We prove the convergence of \cref{eq:in distribution td learning} by showing that the update rule is a $\gamma$-contraction mapping. The contraction property ensures that the update rule converges to a unique fixed point. The proof is provided in \cref{appendix:proof}.
    
This indicates that if the state-action space is fully covered, the two update rules are equivalent. When there are missing transitions, \cref{eq:in distribution td learning} converges on the explored state-action pairs. In contrast, \cref{eq:bellman} does not guarantee convergence even for the explored state-action pairs, as shown in previous work~\citep{fujimoto2022should}. 
% We leave the proof in \cref{appendix:proof} and compare the two learning rules in our experiments.
% but there is no guarantee on missing transitions. 

% This formula indeed tries to solve the empirical MDP formed by transitions in the replay memory, and it is similar to offline RL \citep{levine2020offline}, that tries to learn a best possible policy from a dataset \citep{kumar2020conservative,kostrikov2022offline,zhang2023insample,xiao2023the}. 
% Effective offline RL methods would be able to extract policies with the maximum possible utility out of the available data.

\textbf{Pure Exploitation.} Although the $Q$ function may overestimate, the behavior function $\beta$ can differentiate between frequently and rarely taken actions. By combining $Q$ with $\beta$, we can mask actions with low probabilities in $\beta$ before taking the greedy action of $Q$:
\begin{equation}
    a = {\arg\max}_{a:\beta(a|s)>\epsilon}Q(s,a),
\label{eq:q mask greedy}
\end{equation}
where $\epsilon$ is a small number. This policy is purely exploitative, aiming to maximize rewards based on the current experiences stored in the replay memory.

% This is similar to offline/batch reinforcement learning~\citep{lange2012batch,levine2020offline}. The learning goal is to maximize the cumulative reward limited to a static dataset. This returns a conservative policy with currently best possible performance~\citep{kumar2020conservative,kostrikov2022offline,shrestha2021deepaveragers,xiao2023the,zhang2023insample}.
% Now we have an optimistic $Q$ and a conservative $\hat{Q}$.
% In practice, to reduce the computation burden, we only learn one function $\hat{Q}$ using \cref{eq:in distribution td learning}.
% This learning rule gives us an estimate based on in-distribution data, while will still have bias at unseen state-action pairs. 
% Thus, when taking actions, the optimistic policy can be derived by:
% \begin{equation}
%     \pi = \argmax_a \hat{Q}(s,a),
%     \label{eq:optimistic policy}
% \end{equation}
% and the conservative policy can be derived by:
% \begin{equation}
%     \hat{\pi} = \argmax_{a:\beta(s,a)>0}\hat{Q}(s,a).
%     \label{eq:conservative policy}
% \end{equation}
% In our experiments, we compare learning two separate functions and only one function, and find no obvious difference in performance, and the later way is more computational efficient.

\subsection{Constructing Policy Set}
\label{sec:policy set}
Previous work \citep{pislar2022when} has shown that intra-episodic exploration, i.e., changing the mode of exploitation and exploration in one episode, is the most promising diagram.  With the three policies in \cref{eq:explore cov,eq:q greedy,eq:q mask greedy} with clear purposes, we interpolate exploration and exploitation at the intra-episodic level to construct a set of diverse policies that balance exploration between state coverage and overestimation bias correction.

% The principle idea is that we can combine two modes of behavior, an \textit{exploration} mode and an \textit{exploitation} mode. 
% For example, $\epsilon$-greedy is the combination of a random policy and a greedy policy.
% improves over both step-level (such as $\epsilon$-greedy) and episode-level (pure explore or exploit at one episode) baselines. 
% We interpolate between exploration policy and exploitation policy to get a set of policies that explore at some states and exploit at other states within one episode.
% We construct two kinds of exploration polices. One is to explore rarely taken actions for better state space coverage, the other is to explore overestimated actions and get corrective feedback for bias correction.

\textbf{Polices for State-Action Coverage.} One leaving question in \cref{eq:explore cov} is which action to take when all actions have probabilities higher than $\delta$. We address this by interpolating the pure exploitation policy from \cref{eq:q mask greedy} into the exploration policy in \cref{eq:explore cov} to create a policy that enhances state-action coverage:
\begin{equation}
    \pi_{\text{cov}(\delta)} :=     
    \begin{cases}
    {\arg\max}_{a:\beta(a|s)>\epsilon}Q(s,a), \quad \text{if} \ \ \beta(s,a) > \delta \ \ \forall \ \  a \in \mathcal{A} \\
    \text{Unif}\{a : \beta(a|s) \le \delta \} , \qquad \ \ \ \text{otherwise}
    % \text{DiscreteU}(\{a: \beta(s,a) \le \delta\}), \quad \text{otherwise}
    \end{cases}
    \label{eq:policy cov}
\end{equation}
The intuition behind $\pi_{\text{cov}(\delta)}$ is straightforward: if all actions at a state have been tried several times, we follow the pure exploitation mode to choose actions and reach the boundary of the explored area. Otherwise, we sample an action uniformly from the rarely taken actions, as determined by $\delta$. 

\begin{proposition}
    \label{prop:behavior coverage}
            In the tabular case with finite state action space $\mathcal{S}\times \mathcal{A}$ and finite horizon $H$, taking actions following policy $\pi_{\text{cov}(\delta)}$ guarantees infinite state-action visitations for all state action pairs. However, the expected cumulative regret is linear.
\end{proposition}

We prove this proposition by first demonstrating its validity on bandit problems and then extending the proof to MDPs. The detailed proof is provided in \cref{appendix:proof}.

Selecting actions according to $\pi_{\text{cov}(\delta)}$ ensures that all state-action pairs are visited infinitely often in the long run, guaranteeing the convergence of value iteration in the tabular case~\citep{watkins1992q,singh2000convergence}. By setting different values of $\delta$, we obtain a range of policies with varying degrees of exploration. Specifically, $\pi_{\text{cov}(0)}$ is the pure exploitation policy as defined in \cref{eq:q mask greedy}, while $\pi_{\text{cov}(\delta)}$ becomes a random policy when $\delta \ge 1/|\mathcal{A}|$.

\cref{prop:behavior coverage} also indicates that the total regret is linear if we only follow $\pi_{\text{cov}(\delta)}$ to choose actions. 
An excessive amount of steps can be wasted traveling through already-explored but unpromising states, reducing overall efficiency. Since the goal of learning is to obtain an accurate action value function, we design exploration to try overestimated actions and correct the estimation bias.

\noindent\textbf{Polices for Overestimation Bias Correction.} 
Action value-based algorithms are known to overestimate action values~\citep{van2016deep,fujimoto2018addressing}. Accurate value estimation is critical for extracting a good policy in DRL. We interpolate between the overestimated policy in \cref{eq:q greedy} and the pure exploitation policy in \cref{eq:q mask greedy} to create a policy that explores overestimated actions for corrective feedback:
\begin{equation}
    \pi_{\textit{cor}(\alpha)} = \arg\max_a(\alpha Q(s,a) + (1-\alpha) \hat{Q}(s,a)).
    \label{eq:policy cor}
\end{equation}
Here, $\hat{Q}$ is the value function where we suppress the overestimated action values identified by $\beta$,
\begin{equation}
    \hat{Q}(s,a) =     
    \begin{cases}
    Q(s,a), \qquad \qquad \quad \text{if} \ \ \beta(s,a) > \epsilon  \\
    \min_{a\in \mathcal{A}} Q(s,a), \ \ \quad \text{otherwise}
    \end{cases}
    \label{eq:q mask}
\end{equation}
The intuition is to follow the current best actions at some states while exploring overestimated actions at others. The parameter $\alpha \in [0,1]$ allows us to set different values to obtain policies with varying degrees of exploration for bias correction. Specifically, $\pi_{\text{cor}(0)}$ recovers the pure exploitation policy in \cref{eq:q mask greedy}, and $\pi_{\text{cor}(1)}$ recovers the overestimated policy in \cref{eq:q greedy}.

\textbf{Constructing Policy Set.} By setting different values for $\delta$ and $\alpha$, 
% in $\pi_{\text{cov}}$ and $\pi_{\text{cor}}$, 
we generate a policy set $\Pi$ that ranges from exploration for better state-action coverage to overestimation bias correction:
\begin{equation}
\label{eq:policy set}
    \Pi = \left\{\pi_{\text{cov}(\delta_1)},\cdots,\pi_{\text{cov}(\delta_m)}, \pi_{\text{cor}(\alpha_1)},\cdots,\pi_{\text{cor}(\alpha_n)} \right\}.
\end{equation}
This policy set does not inject specialized inductive biases, making it a general method across a wide range of tasks. Additionally, the computational cost does not increase when adding more policies with different $\delta$ and $\alpha$ values.

\begin{figure*}[htbp]
    \centering
    \vspace{-0.15in}
    % {\small The first figure shows the state-action pairs in the memory, indicating taking action with low probabilities of $\beta$ will try these missing actions. The second and third figures show the masked/unmasked action values and the corresponding actions taken at each state.}
    \subfigure[]{\Description{}
    \label{fig:toy example(a)}
    \includegraphics[width=0.3\textwidth]{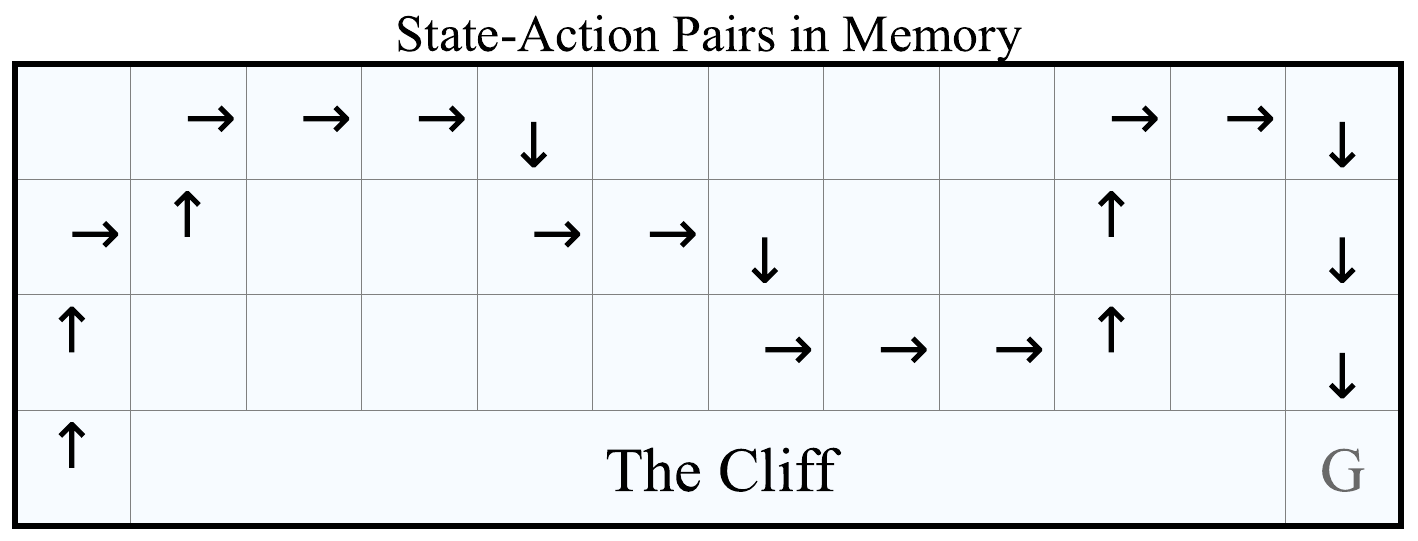}
    \includegraphics[width=0.3\textwidth]{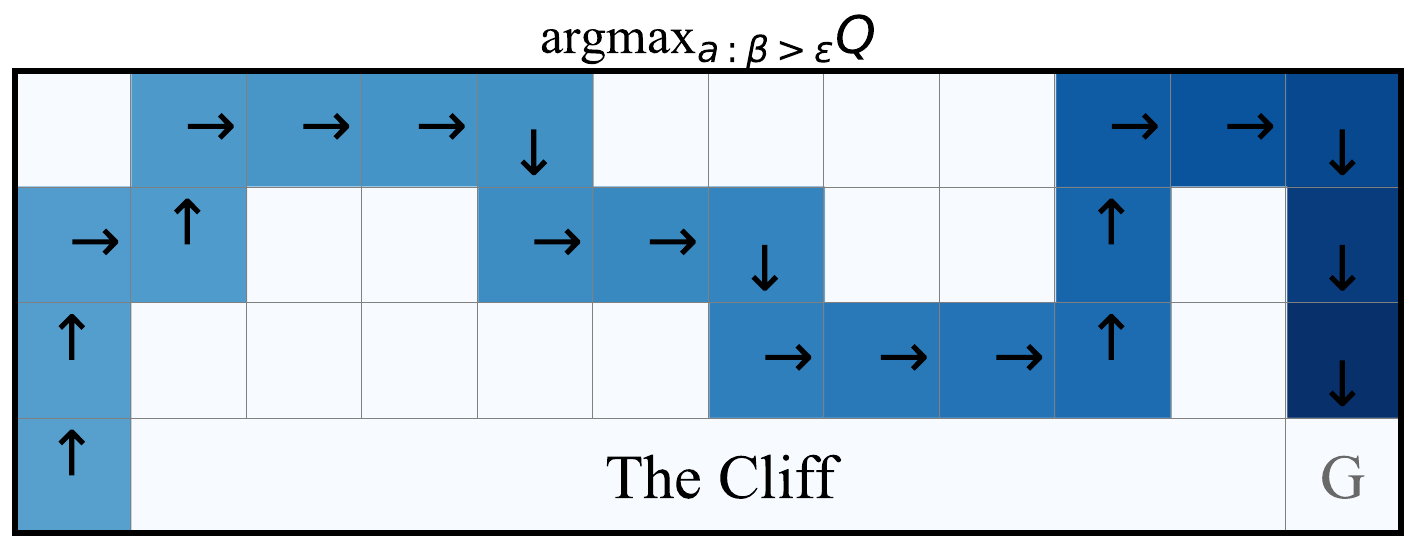}
    \includegraphics[width=0.3\textwidth]{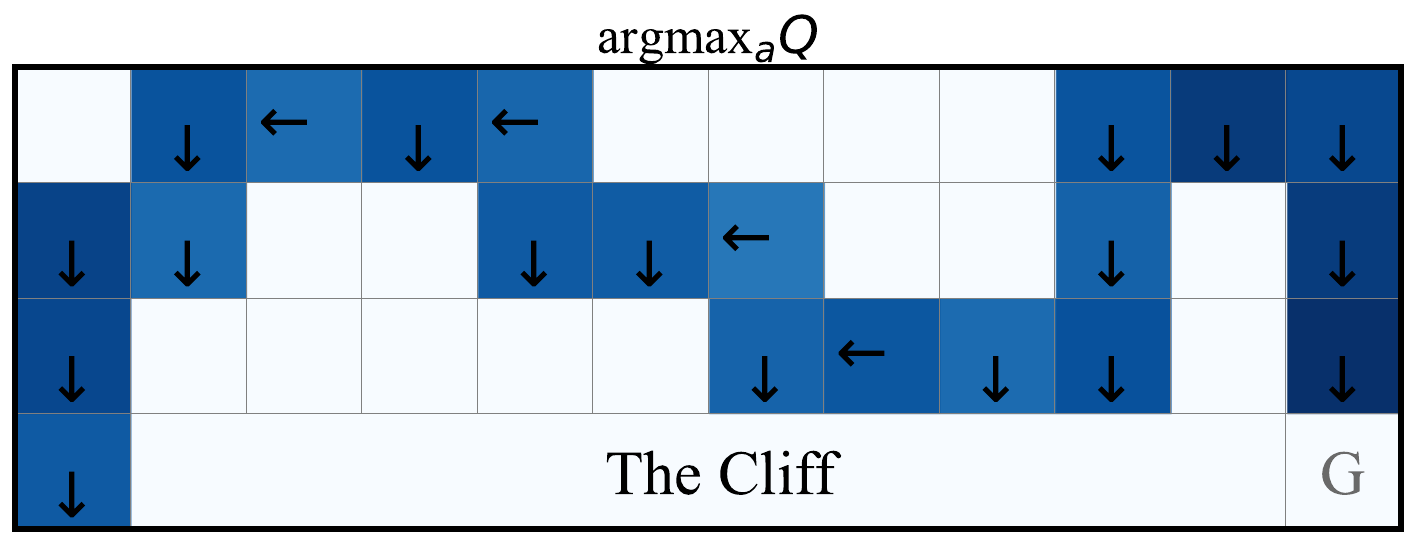}
    \includegraphics[width=0.02\textwidth, height=0.105\textwidth]{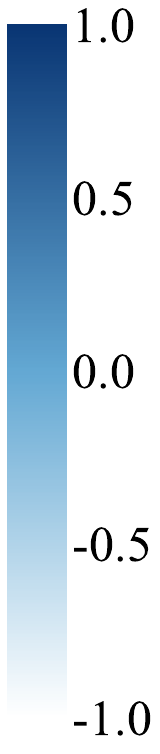}}
    \vskip -0.1in
    % \small Diverse polices achieved when we assign different $\delta$ and $\alpha$.
    
    \includegraphics[width=0.23\textwidth]{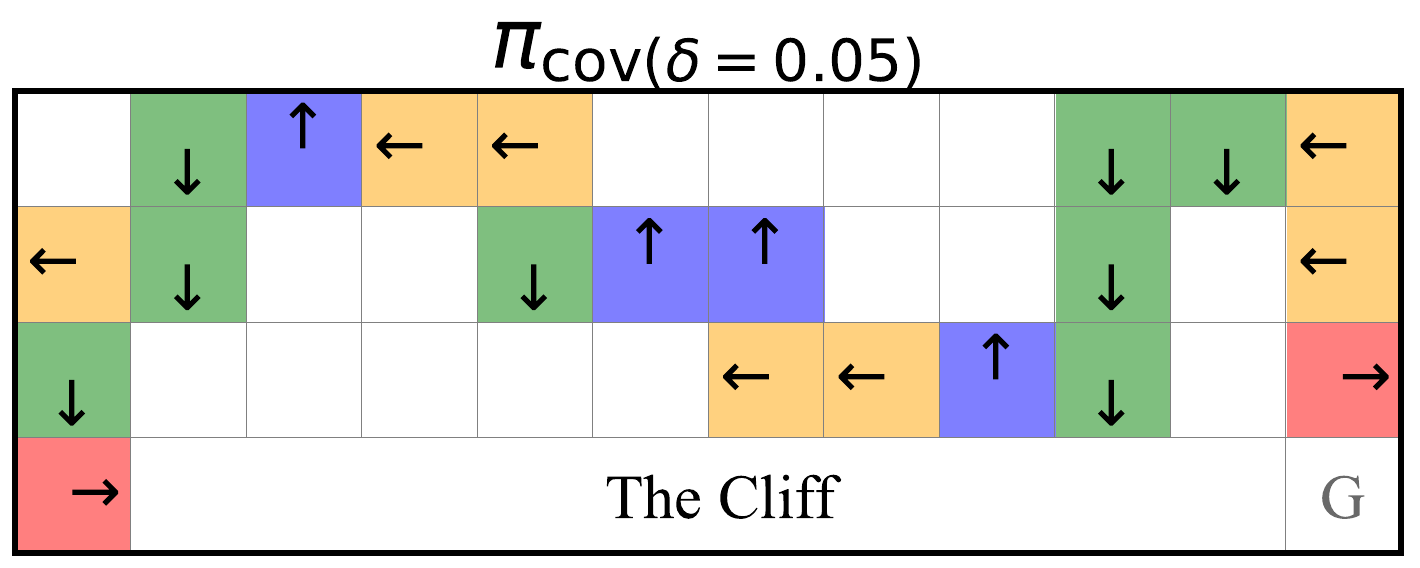}
    \includegraphics[width=0.23\textwidth]{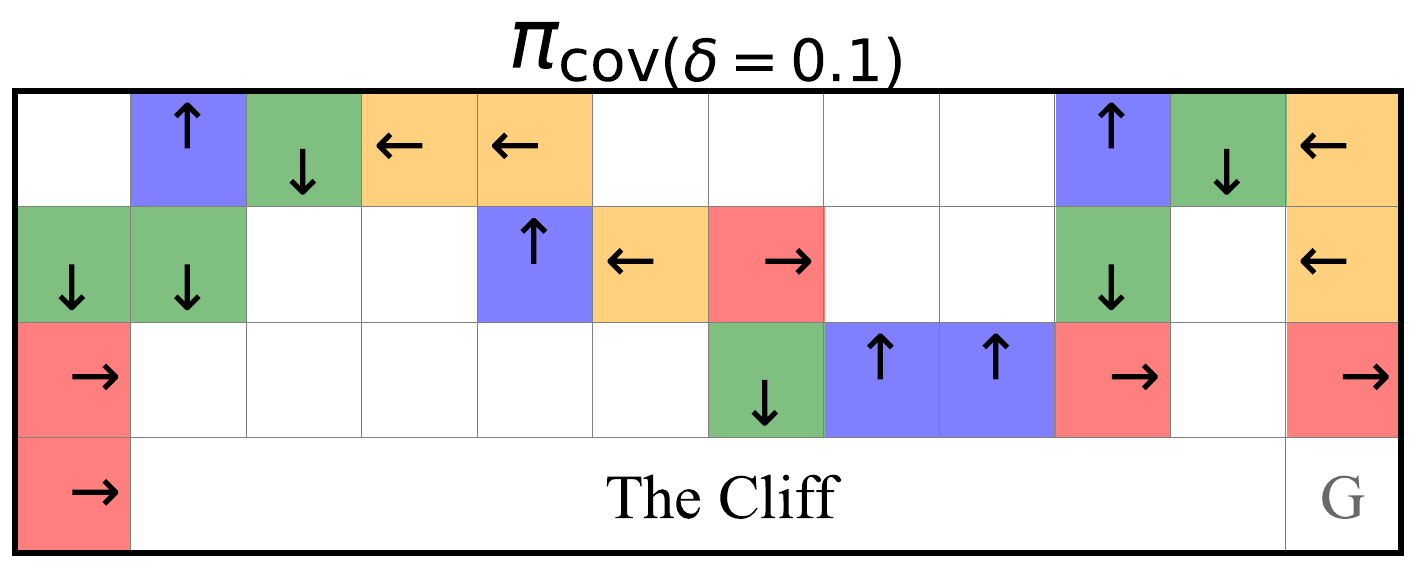}
    \includegraphics[width=0.23\textwidth]{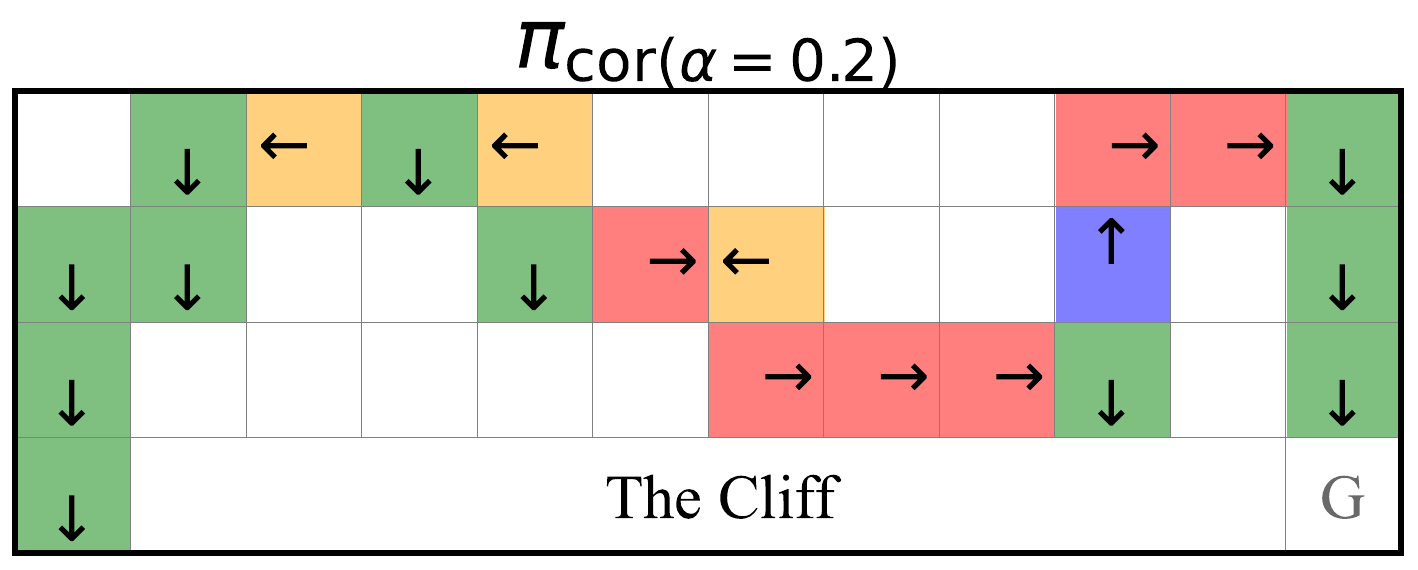}
    \includegraphics[width=0.23\textwidth]{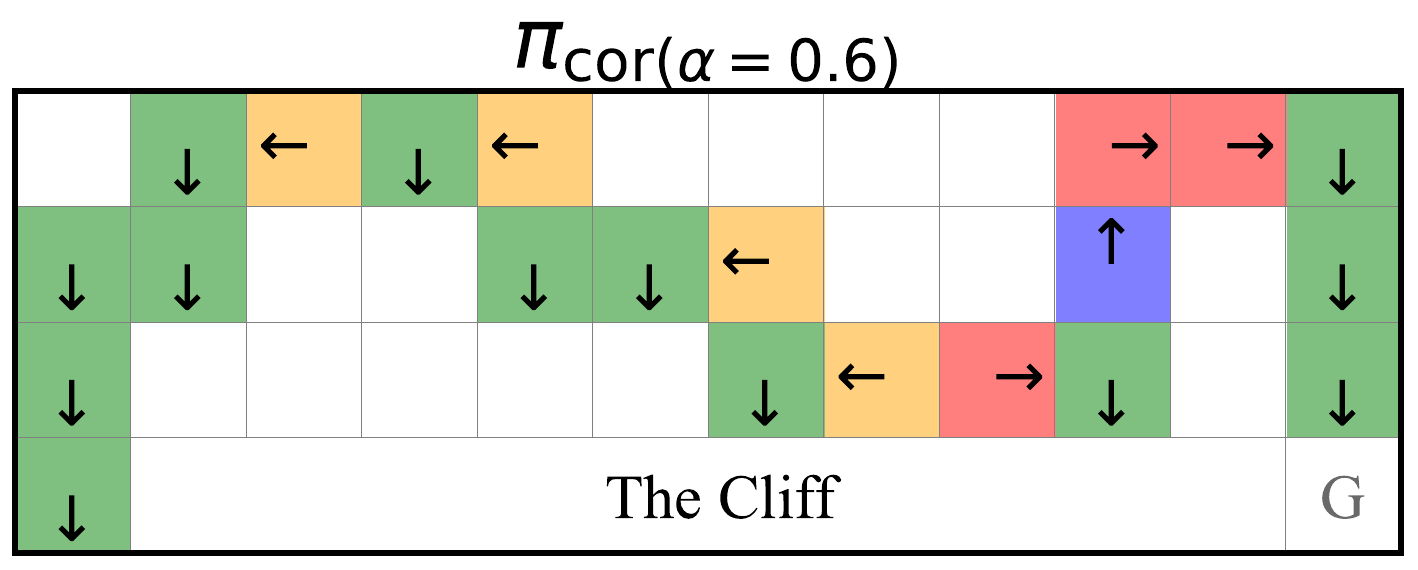}
    \vskip -0.02in
    \subfigure[]{\Description{}\label{fig:toy example(b)}
    \includegraphics[width=0.35\textwidth]{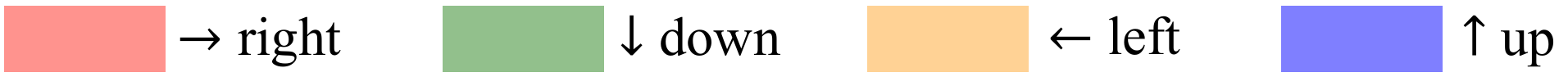}}
    \vskip -0.2in
    % \caption{Policy diversity in a specific case: Consider a scenario where there is only one suboptimal trajectory in the replay memory. Our method constructs diverse policies that benefit learning.}
    \caption{(a) The first image shows the state-action pairs in memory. It implies that taking actions with low probabilities according to $\beta$ will try missing actions. The second and third images show the masked/unmasked Q values and the corresponding actions. (b) Policies with different $\delta$ and $\alpha$, demonstrating the effectiveness of our method in constructing a diverse policy set.}
    % \vskip -0.15in
    \label{fig:toy example}
\end{figure*}

\begin{figure*}[htbp]
    % \vskip -0.1in
\begin{center}
    \subfigure{\Description{}\includegraphics[width=0.9\textwidth]{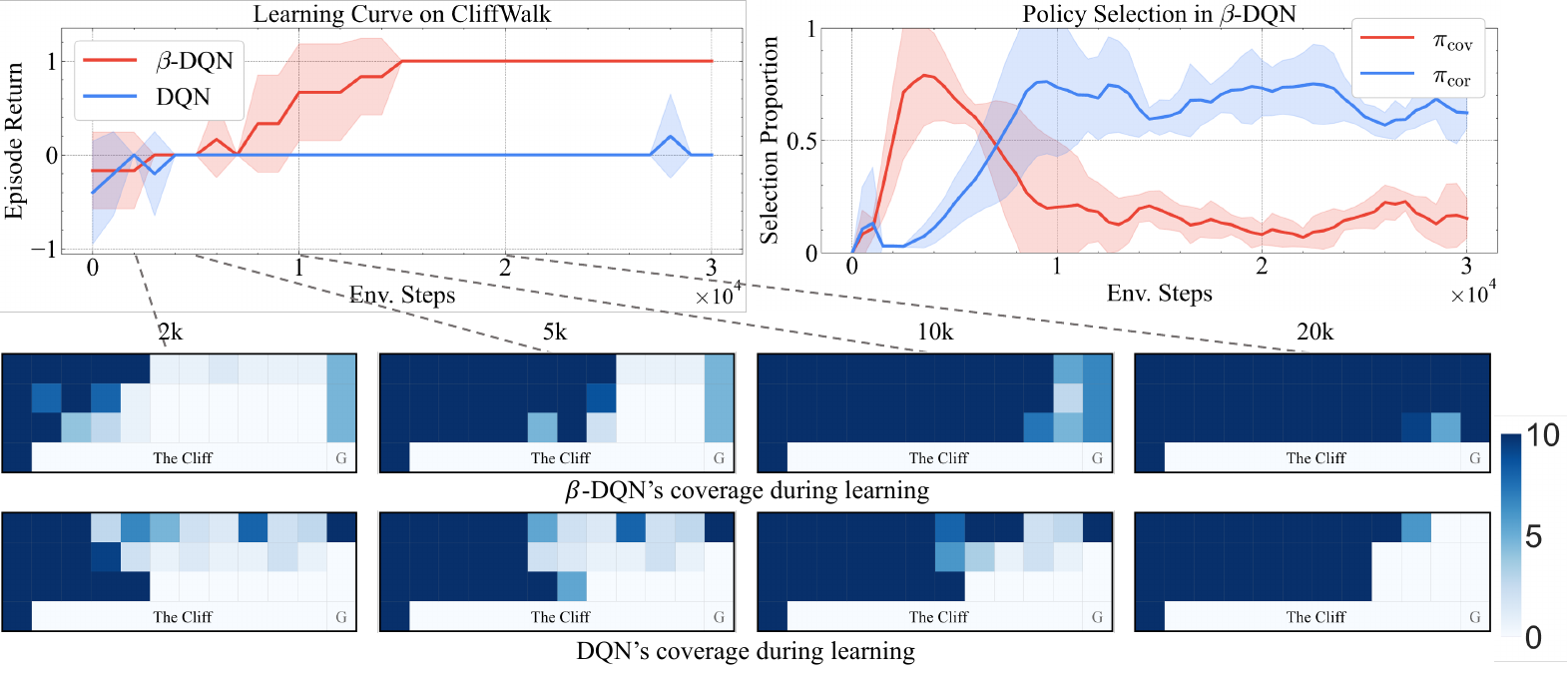}}
    \vskip -0.25in
    \caption{
    \textbf{Details during the learning.}
    (1) The top left learning curves show that $\beta$-DQN successfully reaches the goal state, while DQN only avoids the cliff. 
    (2) The heatmaps illustrate that $\beta$-DQN explores the entire state space efficiently.
    (3) The top right curves show that $\beta$-DQN initially uses $\pi_{\text{cov}}$ for state space exploration, then switches to $\pi_{\text{cor}}$ for correcting biased estimations.
    % (3) The top right curves detail the policy selection within $\beta$-DQN, initially focusing on exploring the state space using $\pi_{\text{cov}}$ before transitioning to correct biased estimations with $\pi_{\text{cor}}$.
     % The top left shows the learning curves of $\beta$-DQN and DQN. The heatmaps show the state coverage during learning. Our method explores the whole space quickly and finds the path to the goal state which gives reward 1. In contrast, DQN agent learns how to avoid falling into the cliff but fails to reach the goal state G. The top right figure further shows the policy selection in $\beta$-DQN. It first tries to explore the state space with $\pi_{\textit{cov}}$, and then converts to correct inaccurate estimation using $\pi_{\textit{cor}}$.
    }
    % \vskip -0.1in
    \label{fig:data coverage}
\end{center}
\end{figure*}

\begin{algorithm}[t!]
    \caption{$\beta$-DQN}
    \label{alg:alg1}
  \begin{algorithmic}[1]
    \STATE Initialize replay memory $\mathcal{D}$ with fixed size
    \STATE {\color{blue}Initialize functions $\beta,Q$ and construct policy set $\Pi$ following \cref{eq:policy cov,eq:policy cor,eq:policy set}}
    \FOR{ episode $k$ = 0 {\bfseries to} $K$}
    \STATE {\color{blue} Select a policy $\pi$ according to \cref{eq:policy select}}
    \STATE Initialize the environment $s_0\leftarrow Env$
    \FOR{ environments step $t$ = 0 {\bfseries to} $T$}
    \STATE Select an action $a_t \sim \pi(\cdot|s_t)$ 
    \STATE Execute $a_t$ in $Env$ and get $r_t,s_{t+1}$
    \STATE Store transition $(s_t,a_t,r_t,s_{t+1})$ in $\mathcal{D}$
    \STATE {\color{blue} Update $\beta$ and $Q$ following \cref{eq:behavior policy learning,eq:in distribution td learning}}
    \ENDFOR
    \ENDFOR
\end{algorithmic}
\end{algorithm}

\vspace{-0.1in}
\subsection{Meta-Controller for Policy Selection}
\label{sec:policy choosen}

After constructing a set of policies, we need to select an effective policy to interact with the environment for each episode. Similar to previous work~\citep{badia2020agent57,fan2022generalized,fan2023learnable,kim2023lesson}, we consider the policy selection problem as a non-stationary multi-armed bandit (MAB)~\citep{garivier2008upper,lattimore2020bandit}, where each policy in the set is an arm. We design a meta-controller to select policies adaptively.

Assume there are $N$ policies in the policy set $\Pi=\{\pi_0,\cdots, \pi_{N-1}\}$. For each episode $k \in \mathbb{N}$, the meta-controller selects a policy $A_k = \pi_i$ and receives an episodic return $R_k(A_k)$. Our objective is to obtain a policy $\pi$ that maximizes the return within a given interaction budget $K$.

Due to the non-stationarity of the policies, we consider the recent $L<K$ results. Let $N_k(\pi_i,L)$ be the number of times policy $\pi_i$ has been selected after $k$ episodes, and $\mu_k(\pi_i,L)$ be the mean return $\pi_i$ obtained by $\pi_i$. We design a bonus $b$ to encourage exploration. An action is considered exploratory if it differs from the pure exploitation action taken by \cref{eq:q mask greedy}. The exploration bonus for policy $\pi_i$ is computed as:
\begin{equation}
    b_k(\pi_i,L) = \frac{1}{N_k(\pi_i,L)}\sum^{k-1}_{m=\max(0,k-L)}B_m(\pi_i)\mathbb{I}(A_m=\pi_i),
\end{equation}
where $B_m(\pi_i)$ computes the ratio of exploration actions taken by policy $\pi_i$ at episode $m$ and $\mathbb{I}(\cdot)$ is the indicator function.

To select a policy for episode $k$, we consider the return and exploration bonus of each policy within the sliding window $L$:
\begin{equation}
\label{eq:policy select}
    A_k =     
    \begin{cases}
    \pi_i, \qquad \qquad \qquad \qquad \qquad \quad \text{if} 
    % \ \ \exists \ \pi_i \in \Pi \ \mathrm{s.t.} 
    \ N_k(\pi_i,L) = 0, \\
    \arg\max_{\pi_i} (\mu_k(\pi_i,L) + b_k(\pi_i,L)),  \quad \text{otherwise.}
    \end{cases}
\end{equation}
In this formula, if a policy has not been selected in the last $L$ episodes, we will prioritize selecting that policy. Otherwise, the policy that explores more frequently and also gets higher returns is preferred. \cref{alg:alg1} summarizes our method.

% In implementation, we normalize the return into $[0,1]$ by $(R_k-R_{\text{min}})/(R_{\text{max}}-R_{\text{min}})$, where $R_{\text{max}}$ and $R_{\text{min}}$ are the maximum and minimum return in the sliding window. Thus $\mu_k$ and $b_k$ are at the same magnitude.

% We propose to follow previous work \citep{badia2020agent57,fan2022generalized,fan2023learnable} and adaptation of these settings to a meta-controller, which is implemented as a non-stationary multi-armed bandit (MAB) that maximises the current episodic return and also considering exploration that may benefit in the future.
% For each episode $k \in \mathbb{N}$, a MAB algorithm chooses an arm $A_k$ among the possible arms ${0,\cdots, N-1}$, i.e. a policy $\pi \in \Pi$, that is conditioned on the sequence of previous actions and rewards. 
% Doing so, it receives a reward $R_k(A_k) in \mathbb{R}$.
% The rewards $\{R_k(a)\}_{k \ge 0}$ are modelled by a sequence of independent random variables whose distributions could change through time due to the non-stationarity.
% We use a sliding-window to store the most recent rewards to reflect the performance of exploitation and exploration for the possible arms, i.e. the policy set $\Pi$. 
% The goal of a MAB algorithm is to find a policy $\pi$ that maximizes the expected cumulative reward for a given horizon $K$ : $\mathbb{E}_\pi[\sum^{K-1}_{k=0}R_k(A_k)]$.
% Let the length of sliding window be $L \in \mathbb{N^*}$ to adapt to the change of reward distribution.
% It is commonly understood that the window length $L$ should be way smaller that the horizon $K$.

\section{Experiments}
\label{sec:experiments}
In this section, we aim to answer the following questions:
% \textbf{(1)} Does our method lead to diverse exploration, thereby benefiting learning?
% \textbf{(2)} Can our method improve performance in both dense and sparse reward domains while maintaining a mild computational cost?
% \textbf{(3)} What roles do the exploration policies $\pi_{\text{cov}}$ and $\pi_\text{cor}$ play in different environments?
% \textbf{(4)} Is there a difference when learning $Q$ using TD learning with and without constraints by $\beta$?
\begin{itemize}[leftmargin=*,noitemsep]
\item Does our method lead to diverse exploration, thereby enhancing the learning process and overall performance?
\item Can $\beta$-DQN improve performance in both dense and sparse reward environments while maintaining a low computational cost?
\item How do the exploration policies $\pi_{\text{cov}}$ and $\pi_{\text{cor}}$ contribute to the learning process in different environments?
\item Is there a difference when learning $Q$ with constraints imposed by $\beta$ compared to without such constraints?
\end{itemize}

\subsection{A Toy Example}
\label{sec:toy example}
In this section, we present a toy example using the CliffWalk environment~\citep{sutton2018reinforcement}, to illustrate the policy diversity and the learning efficacy of our method. The CliffWalk environment comprises 48 states and 4 actions, as depicted in~\cref{fig:toy example}. Starting from the bottom left, the goal is to navigate to state G located at the bottom right. Reaching G yields a reward of 1, while falling into the cliff incurs a penalty of -1; all other moves have a reward of 0. 

\begin{table*}[t]
    % \vskip -0.3in
    \vspace{-0.15in}
    \caption{
    \textbf{Overall performance on MiniGrid (Success Rate in $[0,1]$) and MinAtar (Episode Return).} Bold numbers indicate the method that achieves the best performance. Our method outperforms others in most games with a mild computational cost.
    }
    \label{table:overall performance}
    \vspace{-0.1in}
        \centering
        \scalebox{1}{
            \begin{tabular}{cccccccc}
                \toprule[1pt]
                \multicolumn{2}{c}{Environment} &DQN&Bootstrapped DQN&$\epsilon$z-greedy&RND&LESSON&$\beta$-DQN (Ours)\\ \midrule
                \multirow{7}*{MiniGrid} &DoorKey &0.44 & 0.11 & 0.0 & \textbf{0.99} & 0.86 & 0.98\\
                ~&Unlock &0.22 & 0.17 & 0.0 & 0.95 & 0.64 & \textbf{0.99} \\
                % ~&RedBlueDoors &0.43 & 0.46 & 0.0 & \textbf{1.0} & 0.73 & 0.38 \\
                ~&SimpleCrossing-Easy &\textbf{1.0} & \textbf{1.0} & 0.95 & 0.95 & 0.97 & 0.99\\
                ~&SimpleCrossing-Hard &\textbf{1.0} & 0.81 & 0.05 & 0.93 & 0.6 & \textbf{1.0}\\
                ~&LavaCrossing-Easy &0.29 & 0.66 & 0.26 & 0.68 & 0.75 & \textbf{0.84}\\
                ~&LavaCrossing-Hard &0.0 & 0.01 & 0.0 & \textbf{0.39} & 0.06 & 0.16 \\ \cdashline{2-8}\cdashline{2-8}
                ~&Average &0.49 & 0.46 & 0.21 & 0.82 & 0.65 & \textbf{0.83}\\
                \hline
                \multirow{5}*{MinAtar} &Asterix &22.78 & 22.54 & 18.79 & 13.4 & 18.43 & \textbf{39.09}\\
                ~&Breakout &16.69 & 21.88 & 19.06 & 14.1 & 17.71 & \textbf{29.04} \\
                ~&Freeway &60.78 & 59.94 & 59.68 & 49.26 & 54.38 & \textbf{62.56}\\
                ~&Seaquest &14.66 & 14.31 & 16.98 & 5.61 & 9.41 & \textbf{33.23}\\
                ~&SpaceInvaders &67.28 & 69.91 & 68.7 & 31.58 & 55.94 & \textbf{98.28}\\ \cdashline{2-8}\cdashline{2-8}
                ~&Average &36.44 & 36.55 & 36.64 & 22.79 & 31.17 & \textbf{52.44}\\
                \hline
                \multicolumn{2}{c}{\multirow{1}{*}[-0.05cm]{Computational Cost}}  & \multirow{1}{*}[-0.05cm]{100\%} & \multirow{1}{*}[-0.05cm]{195.34 \%} & \multirow{1}{*}[-0.05cm]{\textbf{94.32 \%}} & \multirow{1}{*}[-0.05cm]{152.57 \%} & \multirow{1}{*}[-0.05cm]{371.07 \%}& \multirow{1}{*}[-0.05cm]{138.78 \%}\\
                \hline
                \multicolumn{2}{c}{\multirow{1}{*}[-0.05cm]{Performance/Computational Cost}}  & \multirow{1}{*}[-0.05cm]{1} & \multirow{1}{*}[-0.05cm]{0.50} & \multirow{1}{*}[-0.05cm]{0.76} & \multirow{1}{*}[-0.05cm]{0.75} & \multirow{1}{*}[-0.05cm]{0.29}& \multirow{1}{*}[-0.05cm]{\textbf{1.13}} \\
                \bottomrule[1pt]
            \end{tabular}
        }
            % \vspace{1.5ex}
    \end{table*}

For a clear illustration of policy diversity, we design a scenario with only one suboptimal trajectory in the replay memory (the left image in \cref{fig:toy example(a)}). The function $\beta$, learned from this trajectory, assigns a probability of 1 to the only existing action at each state, while assigning zero probabilities to other actions. The second and third images in \cref{fig:toy example(a)} show the action values masked/unmasked by $\beta$, and the corresponding actions taken at each state. This gives us a clear understanding of what actions are taken by different policies introduced in \cref{sec:basic polices}.

In \cref{fig:toy example(b)}, by assigning different values to $\delta$ and $\alpha$, we generate a group of diverse policies. Each policy takes different actions, leading to novel states or those with biased estimates. 
% Different colors represent different actions to clearly illustrate the policy diversity. 
Colors are used to differentiate actions and illustrate policy diversity.
This indicates that our method can create a diverse set of policies by simply interpolating between exploration and exploitation policies.
% The second and third images in \cref{fig:toy example(a)} illustrates the estimation errors of the function $Q$ learned by~\cref{eq:in distribution td learning} for both seen and unseen actions. We can observe that $Q$ learns accurate estimates for seen actions but inaccurate for unseen ones, indicating a potential overestimation in some states. 

\Cref{fig:data coverage} further details the online learning process. In the top left, the learning curves show $\beta$-DQN outperforms standard DQN by reaching the goal state and obtaining the reward, while DQN primarily learns to avoid the cliff without reaching the goal. This distinction is further illustrated in state coverage shown in the images below. Unlike DQN, which avoids areas near the cliff, $\beta$-DQN consistently explores the entire state space, including the challenging regions near the cliff and the goal state. 
Regarding policy selection, the top right image indicates that $\beta$-DQN initially favors space coverage strategies ($\pi_{\text{cov}}$). 
Once good coverage is achieved, the exploration shifts towards bias correction strategies ($\pi_{\text{cor}}$) in this sparse reward environment. This strategic shift highlights $\beta$-DQN’s adaptability in optimizing exploration to enhance overall learning performance in challenging settings.
For a broader perspective, an additional example is provided in~\cref{appendix:Additional Experimental Results}~\cref{fig:another toy example}.

\subsection{Overall Performance}

\textbf{Environments.} We evaluate our method on MiniGrid~\citep{gym_minigrid} and MinAtar~\citep{young19minatar} based on the OpenAI Gym interface~\citep{brockman2016openai}. MiniGrid presents numerous tasks in a grid world, characterized by sparse rewards that pose significant challenges in achieving high success rates. 
MinAtar is an image-based miniaturized version of Atari games~\citep{bellemare2013arcade}, preserves the core mechanics of the original games while significantly enhancing processing speed, which facilitates faster model training. In MiniGrid, maps are randomly generated in each episode, and in MinAtar, object placements vary across different time steps, necessitating robust policy generalization. For the evaluation metrics, MiniGrid measures the success rate between 0 and 1, while MinAtar employs episode return. 
% More details about these environments can be found in Appendix~\ref{appendix:environment details}.

\textbf{Baselines and Implementation Details.} We compare $\beta$-DQN with DQN~\citep{mnih2015human}, Bootstrapped DQN~\citep{osband2016deep}, $\epsilon$z-greedy~\citep{dabney2021temporallyextended}, RND~\citep{burda2018exploration}, and LESSON~\citep{kim2023lesson}. These algorithms are derivatives of DQN, differing primarily in their exploration strategies. RND targets environments with sparse rewards, while the others are general for all kinds of domains. A policy gradient method~\citep{schulman2017proximal} is included as a reference in~\cref{appendix:Additional Experimental Results}~\cref{fig:performance}.
For fair comparison, we use the same network architecture for all algorithms as used in MinAtar~\citep{young19minatar}. Learning rates for the baselines are searched over $\{3e^{-4}, 1e^{-4}, 3e^{-5}\}$, reporting the highest performance achieved. 
Our approach introduces additional hyper-parameters, yet employs a consistent parameters set across all environments. We instantiate the policy set as $\Pi = \{\pi_{\text{cov}(0.05)},$ $\pi_{\text{cov}(0.1)},$ $\pi_{\text{cor}(0)},$ $\pi_{\text{cor}(0.1)},$ $\pi_{\text{cor}(0.2)},$ $\cdots, \pi_{\text{cor}(1)} \}$. 
% For parameter $L$, a grid search over $\{100, 500, 1000, 2000\}$ on SimpleCrossing-Easy and Asterix reveals no significant performance differences for $L \geq 500$, but lower performance for $L=100$. Thus, $L=1000$ is fixed for all environments.
% For parameter $\epsilon$, we fix it at 0.05, with the sensitivity analysis provided in~\cref{appendix:Additional Experimental Results}~\cref{fig:parameter search epsilon}.
For the parameter $L$, we search over $\{100,$ $500,$ $1000,$ $2000\}$ on SimpleCrossing-Easy and Asterix. We find no significant differences between 500, 1000 and 2000, but observe lower performance for $L=100$. Based on these results, we fix $L=1000$ for all environments. 
For the parameter $\epsilon$, we use a fixed value of $\epsilon=0.05$, with the sensitivity analysis provided in~\cref{appendix:Additional Experimental Results}~\cref{fig:parameter search epsilon}.
Other common parameters are outlined in~\cref{appendix:implementation details}~\cref{appendix:DQN_hyperparameter}. Each algorithm is assessed using 10 different random seeds, with each run consisting of 5 million steps. Performance evaluation occurs every 100k steps over 30 episodes.

% \begin{table}[b]
%     % \vskip -0.1in
%     % \vskip 0.2in
%     \caption{Wall-clock time comparison between different methods. We use \emph{Frames Per Second} (FPS) to measure the speed of interaction with environments during training. Our method adds mild computational overhead on DQN. \hongming{performance/compute cost}}
%     \label{table:wall clock time comparison}
%         \centering
%         \scalebox{1.}{
%             \begin{tabular}{ccc}
%                 \toprule[1pt]
%                 \textbf{Method} &\textbf{FPS} (mean $\pm$ std) & \textbf{Computational Cost}\\ \midrule
%                 DQN &  870.64 $\pm$ 7.59 & 100\%  \\
%                 Bootstrapped DQN & 445.72 $\pm$ 14.68 & 195.34\%  \\ 
%                 $\epsilon$z-greedy & 923.04 $\pm$ 38.58 & 94.32\%  \\
%                 RND & 570.63 $\pm$ 32.31 & 152.57\%  \\
%                 LESSON & 234.63 $\pm$ 12.71 & 371.0\%  \\
%                 $\beta$-DQN (Ours) & 627.36 $\pm$ 5.34 & 138.78\%  \\
%                 \bottomrule[1pt]
%             \end{tabular}
%         }
%         %	\vspace{1.5ex}
%     \end{table}

\textbf{Performance.} The final performance is presented in~\cref{table:overall performance}, displaying the mean success rate on MiniGrid and mean return on MinAtar. Our method consistently outperforms others across a diverse range of environments, effectively addressing both dense and sparse reward scenarios. Bootstrapped DQN shows modest improvement on MinAtar and MiniGrid, indicating its generality but limited improvement on performance. However, $\epsilon z$-greedy fails to deliver significant improvements on MinAtar and suffers a substantial decrease on MiniGrid, as repetitive unguided actions often result in the agent colliding with obstacles, thereby squandering numerous trials. This inefficiency underscores the limitations of state-independent exploration, even when augmented by temporal persistence. RND excels in sparse reward settings but is the least effective on MinAtar, highlighting its lack of versatility. LESSON, while somewhat improving MiniGrid, performs poorly on MinAtar and incurs a significantly higher computational cost (371\%) compared to DQN. In contrast, our method increases computational demand by only 38\%. $\epsilon z$-greedy run slightly faster than DQN by selecting random actions for random durations, reducing demands on the $Q$ network’s inference. 

The last row of~\cref{table:overall performance} highlights the performance/computational-cost ratio, which considers both performance and computational cost. The ratio is calculated as the performance divided by the computational cost relative to DQN. Our method achieves the highest value of 1.13, indicating superior performance relative to computational cost, while other methods fall short. 
In summary, $\beta$-DQN proves to be general, effective, and computationally efficient. 
Detailed learning curves, including mean values and confidence intervals, are provided in~\cref{fig:performance} of~\cref{appendix:Additional Experimental Results}.

\subsection{Analysis of Our Method}
\label{sec:Analysis of our method}
Our method construct two types of exploration polices in our policy set based on the behavior function $\beta$: $\pi_{\text{cov}}$ for state-action coverage and $\pi_{\text{cor}}$ for overestimation bias correction. A meta-controller is then used to select an effective policy for each episode. 
We explore several interesting questions:
(1) What type of policy in the policy set is preferred by the meta-controller during learning?
(2) Which policy performs better, $\arg\max_a Q$ or ${\arg\max}_{a:\beta(a|s)>\epsilon}Q(s,a)$?

\begin{figure}[t]
    \vskip -0.15in
\begin{center}
    \subfigure{\Description{}\includegraphics[width=0.235\textwidth]{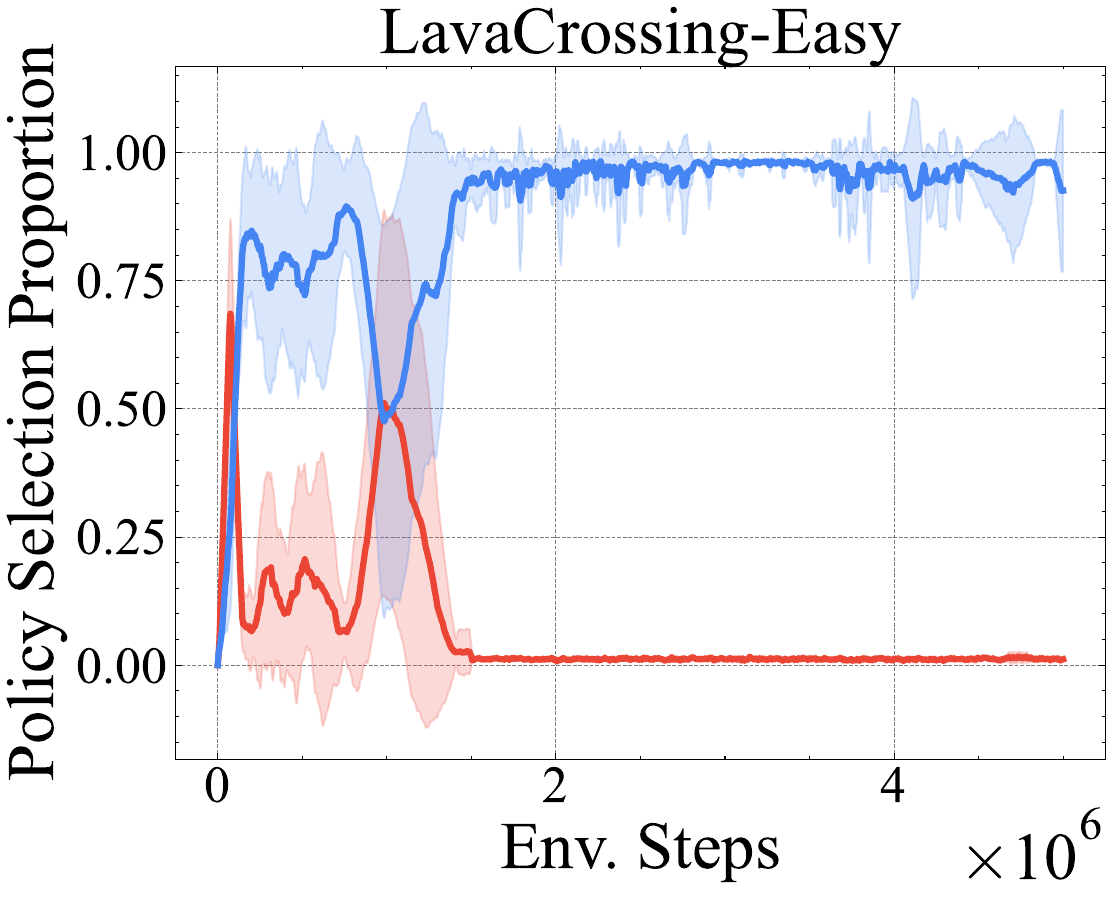}}
    \subfigure{\Description{}\includegraphics[width=0.235\textwidth]{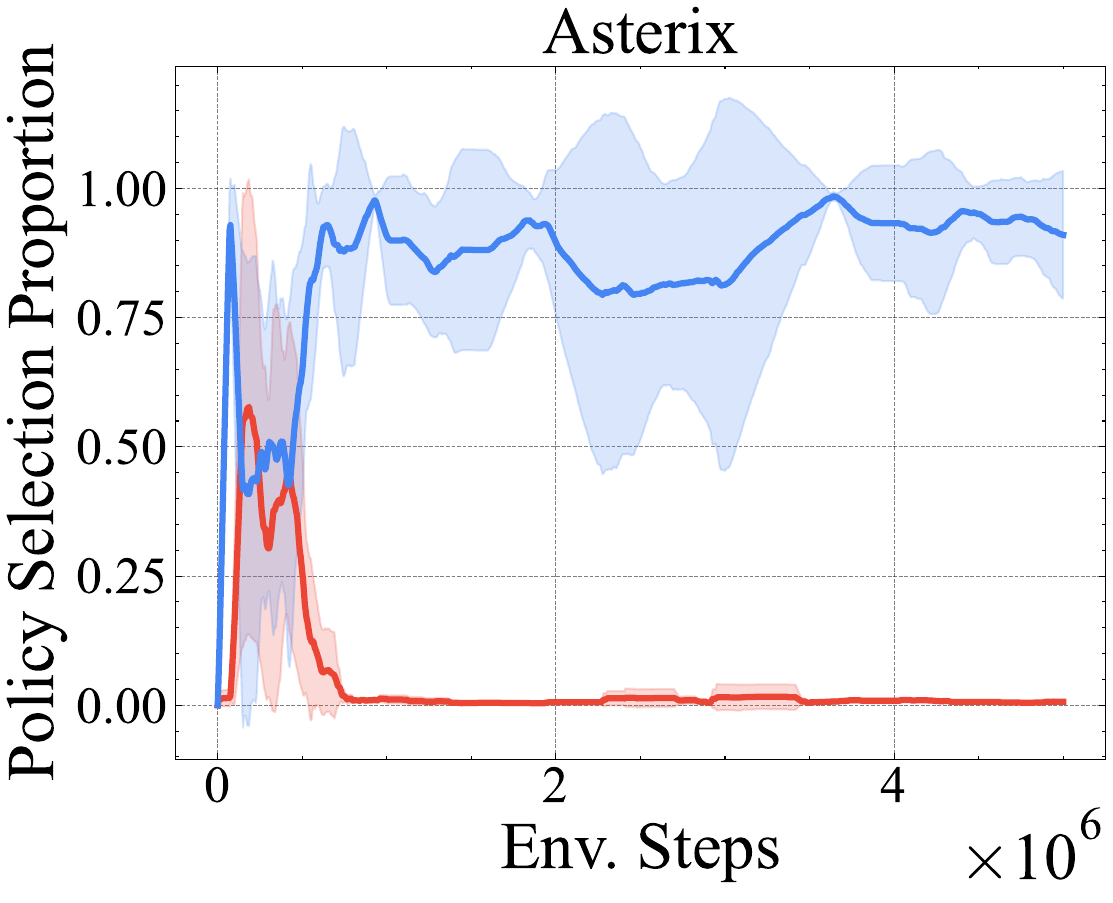}}
    \vskip -0.15in
    \subfigure{\Description{}\includegraphics[width=0.23\textwidth]{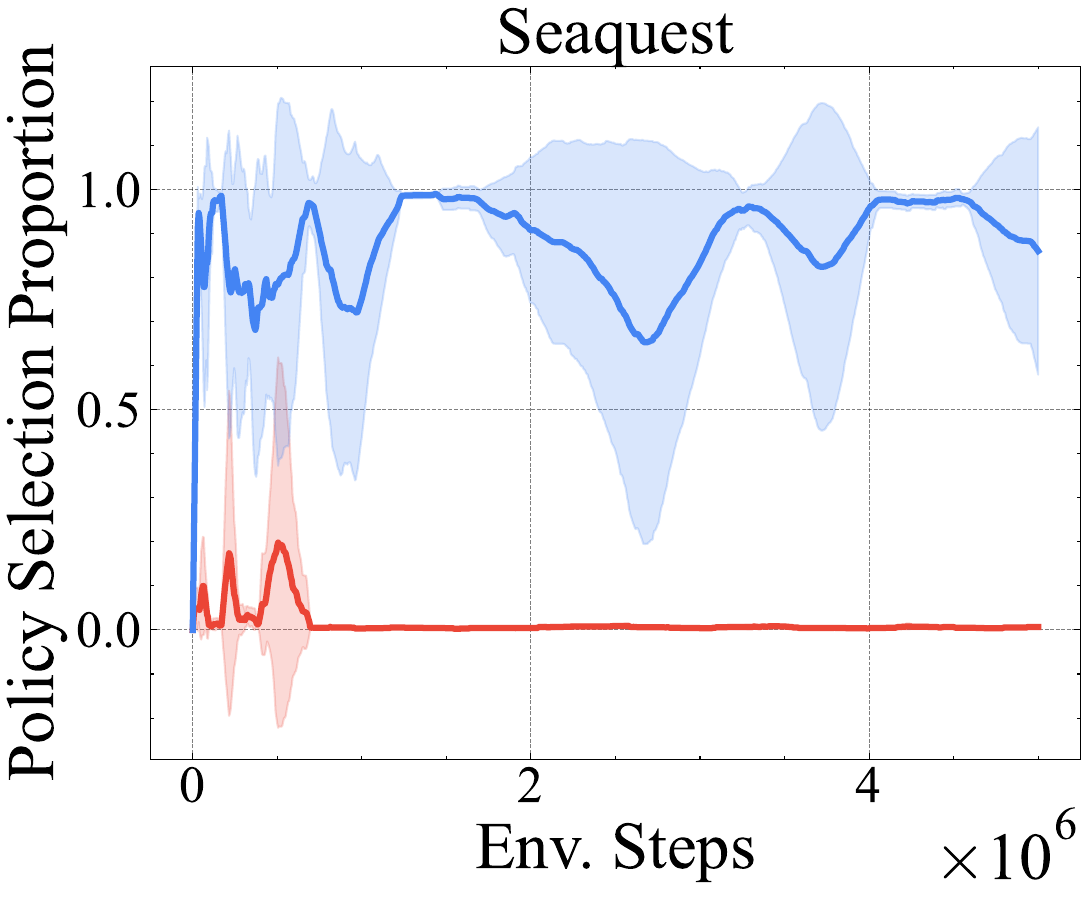}}
    \subfigure{\Description{}\includegraphics[width=0.235\textwidth]{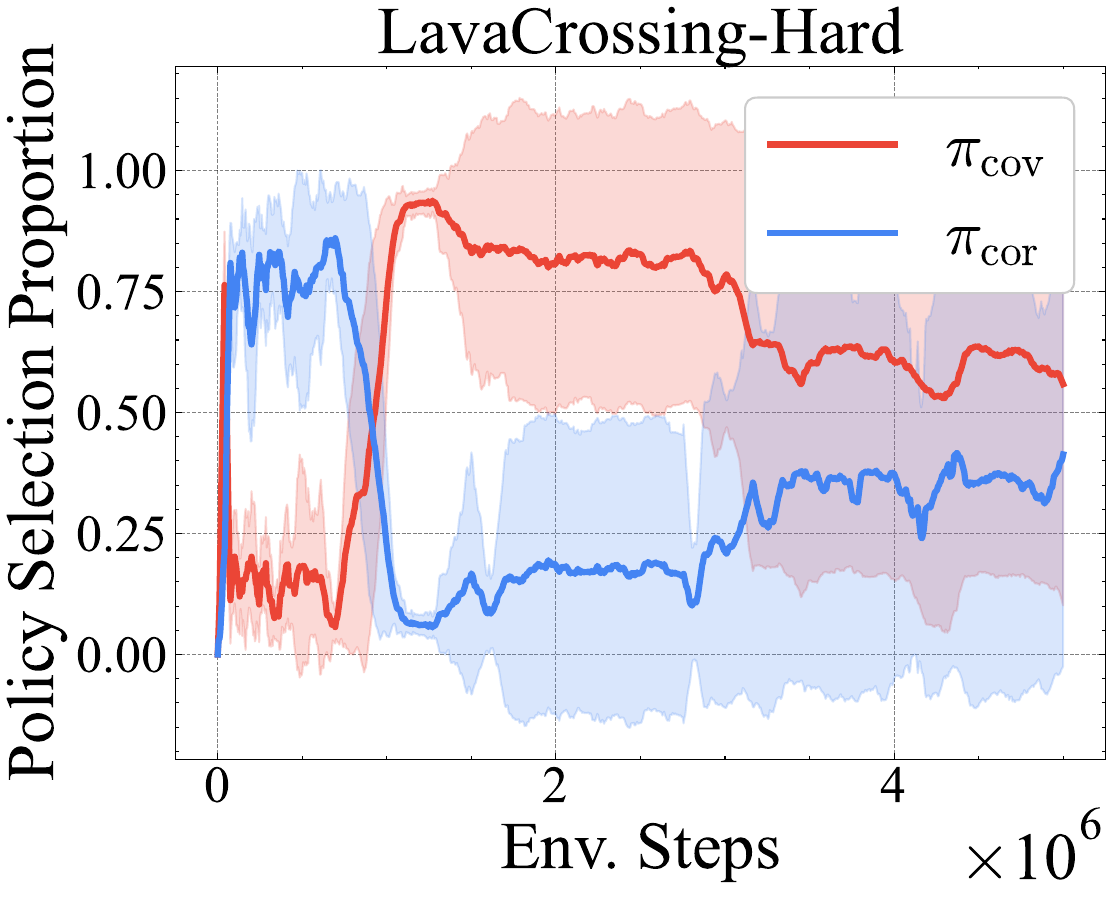}}
    \vskip -0.2in
    \caption{
    Policy selection varies across different tasks.
    In simple (LavaCrossing-Easy) or dense reward (Asterix) tasks, exploration primarily corrects estimation biases. In harder tasks (LavaCrossing-Hard), two types of exploration alternate, leading to a more complex policy selection strategy.
    }
    \vskip -0.15in
    \label{fig:role of two kinds of exploration policies}
\end{center}
\end{figure}

\textbf{Policy Selection During Learning.} For the first question, we illustrate the selection proportions of the two types of polices within the siding window $L$ during the learning process in~\cref{fig:role of two kinds of exploration policies}.
We group $\{\pi_{\text{cov}(0.05)},\pi_{\text{cov}(0.1)}\}$ together as $\pi_{\text{cov}}$ for state-action coverage, and $\{\pi_{\text{cor}(0.1)},$ $\pi_{\text{cor}(0.2)},$ $\cdots,\pi_{\text{cor}(1)} \}$ together as $\pi_{\text{cor}}$ for overestimation bias correction. 
We leave $\pi_{\text{cor}(0)}$ in \cref{appendix:additional analysis}~\cref{fig:Policy selection proportions}, as it is a pure exploitation policy. 

In simple environments such as LavaCrossing-Easy and dense reward environments such as Asterix, exploring for bias correction plays a more significant role. This suggests that exploring some overestimated actions is sufficient to achieve good performance, without the need to focus extensively on discovering hard-to-reach rewards. In contrast, in hard exploration environments such as LavaCrossing-Hard, the two types of policies interleave, resulting in a more intricate selection pattern. This indicates that relying on a single type of policy may not be enough to achieve good performance in hard exploration environments. Novel states require effort to explore, and overestimated state-actions also need to be corrected.

In summary, our method dynamically selects the exploration policy based on the environment. In simple environments, it is usually more beneficial to find low-hanging-fruit rewards rather than spending much effort exploring novel areas. In hard exploration environments, state-novelty exploration plays a more important role in finding new states with high rewards. This policy selection mechanism parallels the principles of depth-first search (DFS) and breadth-first search (BFS). When encountering positive rewards, our approach adopts a depth-first exploration, delving deeper into the discovered areas for further exploration. Conversely, in the absence of immediate rewards, we shift towards a breadth-first strategy, exploring widely in search of promising areas.

\begin{figure}[t]
    \vskip -0.15in
\begin{center}
    \subfigure{\Description{}\includegraphics[width=0.235\textwidth]{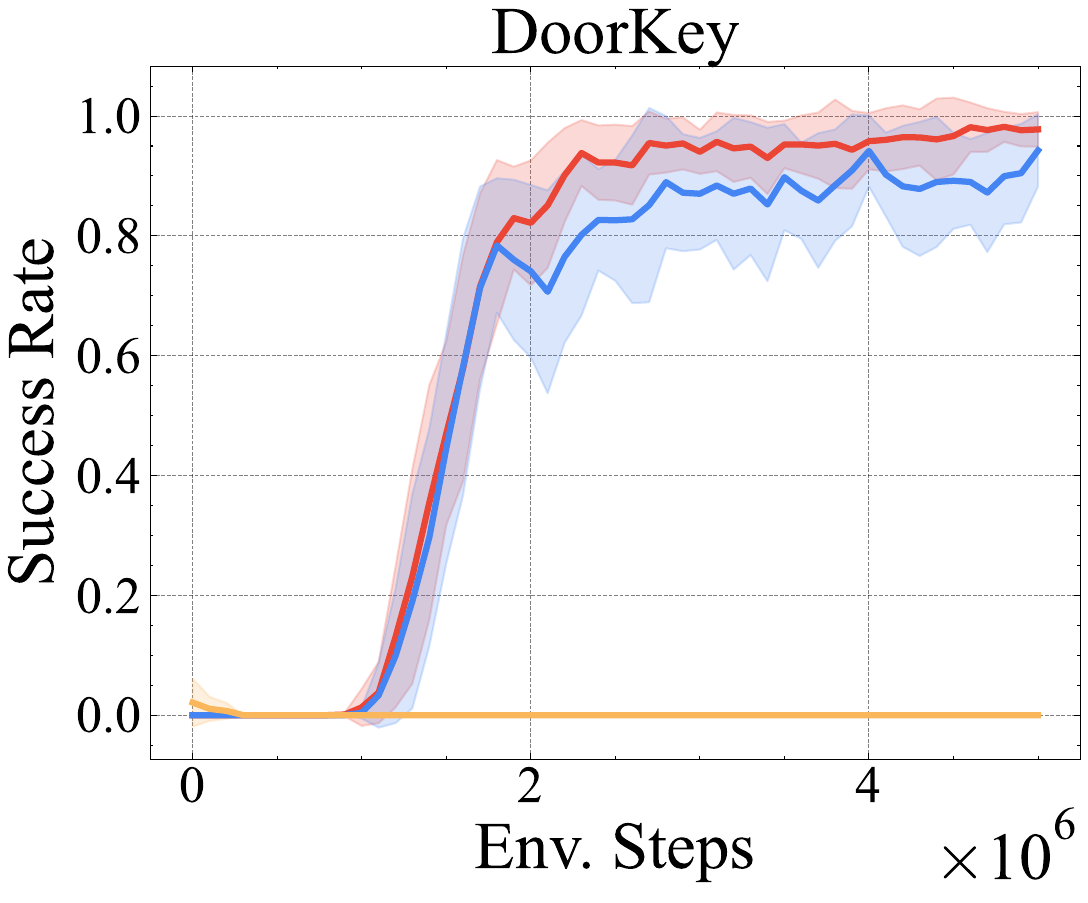}}
    \subfigure{\Description{}\includegraphics[width=0.235\textwidth]{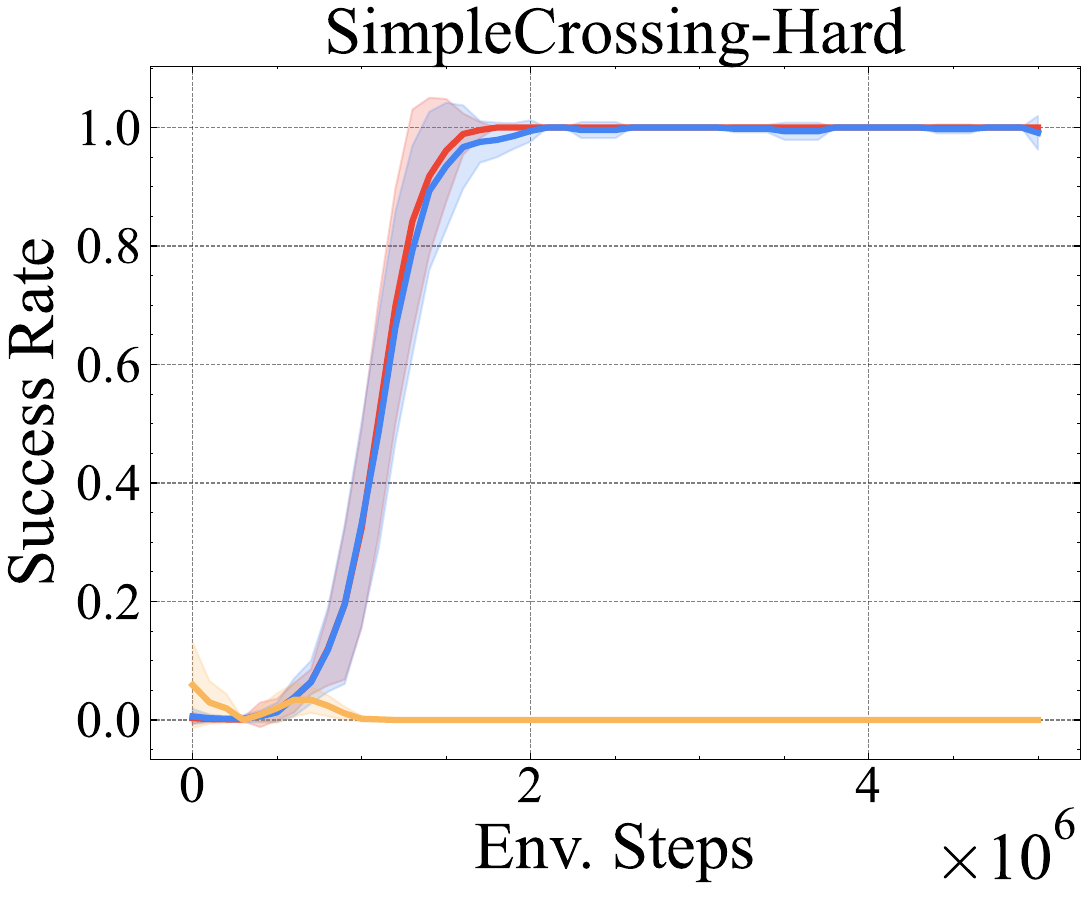}}
    \vskip -0.15in
    \subfigure{\Description{}\includegraphics[width=0.235\textwidth]{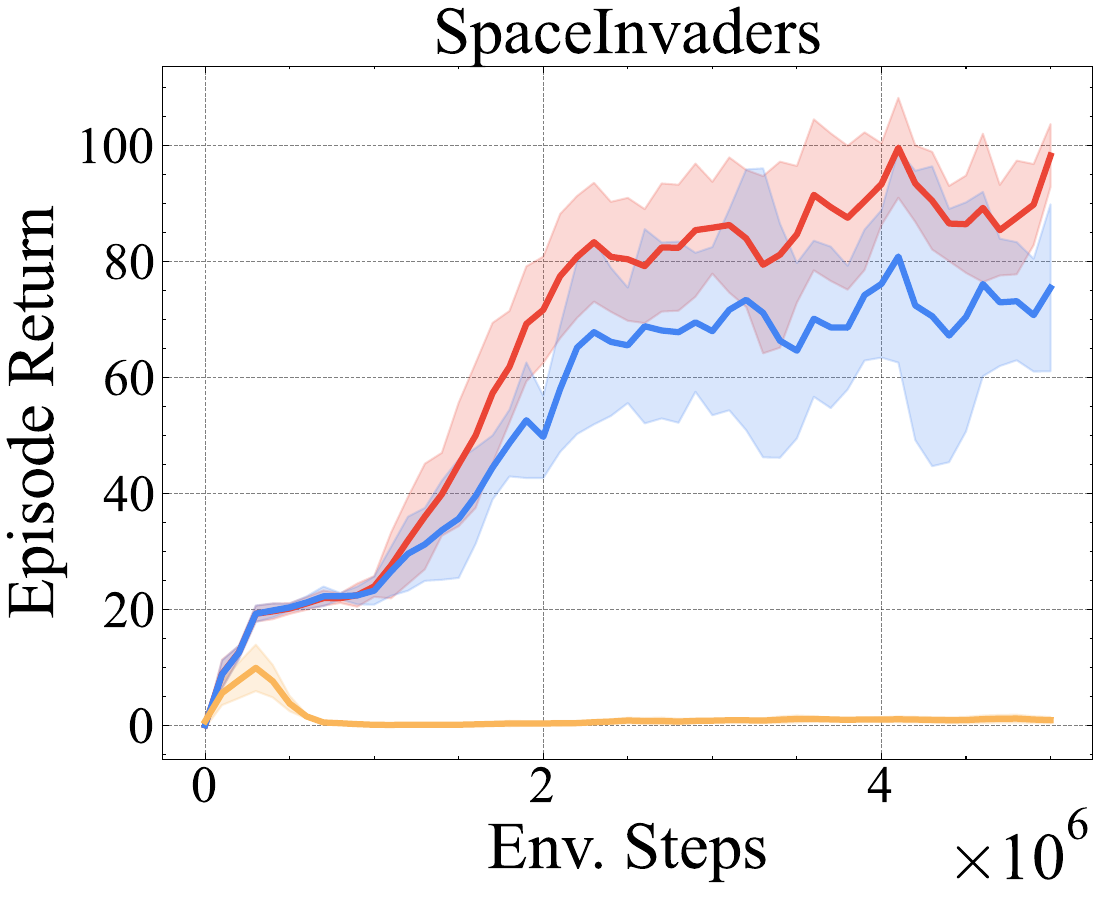}}
    \subfigure{\Description{}\includegraphics[width=0.23\textwidth]{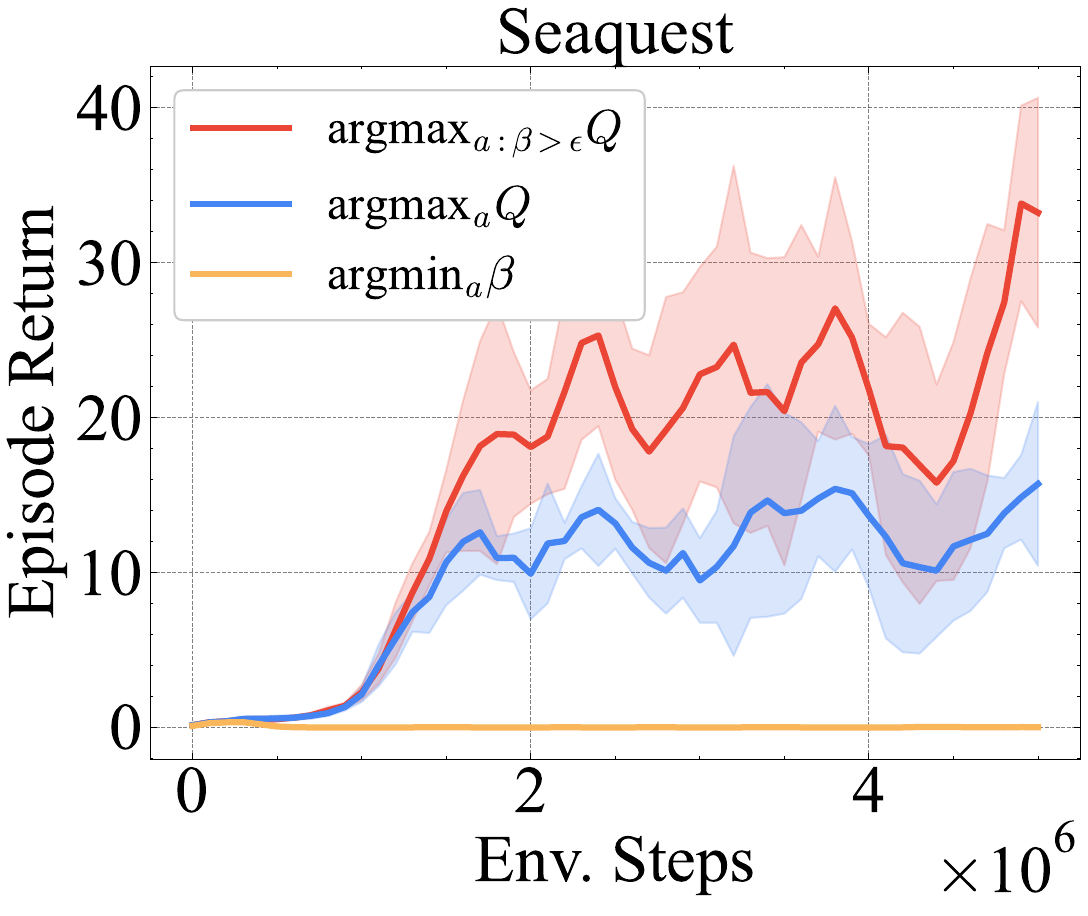}}
    \vskip -0.2in
    \caption{
    The performance of the three basic polices.
    $\arg\min_a \beta$ learns nothing since it does not consider rewards. ${\arg\max}_{a:\beta>\epsilon}Q$ chooses in-sample greedy actions and performs the best. $\arg\max_a Q$ takes greedy actions among the entire action space and may take overestimated actions.
    }
    \vskip -0.15in
    \label{fig:The performance of polices in the policy set}
\end{center}
\end{figure}

\textbf{Polices Performance in the Policy Set.} For the second question, we show the performance of the three basic polices defined in \cref{eq:explore cov,eq:q greedy,eq:q mask greedy}, which form the basis of our policy set.

As shown in~\cref{fig:The performance of polices in the policy set}, the policy $\arg\min_a \beta$ always selects actions with the lowest probability, disregarding performance and thus learning nothing. The policy ${\arg\max}_{a:\beta(a|s)>\epsilon}Q(s,a)$ chooses greedy actions that are well-supported in the replay memory and performs the best. The policy $\arg\max_a Q$ takes greedy actions across the entire action space, which may take overestimated actions, resulting in performance that is not as good as the policy ${\arg\max}_{a:\beta(a|s)>\epsilon}Q(s,a)$. This result aligns with our expectations, highlighting the distinct purposes of the three basic policies. 

In some environments such as SimpleCrossing-Hard, $\arg\max_a Q$ performs similar to ${\arg\max}_{a:\beta(a|s)>\epsilon}Q(s,a)$. This indicates that the space has been fully explored and there is little estimation bias in the $Q$ function. In contrast, in environments such as SpaceInvaders, there is a significant performance gap between $\arg\max_a Q$ and ${\arg\max}_{a:\beta(a|s)>\epsilon}Q(s,a)$. This suggests that many underexplored, overestimated actions still need correction.

% \begin{figure}[htbp]
%     % \vskip -0.1in
% \begin{center}
%     \subfigure[Learning two separate $Q$s.]{
%     \Description{}\includegraphics[width=0.235\textwidth]{ICLR2024/pic/learn2Q/DoorKey_ablation_learn2Q.pdf}
%     \Description{}\includegraphics[width=0.235\textwidth]{ICLR2024/pic/learn2Q/SpaceInvaders_ablation_learn2Q_legend.pdf}
%     \label{fig:Learn individual Q}}
%     \subfigure[The influence of the policy set size.]{
%     \Description{}\includegraphics[width=0.235\textwidth]{
%     ICLR2024/pic/policy_set_size/DoorKey_ablation_policy_set_size.pdf}
%     \Description{}\includegraphics[width=0.235\textwidth]{ICLR2024/pic/policy_set_size/SpaceInvaders_ablation_policy_set_size_legend.pdf}
%     \label{fig:Size of policy set}}
%     \vskip -0.1in
%     \caption{
%     Ablation studies on different configurations. (a) We learn two $Q$s separately and find no obvious difference, which means it is enough to derive two $Q$s from~\cref{eq:in distribution td learning}. (b) We construct policy sets with different sizes. 
%     There is clear performance improvement when we include all the three functions in the policy set.
%     }
%     \vskip -0.2in
% \end{center}
% \end{figure}

% \textbf{Performance Analysis with Different Configurations.}
\subsection{Ablation Studies}
In this section, we study two questions:
(1) Is there a difference when learning the $Q$ function with and without the constraint of $\beta$?
(2) Since we can set different values for $\delta$ and $\alpha$ to construct the policy set, what is the influence of the policy set size?

\begin{figure}[t]
    \vskip -0.15in
\begin{center}
    \subfigure{\Description{}\includegraphics[width=0.235\textwidth]{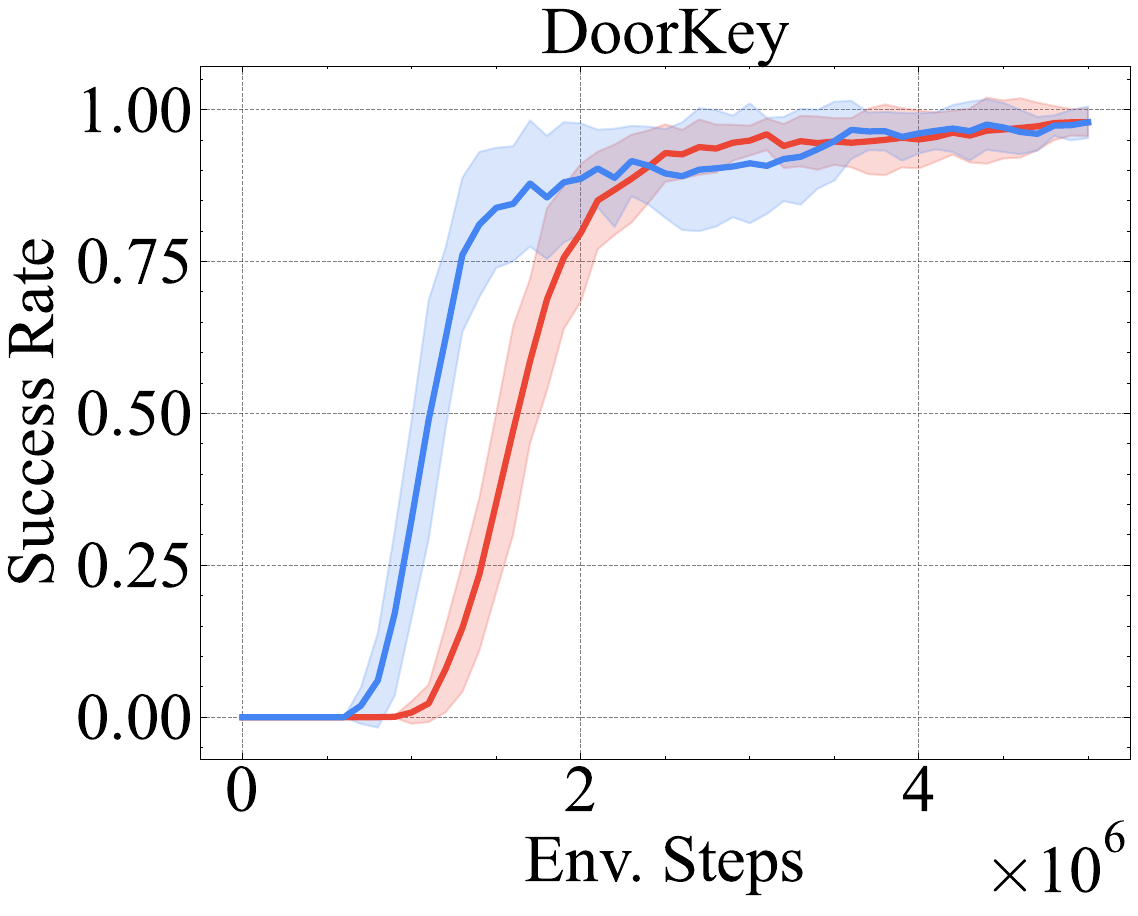}}
    \subfigure{\Description{}\includegraphics[width=0.235\textwidth]{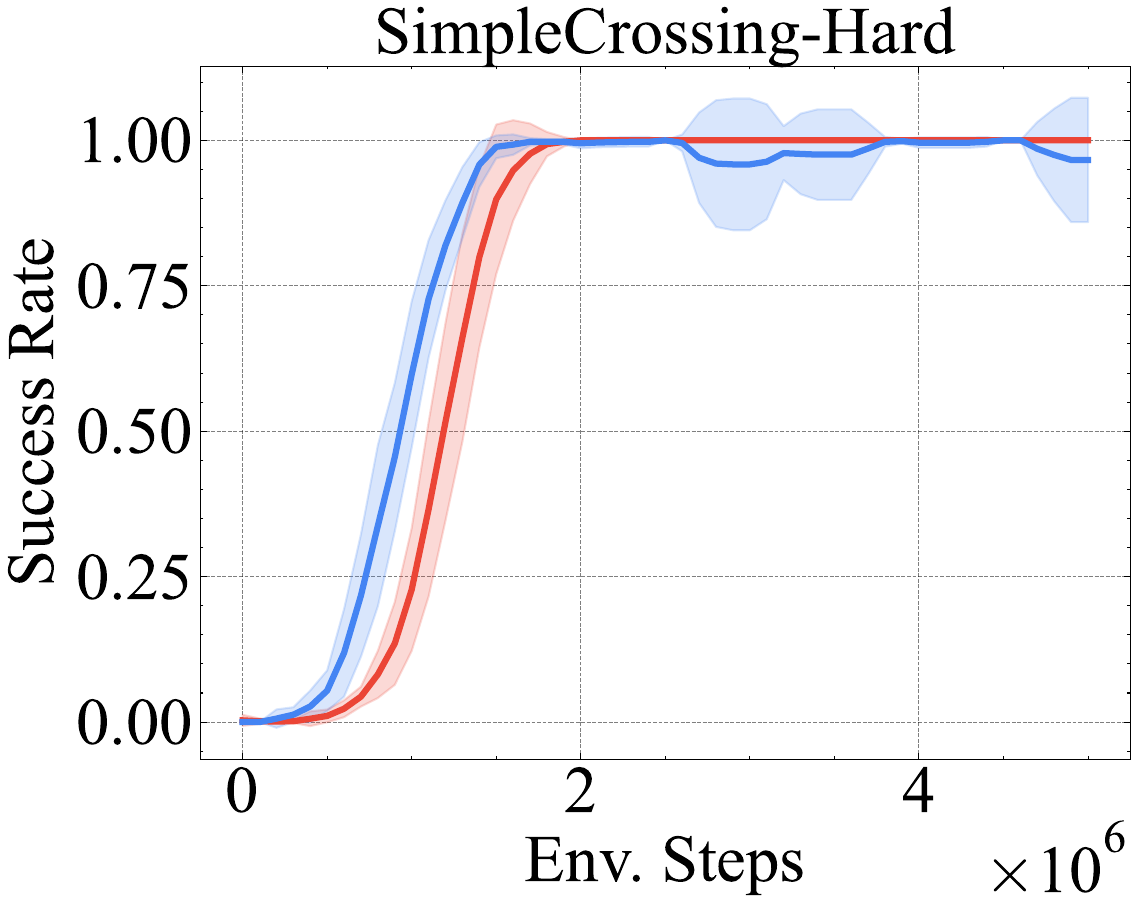}}
    \vskip -0.15in
    \subfigure{\Description{}\includegraphics[width=0.235\textwidth]{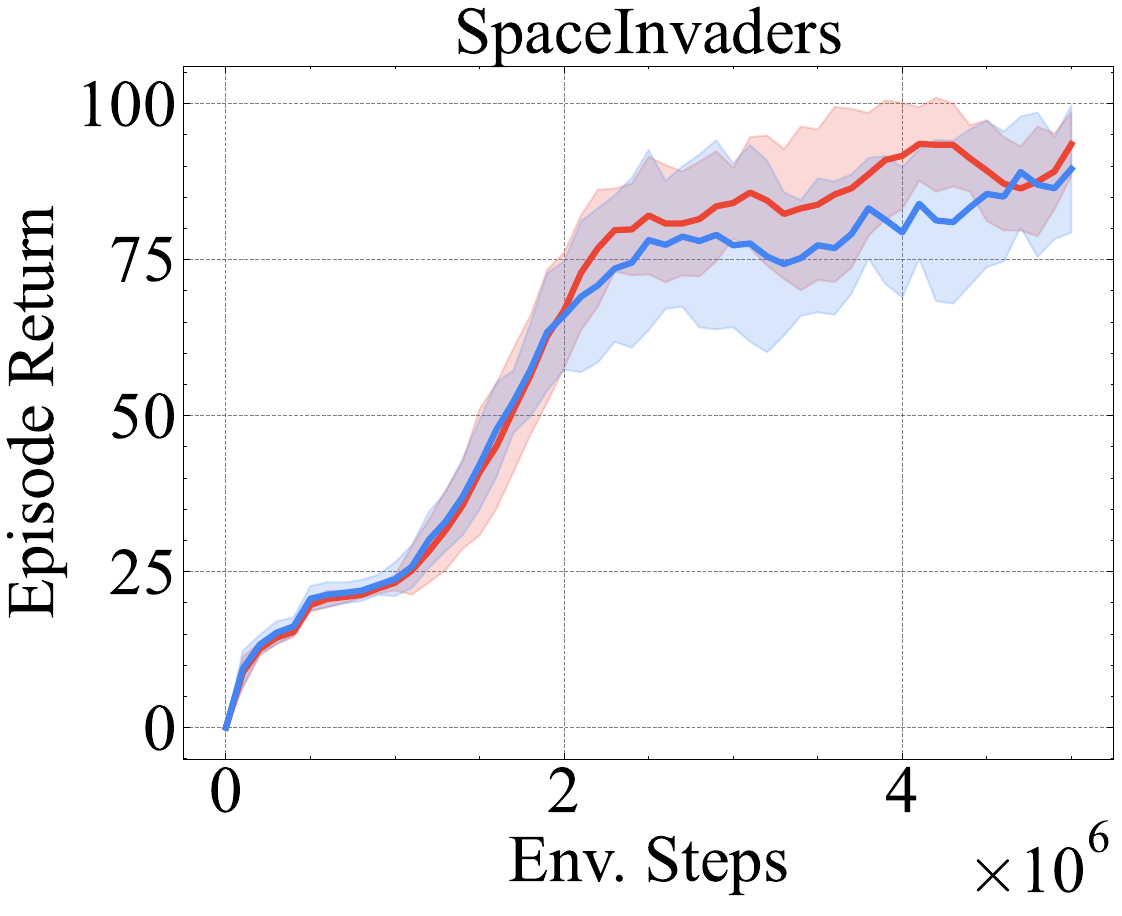}}
    \subfigure{\Description{}\includegraphics[width=0.23\textwidth]{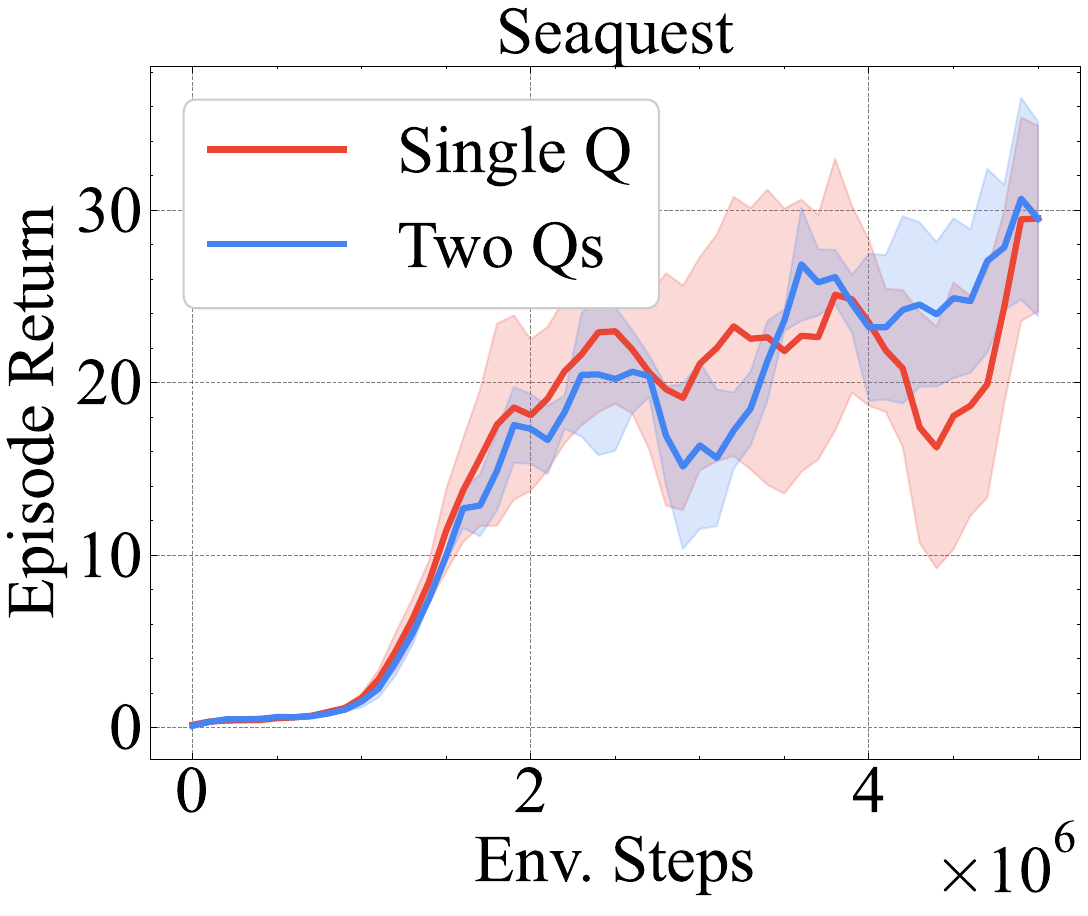}}
    \vskip -0.2in
    \caption{No significant difference is observed between two separate $Q$ functions and a single $Q$ function.
    }
    \vskip -0.15in
    \label{fig:Learn individual Q}
\end{center}
\end{figure}

\textbf{Learning Two Separate $Q$ Functions.} 
In our method, we learn a single $Q$ function using~\cref{eq:in distribution td learning}. Taking the argmax of $Q$ yields an optimistic policy that explores for overestimation bias correction. Masking $Q$ with $\beta$ before taking the argmax provides a pure exploitation policy. The intuition is that while \cref{eq:in distribution td learning} offers a conservative estimate based on in-distribution data, it may still overestimate at unseen state-action pairs. The practical benefit is that learning one $Q$ function is more computationally efficient.
% The practical benefit is that we only need to learn one $Q$ function, which is more computationally efficient.

An alternative approach is to learn two separate $Q$ functions: one for conservative estimation and the other for optimistic estimation. The conservative $Q$ is learned with the constraint of $\beta$ and then masked to obtain the pure exploitation policy. The optimistic $Q$ is learned without the constraint, and the argmax is taken to derive the optimistic policy and try overestimated actions. 

We compare the performance of these two approaches in~\cref{fig:Learn individual Q}.
% In \cref{fig:Learn individual Q}, we compare the performance of learning two separate $Q$ functions versus learning a single $Q$ function. 
We find no significant performance difference across environments, indicating that learning a single $Q$ function is sufficient to achieve both conservative and optimistic estimations while being more computationally efficient.

\begin{figure}[t]
    \vskip -0.15in
\begin{center}
    \subfigure{\Description{}\includegraphics[width=0.235\textwidth]{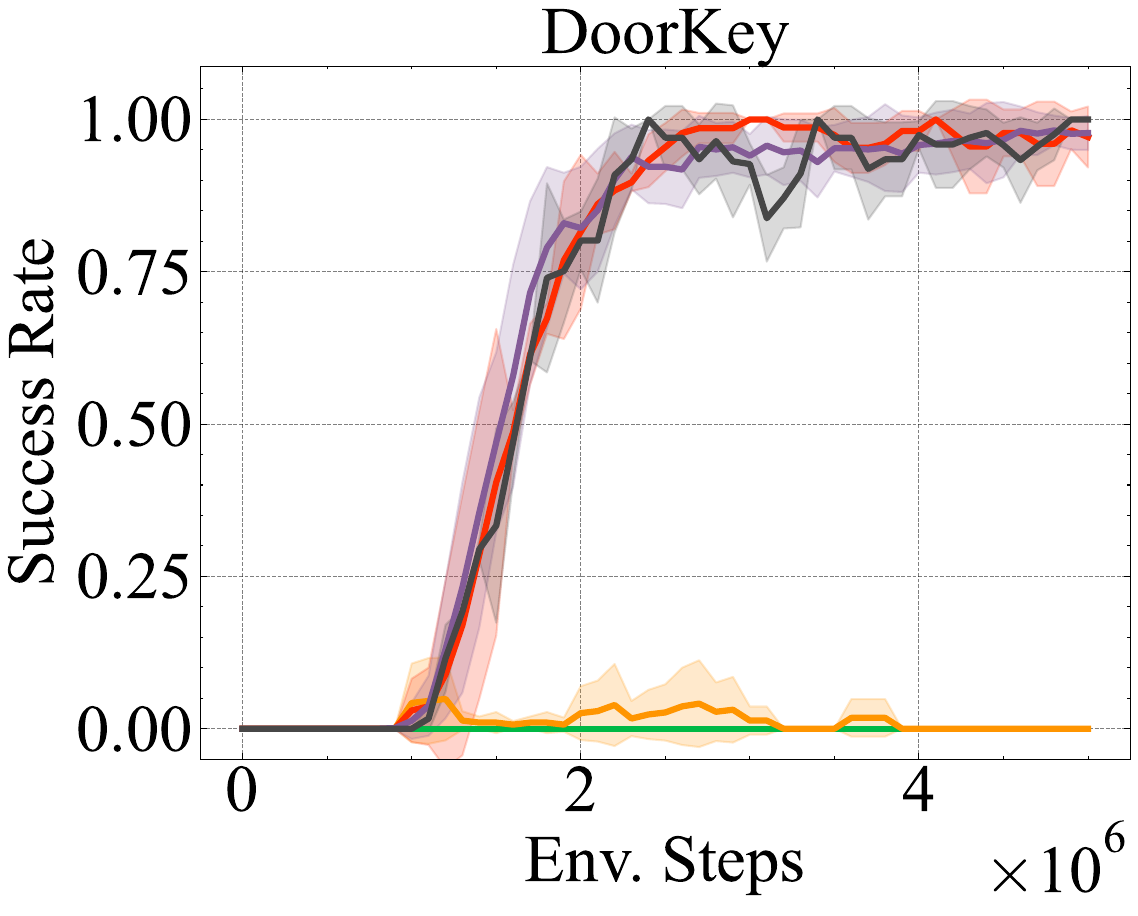}}
    \subfigure{\Description{}\includegraphics[width=0.235\textwidth]{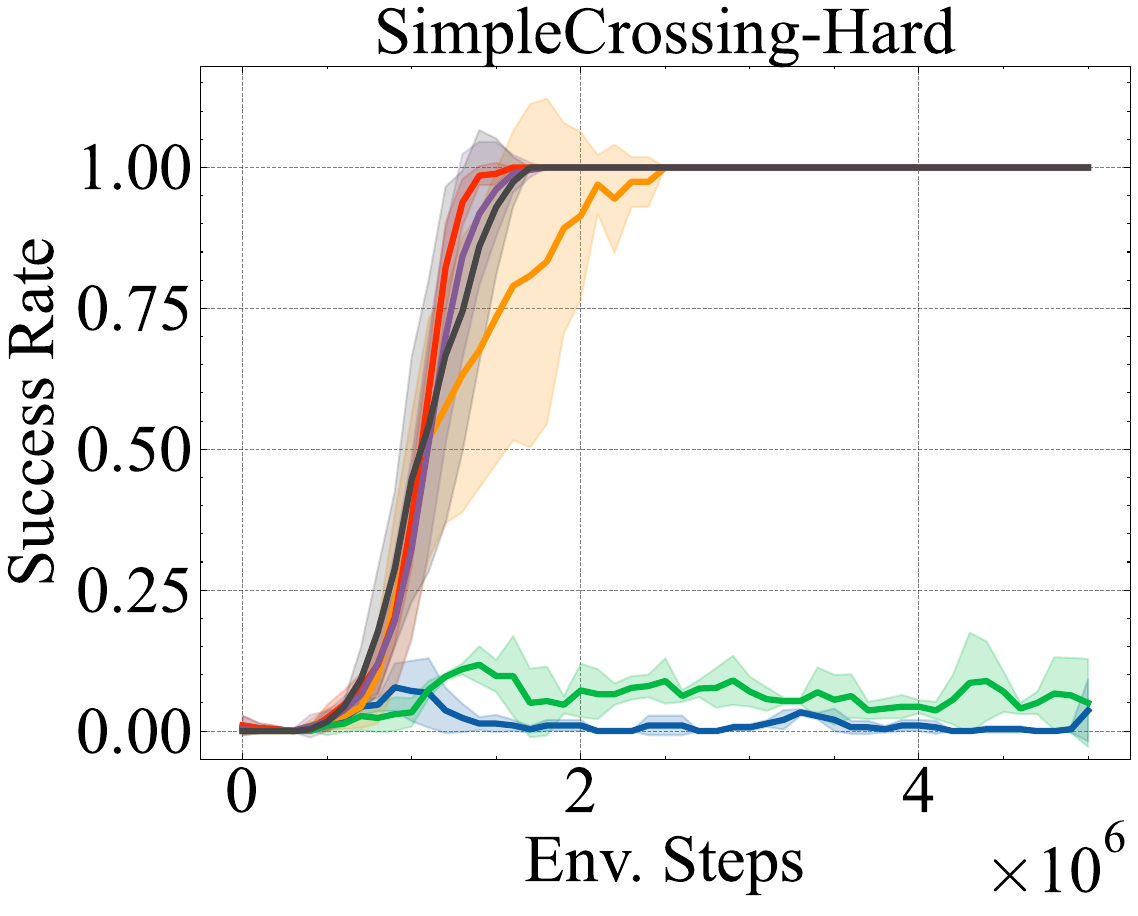}}
    \vskip -0.15in
    \subfigure{\Description{}\includegraphics[width=0.235\textwidth]{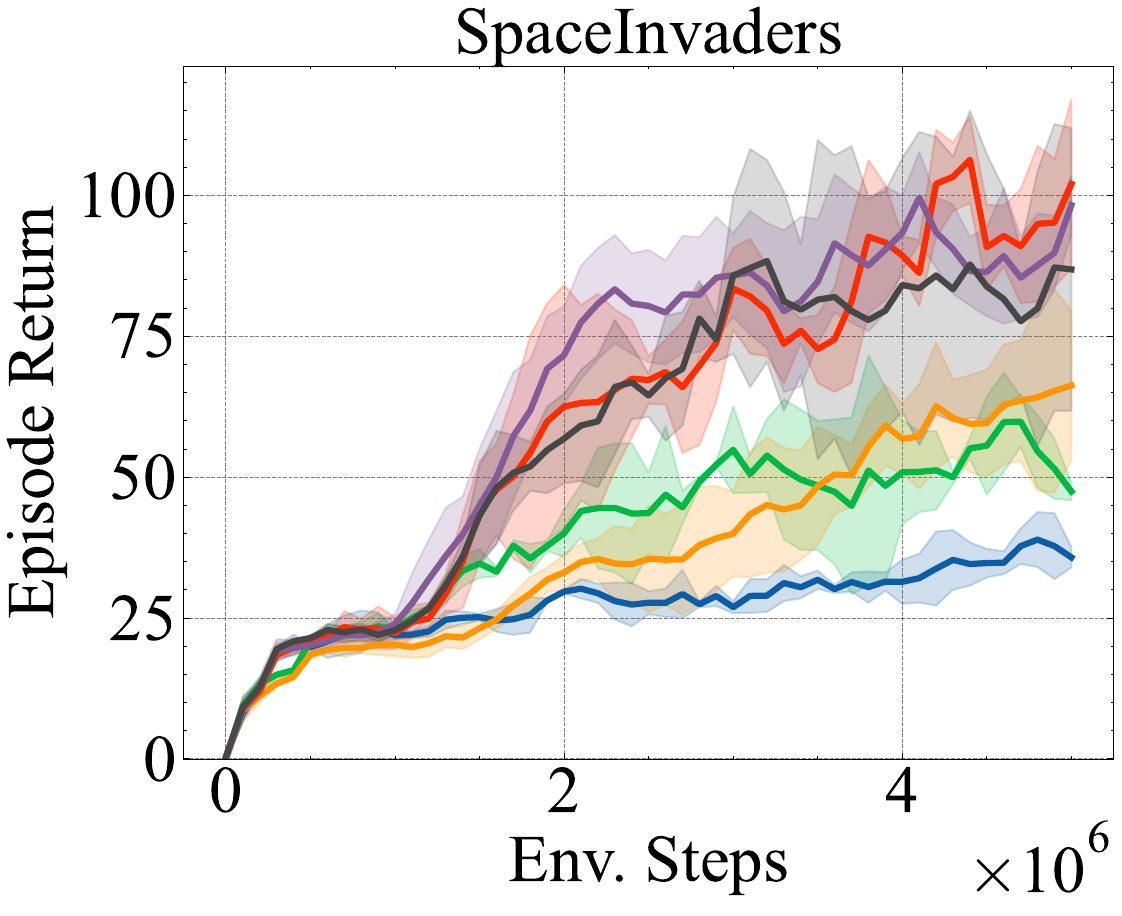}}
    \subfigure{\Description{}\includegraphics[width=0.23\textwidth]{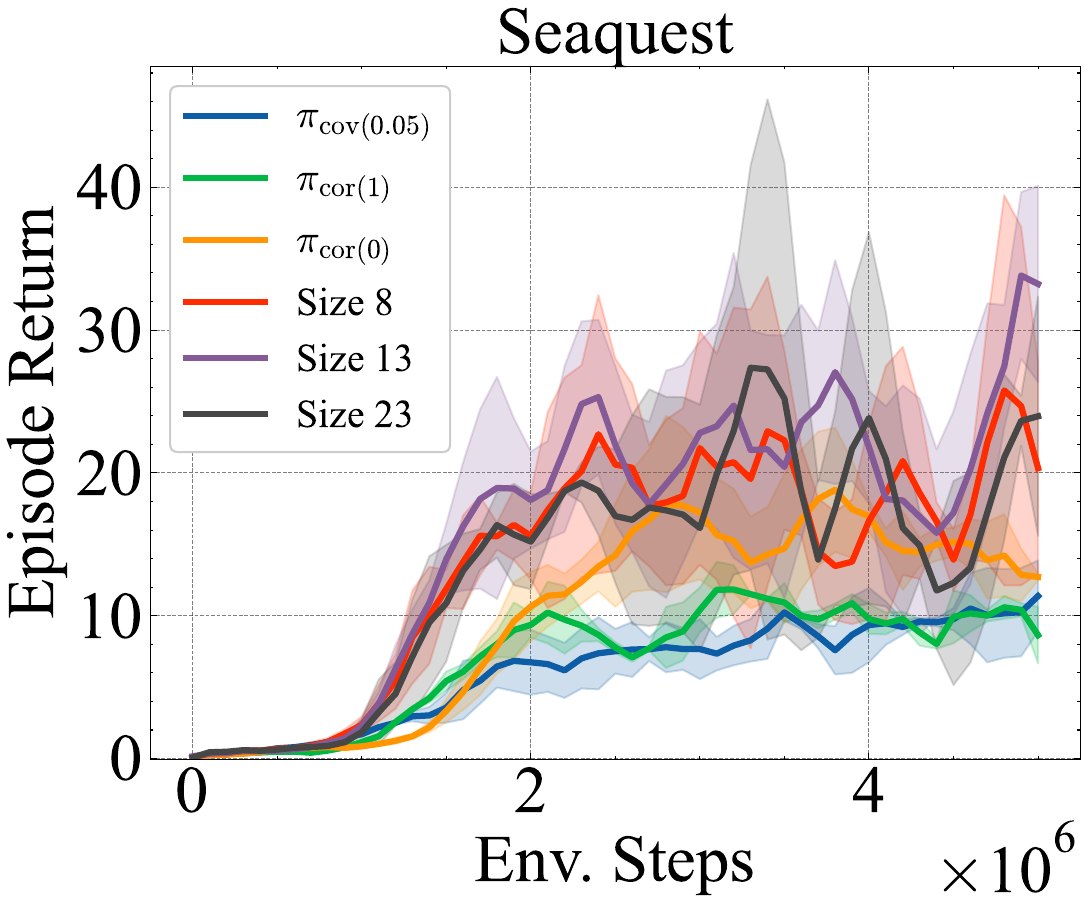}}
    \vskip -0.2in
    \caption{The influence of the policy set size. Performance improves as the policy set size increases, with significant gains observed when all three basic functions are included.}
    \vskip -0.15in
    \label{fig:Size of policy set}
\end{center}
\end{figure}

\textbf{Size of Policy Set.} One benefit of our method is that we can construct policy sets of varying sizes without increasing computational cost. By adding different $\delta$ and $\alpha$, we can create larger policy set. We construct policy sets of different sizes and compare their performance, as shown in~\cref{fig:Size of policy set}.

The policies $\pi_{\text{cov}(0.05)},\pi_{\text{cor}(0)},\pi_{\text{cor}(1)}$ indicate that there is only one policy.
And others show the size of the policy set.
\textbf{Size 8} denotes the policy set $\Pi = \{\pi_{\text{cov}(0.05)},$ $\pi_{\text{cov}(0.1)}, $ $\pi_{\text{cor}(0)},$ $\pi_{\text{cor}(0.2)},$ $\pi_{\text{cor}(0.4)},$$ \cdots,\pi_{\text{cor}(1)} \}$.
\textbf{Size 13} denotes the policy set $\Pi = \{\pi_{\text{cov}(0.05)},$ $\pi_{\text{cov}(0.1)},$ $\pi_{\text{cor}(0)},$ $\pi_{\text{cor}(0.1)},$ $\pi_{\text{cor}(0.2)},$ $\cdots,\pi_{\text{cor}(1)} \}$, which we used in our main results.
\textbf{Size 23} denote the policy set $\Pi = \{\pi_{\text{cov}(0.05)},$ $\pi_{\text{cov}(0.1)},$ $\pi_{\text{cor}(0)},$ $\pi_{\text{cor}(0.05)},$ $\pi_{\text{cor}(0.1)},$ $\cdots,\pi_{\text{cor}(1)} \}$.

We find that $\pi_{\text{cov}(0.05)}$ does not learn effectively in most of environments, indicating that solely focusing on state-action coverage does no benefit learning. 
This may be because novel states do not always correlate with improved rewards~\citep{bellemare2016unifying,simmons2021reward}. Although both $\pi_{\text{cor}(0)}$ and $\pi_{\text{cor}(1)}$ learn something, they perform worse than a larger policy set. This suggests that a single policy is insufficient for achieving good performance due to the lack of diverse exploration. 
In contrast, combining the three basic polices results in significant performance gains with larger policy set sizes, emphasizeing the importance of diverse exploration. When we increase the policy size to 13 and 23, there is no significant difference.
% within 5 million environmental steps. 
This may indicate the diversity is similar in this two policy sets.

\section{Conclusion}
In this paper, we enhance exploration by constructing a group of diverse polices through the additional learning of a behavior function $\beta$ from the replay memory using supervised learning.
With $\beta$, we create a set of exploration policies that range from exploration for state-action coverage to overestimation bias correction.
An adaptive meta-controller is then designed to select the most effective policy for interacting with the environment in each episode. 
Our method is simple, general, and adds minimal computational overhead to DQN. Experiments conducted on MinAtar and MiniGrid demonstrate that our method is effective and broadly applicable in both easy and hard exploration tasks.
Future work could extend our method to environments with continuous action spaces. 
  
%%%%%%%%%%%%%%%%%%%%%%%%%%%%%%%%%%%%%%%%%%%%%%%%%%%%%%%%%%%%%%%%%%%%%%%%

%%% The acknowledgments section is defined using the "acks" environment
%%% (rather than an unnumbered section). The use of this environment 
%%% ensures the proper identification of the section in the article 
%%% metadata as well as the consistent spelling of the heading.

\begin{acks}
Zhang and M\"uller acknowledge support from NSERC, the Natural Sciences and Engineering Research Council of Canada, UAHJIC, the Digital Research Alliance of Canada, and the Canada CIFAR AI Chair program. Xiao acknowledges support from the National Natural Science Foundation of China (62406271).
\end{acks}

%%%%%%%%%%%%%%%%%%%%%%%%%%%%%%%%%%%%%%%%%%%%%%%%%%%%%%%%%%%%%%%%%%%%%%%%

%%% The next two lines define, first, the bibliography style to be 
%%% applied, and, second, the bibliography file to be used.

\bibliographystyle{ACM-Reference-Format}
\balance
\bibliography{sample}

%%%%%%%%%%%%%%%%%%%%%%%%%%%%%%%%%%%%%%%%%%%%%%%%%%%%%%%%%%%%%%%%%%%%%%%%

\clearpage

\appendix

\section{Analysis}
\label{appendix:proof}

\textbf{\cref{prop:in-sample td convergence}.} \ \ \textit{In the tabular case with finite state action space $\mathcal{S}\times \mathcal{A}$, the temporal difference learning masked by $\beta$ given in \cref{eq:in distribution td learning} uniquely converges to the optimal in-sample value $\widehat{Q}^*$ on explored state-action pairs. When $\beta(a|s)>\epsilon$ for all $a \in \mathcal{A}$, $\widehat{Q}^*$ equals to $Q^*$, which recovers the original temporal difference learning without action mask in \cref{eq:bellman}.}
\\

The Bellman (optimality) operator $\mathcal{B}$ for the original temporal difference learning is defined as:
\begin{equation}
\label{eq:bellman operator}
(\mathcal{B}Q)(s,a) = \sum_{s^\prime\in \mathcal{S}}P(s^\prime|s,a)[r+\gamma \max_{a^\prime}Q(s^\prime,a^\prime)].
\end{equation}
Previous works have shown the operator $\mathcal{B}$ is a $\gamma$-contractor with respect to supremum norm if all state-action pairs are available~\citep{szepesvari2010algorithms}:
\begin{equation}
\Vert \mathcal{B}Q_1 - \mathcal{B}Q_2 \Vert_\infty \leq \gamma\Vert Q_1 - Q_2 \Vert_\infty,
\end{equation}
where the supremum norm $\Vert v \Vert_\infty = \max_{1\leq i \leq d} |v_i|$, $d$ is the dimension of vector $v$.
Following Banach's fixed-point theorem \citep{granas2003fixed}, $Q$ converges to optimal action value $Q^*$ if we consecutively apply operator $\mathcal{B}$ to $Q$,  $\lim_{n\rightarrow \infty} (\mathcal{B})^{n} Q = Q^*$.
Further, the update rule in \cref{eq:bellman}, i.e. $Q$-learning, is a sampling version that applies the $\gamma$-contraction operator $\mathcal{B}$ to $Q$. It can be considered as a random process and will converge to $Q^*$, $\lim_{t\rightarrow \infty}Q_t = Q^*$, with some mild conditions \citep{szepesvari2010algorithms,robbins1951stochastic,jaakkola1993convergence,melo2001convergence}.

However in practice, it is hard to access all state-action pairs and the $\gamma$-contraction property may not hold for $\mathcal{B}$. Fortunately, the constrained TD learning in \cref{eq:in distribution td learning} is a $\gamma$-contractor under the supremum norm. Similarly, we define the empirical Bellman (optimality) operator $\mathcal{\hat{B}}$ for \cref{eq:in distribution td learning}:
\begin{equation}
\label{eq:empirical bellman operator}
(\mathcal{\hat{B}}\hat{Q})(s,a) = \sum_{s^\prime\in \mathcal{S}}P(s^\prime|s,a)[r+\gamma \max_{a^\prime:\beta(a^\prime|s^\prime)>\epsilon}\hat{Q}(s^\prime,a^\prime)].
\end{equation}
We can rewrite $\Vert \mathcal{\hat{B}}\hat{Q}_1 - \mathcal{\hat{B}}\hat{Q}_2 \Vert_\infty$ as
\begin{equation*}
\begin{split}
& \quad \Vert \mathcal{\hat{B}}\hat{Q}_1 - \mathcal{\hat{B}}\hat{Q}_2 \Vert_\infty  \\
&= \max_{s,a} \Big\lvert \sum_{s^\prime\in \mathcal{S}}P(s^\prime|s,a)[r + \gamma\max_{a^\prime_1:\beta(a^\prime_1|s^\prime)>\epsilon}\hat{Q}_1(s^\prime,a^\prime_1)] \\
& \qquad \qquad - P(s^\prime|s,a)[r + \gamma\max_{a^\prime_2:\beta(a^\prime_2|s^\prime)>\epsilon}\hat{Q}_2(s^\prime,a^\prime_2)] \Big\rvert \\ 
&= \resizebox{1\hsize}{!}{$\max_{s,a}\gamma \Big\lvert \sum_{s^\prime\in \mathcal{S}}P(s^\prime|s,a)[\max_{a^\prime_1:\beta(a^\prime_1|s^\prime)>\epsilon}\hat{Q}_1(s^\prime,a^\prime_1) - \max_{a^\prime_2:\beta(a^\prime_2|s^\prime)>\epsilon}\hat{Q}_2(s^\prime,a^\prime_2)] \Big\rvert$} \\
&\leq \resizebox{1\hsize}{!}{$\max_{s,a}\gamma \sum_{s^\prime\in \mathcal{S}}P(s^\prime|s,a) \Big\lvert \max_{a^\prime_1:\beta(a^\prime_1|s^\prime)>\epsilon}\hat{Q}_1(s^\prime,a^\prime_1) - \max_{a^\prime_2:\beta(a^\prime_2|s^\prime)>\epsilon}\hat{Q}_2(s^\prime,a^\prime_2) \Big\rvert $} \\
&\leq \max_{s,a}\gamma \sum_{s^\prime\in \mathcal{S}}P(s^\prime|s,a) \max_{\Tilde{a}:\beta(\Tilde{a}|s^\prime)>\epsilon} \Big\lvert \hat{Q}_1(s^\prime,\Tilde{a})-\hat{Q}_2(s^\prime,\Tilde{a})\Big\rvert \\
&\leq \max_{s,a}\gamma \sum_{s^\prime\in \mathcal{S}}P(s^\prime|s,a) \max_{\Tilde{s},\Tilde{a}:\beta(\Tilde{a}|\Tilde{s})>\epsilon} \Big\lvert \hat{Q}_1(\Tilde{s},\Tilde{a})-\hat{Q}_2(\Tilde{s},\Tilde{a})\Big\rvert \\
&= \max_{s,a}\gamma \sum_{s^\prime\in \mathcal{S}}P(s^\prime|s,a) \Vert \hat{Q}_1 - \hat{Q}_2 \Vert_\infty \\
&= \gamma \Vert \hat{Q}_1 - \hat{Q}_2 \Vert_\infty,
\end{split}
\end{equation*}
where the last line follows from $\sum_{s^\prime\in \mathcal{S}}P(s^\prime|s,a)=1$.
Further, when we explore by taking actions with probabilities lower than $\epsilon$ and $\beta(a|s)\ge \epsilon$, the mask in \cref{eq:in distribution td learning} takes no effect and we naturally have $\widehat{Q}^* = Q^*$.

\textbf{\cref{prop:behavior coverage}.} \ \ \textit{In the tabular case with finite state action space $\mathcal{S}\times \mathcal{A}$ and finite horizon $H$, taking actions following policy $\pi_{\text{cov}(\delta)}$ guarantees infinite state-action visitations for all state action pairs. However, the expected cumulative regret is linear.}
\\

\begin{figure}[htbp]
    \vskip -0.15in
\begin{center}
    \subfigure{\Description{}\includegraphics[width=0.1\textwidth]{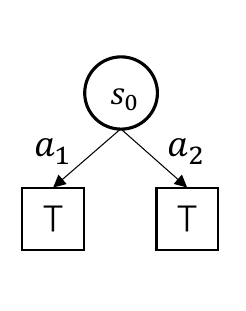}} \qquad \qquad
    \subfigure{\Description{}\includegraphics[width=0.2\textwidth]{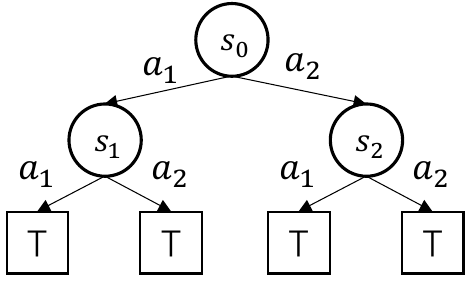}}
    \vskip -0.15in
    \caption{A sample example of multi-armed bandit (left) and tree structured MDP (right) with two actions. $s$ denotes a state, $a$ denotes an action, T means an episode terminates.}
    \vskip -0.1in
    \label{fig:example bandit and tree}
\end{center}
\end{figure}

We first consider the bandit case. The state space $\mathcal{S}=\{s_0\}$ contains only one state. The action space is finite with $\mathcal{A}=\{a_1,\cdots,a_k\}$ and $|\mathcal{A}|=k$. At each time step $t$, an action $A_t=a_i$ is taken and yield a reward $X_t$ from a certain unknown distribution with mean $\mu_{a_i}$. $\mu^*$ denotes the mean value of the optimal action. \cref{fig:example bandit and tree} left shows a case with two actions.

Let $N_n(s_0, a_i)$ be the number of times that $(s_0,a_i)$ is visited up to time $n$:
\begin{equation}
    N_n(s_0,a_i):= \sum_{t=1}^{n}\mathbb{I}(A_t=a_i).
\end{equation}
The sample mean of the rewards for action $a_i$ up to time $n$ is given by:
\begin{equation}
    \hat{\mu}_{a_i}:=\frac{1}{N_n(s_0,a_i)}\sum_{t=1}^{n}X_t\mathbb{I}(A_t=a_i).
\end{equation}
Let $\Delta_{a_i}:=\mu^*-\mu_{a_i}$ be the sub-optimality gap of action $a_i$.
The regret is defined as $R_n:=\sum_{i=1}^k\mathbb{E}[N_n(s_0,a_i)]\Delta_{a_i}$.

In this case, consider taking action following policy $\pi_{\text{cov}(\delta)}$, the algorithm can be write as in~\cref{alg:policy cov}:

\begin{algorithm}[b]
    \caption{Taking actions following policy $\pi_{\text{cov}(\delta)}$}
    \label{alg:policy cov}
  \begin{algorithmic}[1]
    \STATE Initialize $\delta \in [0,1]$, $N_n(s_0,a_i)=1$ and compute $\beta(s_0,a_i)$ for $a_i \in \mathcal{A}$.
    \FOR{$t$ = 0 {\bfseries to} $n$}
    \IF{$\beta(s_0,a_i)>\delta$ for all $a_i \in \mathcal{A}$}
    \STATE Select action $A_t=\arg\max_{a_i \in \mathcal{A}} \hat{\mu}_{a_i}$
    \ELSE
    \STATE Sample an action $A_t \sim \text{Unif}\{a_i : \beta(a_i|s_0) \le \delta \}$
    \ENDIF
    \STATE Get reward $X_t$, update $N_n(s_0,a_i), \ \hat{\mu}_{a_i}$ and recompute $\beta$
    \ENDFOR
\end{algorithmic}
\end{algorithm}
Here, $N_n(s_0,a_i)$ is initialized as 1 to avoid division by zero and $\beta$ is thus a uniform policy at the beginning. It can be initialized as other values which does not affect the conclusion. And the tie in $A_t=\arg\max_{a_i \in \mathcal{A}} \hat{\mu}_{a_i}$ is randomly broken.

Since the policy $\pi_{\text{cov}(\delta)}$ firstly takes actions below probability $\delta$, it means each action will be selected at least $\delta t$ times in expectation, i.e. $\mathbb{E}[N_n(s_0,a_i)] \ge \delta t$. 
It means each action will be selected infinite times when take the limit of $t$.
Without loss of generality, let's assume that $a_1$ is the optimal action, then the regret is $R_n=\sum_{i=1}^k\mathbb{E}[N_n(a_i)]\Delta_{a_i}=\sum_{i=2}^k\mathbb{E}[N_n(a_i)]\Delta_{a_i}\ge \delta t \sum_{i=2}^k \Delta_{a_i}$. Because $\sum_{i=2}^k \Delta_{a_i} >0$, the regret is linear.

Then we consider the RL case with finite horizon $H$. In this case, the state space $\mathcal{S}$ is finite. The action space is the same as the bandit case with $\mathcal{A}=\{a_1,\cdots,a_k\}$ and $|\mathcal{A}|=k$.
We consider the transition as a tree structure and define a new action space as the sequence of actions until terminate: $\widetilde{\mathcal{A}} := \mathcal{A} \times \cdots \times \mathcal{A}$, and $|\widetilde{\mathcal{A}}| = |\mathcal{A}|^H$.
\cref{fig:example bandit and tree} right shows a case with two actions and horizon $H=2$.

The reward $X_{\Tilde{t}}$ denotes the accumulated rewards for one episode.
We measure the visitation and regret after each episode $\Tilde{t}$. Then, for episode $\Tilde{t}$, for state-action pair $(s_k,a_i)$ that leads to a leaf node, the expected visitations would be $\mathbb{E}[N_n(s_k,a_i)] \ge \delta^H \Tilde{t}$. 
% where $\Tilde{t}=tH$. 
And for state-action pair $(s_k,a_i)$ at intermediate level $1\le l \le H$, we have $\mathbb{E}[N_n(s_k,a_i)] \ge \delta^l \Tilde{t} \ge \delta^H \Tilde{t}$, which indicates infinite visits in both cases.
And similarly, let $\Delta_{\Tilde{a}_i}$ be the sub-optimality gap of action sequence $\Tilde{a}_i$, we have episode regret $R_n=\sum_{i=1}^{k^H}\mathbb{E}[N_n(s_k,a_i)]\Delta_{\Tilde{a}_i} \ge \delta^H \Tilde{t} \sum_{i=1}^{k^H} \Delta_{\Tilde{a}_i}$, which is linear.

\section{Implementation Details}
\label{appendix:implementation details}

\textbf{Hyper-parameters.} All methods are based on DQN. We maintain most parameters the same as DQN and reduce the interaction steps to run more different random seeds. We run each experiment with 5 million steps of interaction with the environment. 
We proportionally reduce other parameters based on the interaction steps.
The $\epsilon$-greedy exploration is linearly decayed from 1 to 0.01 in 1 million steps. The target network is updated every 1000 steps. 
The replay memory size is set as 100,000.
The minibatch size is 32. The replay ratio is 0.25~\citep{fedus2020revisiting}, that is to say the $Q$ function is updated once per four environmental steps. 
The optimizer for the network is Adam.
The discount factor is 0.99.
Table~\ref{appendix:DQN_hyperparameter} shows the detailed hyper-parameters that used for all methods.

Besides the common parameters, there are other parameters that are specific to different methods. 
For bootstrapped DQN, we follow the parameters setting in the original paper~\citep{osband2016deep}. 
We split $K=10$ separate bootstrap heads after the convolutional layer. 
And the gradients are normalized by $1/K$.
The parameter $p$ in Bernoulli mask $\omega_1,\cdots,\omega_K \sim \text{Ber}(p)$ is set as $1$ to save on minibatch passes.
When evaluate the performance, we combine all the heads into a single ensemble policy by choosing the action with the most votes across heads.
For $\epsilon z$-greedy, to decide the duration of random actions, we use a heavy-tailed distribution zeta distribution ($z(n)\propto n^{-\mu}$) with $\mu = 2$.
For RND, we the intrinsic reward scale $\alpha$ is set as 10.
In LESSON, the temperature parameter $\tau=0.02$. The Intrinsic reward coefficient $\alpha$ is set as default value in \citep{kim2023lesson}.
For our method $\beta$-DQN, we use $\beta(s,a)>\epsilon$ as a constraint for the max operator to bootstraps from actions in~\cref{eq:in distribution td learning}, we use fixed value $\epsilon=0.05$.
We also fix the policy set $\Pi = \{\pi_{\text{cov}(0.05)},\pi_{\text{cov}(0.1)}, \pi_{\text{cor}(0)},\pi_{\text{cor}(0.1)},\pi_{\text{cor}(0.2)},\cdots,\pi_{\text{cor}(1)} \}$, and sliding-window length $L=1000$. We count the same states visited in an episode and avoid visiting the same state-action too much immediately, which is an augment of the behavior function. We search the learning rates for all methods among \{3e-3,1e-3,3e-5\} and report the best performance.
For these baselines, we implement DQN, Bootstrapped DQN, and $\epsilon z$-greedy according to the original papers and refer some awesome public codebases like RLzoo~\citep{ding2021efficient}, Tianshou~\citep{tianshou}
% \footnote{\url{https://github.com/thu-ml/tianshou}} 
and Clearnrl~\citep{huang2022cleanrl}.
% \footnote{\url{https://github.com/vwxyzjn/cleanrl}} 
RND and LESSON are based on publicly released code\footnote{\url{https://github.com/beanie00/LESSON}}.

\begin{table}[t]
\caption{Hyper-parameters of DQN on MiniGrid and MinAtar environments.}
\label{appendix:DQN_hyperparameter}
\centering
\begin{tabular}{cc}\toprule[2pt]
\specialrule{0pt}{1pt}{1pt}
	Hyperparameter & Value  \\ \midrule
	% \hline\specialrule{0pt}{1pt}{1pt}
    Minibatch size & 32  \\\specialrule{0pt}{1pt}{1pt}
    Replay memory size & 100,000  \\\specialrule{0pt}{1pt}{1pt}
    Target network update frequency  & 1,000 \\\specialrule{0pt}{1pt}{1pt}
    Replay ratio & 0.25 \\\specialrule{0pt}{1pt}{1pt}
    Discount factor  & 0.99 \\\specialrule{0pt}{1pt}{1pt}
    Optimizer  & Adam \\\specialrule{0pt}{1pt}{1pt}
    Initial exploration  & 1 \\\specialrule{0pt}{1pt}{1pt}
    Final exploration  & 0.01 \\\specialrule{0pt}{1pt}{1pt}
    Exploration decay steps  & 1M \\\specialrule{0pt}{1pt}{1pt}
    Total steps in environment & 2M \\\specialrule{0pt}{1pt}{1pt}
	 \specialrule{0pt}{1pt}{1pt}\bottomrule[2pt]
\end{tabular}
\end{table}

\textbf{Network Architecture.} 
We use the same network architecture for all algorithms as used in MinAtar baselines~\citep{young19minatar}. 
% The network contains a convolutional layer with 16 $3\times 3$ convolutions with stride 1, and a fully connected hidden layer with 128 units. 
% The architecture is the same as used in  MinAtar baselines.
It consists of a convolutional layer, followed by a fully connected layer. The convolutional layer has 16 $3\times 3$ convolutions with stride 1, the fully connected layer has 128 units. These settings are one quarter of the final convolutional layer and fully connected layer of the network used in DQN~\citep{mnih2015human}.

For bootstrapped DQN, We split the network of the final layer into $K = 10$ distinct heads, each one is a fully connected layer with 128 units.
RND~\citep{burda2018exploration} involves two more neural networks. One is a fixed randomly initialized neural network which takes an observation to an embedding, and a predictor network trained to predict the embedding output by the fixed randomly initialized neural network.
For LESSON~\citep{kim2023lesson}, it involves the prediction networks the same as RND. And the prediction-error maximizing (PEM) intra-policy contains a separate Q-function, which estimates the expected sum of prediction-error intrinsic rewards. Besides, it learns an option selection policy $\{\pi_{\Omega}\}$ and the terminal functions $\{\beta_{\omega}\}$, thus has more networks to learn.

\textbf{Evaluation.} We run each method on each environment with 10 different random seeds, and show the mean score and standard error with solid line and shaded area in~\cref{fig:performance}.
The performance is evaluated by running 30 episodes after every 100K environmental steps.
We use $\epsilon$-greedy exploration at evaluation with $\epsilon=0.01$ to prevent the agent from being stuck at the same state.

\section{Environment Details}
\label{appendix:environment details}

\subsection{Cliffworld}
\begin{figure*}[htbp]
% \vskip 0.1in
\begin{center}
    \vspace{-0.1in}
    \subfigure{
    \Description{}\includegraphics[width=0.3\textwidth]{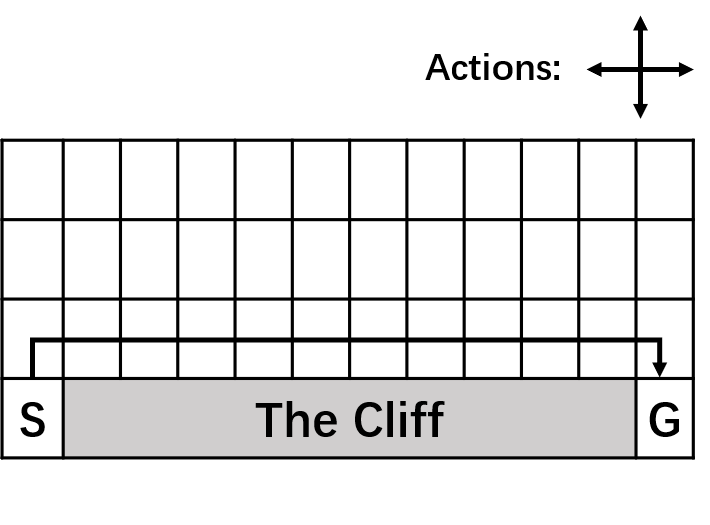}}
    \vskip -0.1in
    \caption{ 
    % A toy example on Cliffworld environment. 
    The illustration of Cliffworld environment. Each grid denotes a state, the black line shows the optimal path from start state S to goal state G.}
    \vskip -0.2in
    \label{fig:cliffworld}
\end{center}
\end{figure*}

Cliffworld is a simple navigation task introduced by \citet{sutton2018reinforcement} as shown in Figure~\ref{fig:cliffworld}. There are 48 states in total which is presented as two-dimensional coordinate axes $x$ and $y$. The size of action space is 4, with left, right, up and down.
The agent needs to reach the goal state G at the bottom right starting from the start state S at the bottom left. 
The reward of reaching the goal is +1, dropped into the cliff gives -1, otherwise is 0. 
We set the discount factor as 0.9 and the max episode steps as 100.
The black line on the figure shows the optimal path. 

\subsection{MiniGrid}
\begin{figure*}[htbp]
    \vskip 0.1in
\begin{center}
    \subfigure[DoorKey]{\Description{}\includegraphics[width=0.2\textwidth]{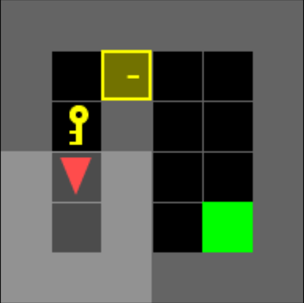}} % 0.2
    \subfigure[Unlock]{\Description{}\includegraphics[width=0.365\textwidth]{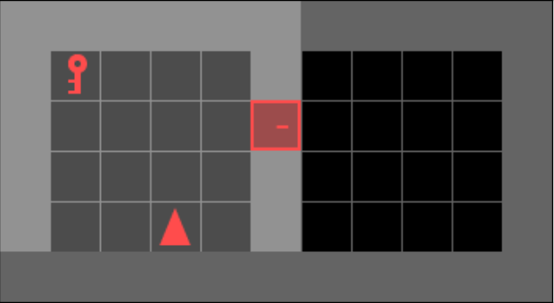}} % 0.365
    \subfigure[SimpleCross-Easy]{\Description{}\includegraphics[width=0.2\textwidth]{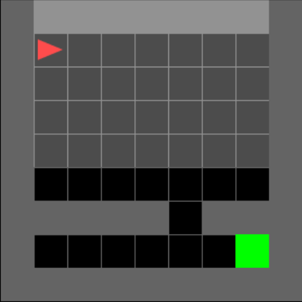}} \\
    \subfigure[SimpleCross-Hard]{\Description{}\includegraphics[width=0.2\textwidth]{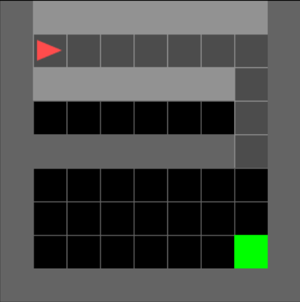}}
    \subfigure[LavaCrossing-Easy]{\Description{}\includegraphics[width=0.2\textwidth]{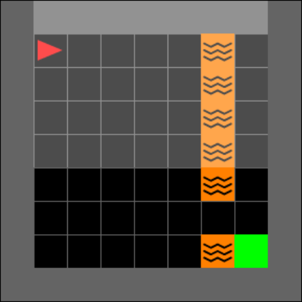}}
    \subfigure[LavaCrossing-Hard]{\Description{}\includegraphics[width=0.2\textwidth]{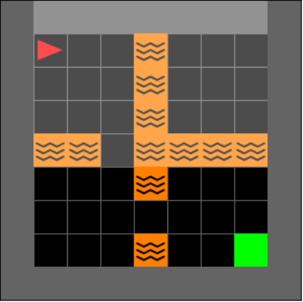}}
    % \vskip -0.1in
    \caption{Visualization of MiniGrid environments.}
    % \vskip -0.2in
    \label{fig:MiniGrid visualization}
\end{center}
\end{figure*}

MiniGrid~\citep{gym_minigrid,MinigridMiniworld23}~\footnote{\url{https://github.com/Farama-Foundation/Minigrid}} is a gridworld Gymnasium~\citep{brockman2016openai} environment, which is designed to be particularly simple, lightweight and fast. 
It implements many tasks in the gridworld environment and most of the games are designed with sparse rewards. 
We choose seven different tasks as shown in Figure~\ref{fig:MiniGrid visualization}. 

The map for each task is randomly generated at each episode to avoid overfitting to a fixed map.
The state is an array with the same size of the map.
The red triangle denotes the player, and other objects are denoted with different symbols. 
The action space is different from tasks. For navigation tasks like SimpleCrossing and LavaCrossing, actions only include turn left, turn right and move forward. For other tasks like DoorKey and Unlock, actions also include pickup a key and open a door.
Let MaxSteps be the max episode steps, MapWidth and MapHeight be the width and height of the map.
We introduce each task as follows.

\textbf{DoorKey.} This task is to first pickup the key, then open the door, and finally reach the goal state (green block). MaxSteps is defined as $10 \times \text{MapWidth} \times \text{MapHeight}$.
Reaching the goal state will get reward +MaxSteps/100, otherwise there is a penalty reward -0.01 for each step.

\textbf{Unlock.} This task is to first pickup the key and then open the door.
MaxSteps is defined as $8 \times \text{MapHeight}^2$.
Opening the door will get reward +MaxSteps/100, otherwise there is a penalty reward -0.01 for each step.

% \textbf{RedBlueDoors.} This task is to first open the red door and then open the blue door.
% MaxSteps is defined as $20 \times \text{MapHeight}^2$.
% The agent will get reward +MaxSteps/100 after the red door and the blue door are opened sequentially, otherwise there is a penalty reward -0.01 for each step.

\textbf{SimpleCross-Easy/Hard.} This task is to navigate through the room and reach the goal state (green block). Knocking into the wall will keep the agent unmoved.
MaxSteps is defined as $4 \times \text{MapWidth} \times \text{MapHeight}$.
Reaching the goal state will get reward +MaxSteps/100, otherwise there is a penalty reward -0.01 for each step.

\textbf{LavaCross-Easy/Hard.} This task is to reach the goal state (green block). Falling into the lava (orange block) will terminate the episode immediately.
MaxSteps is defined as $4 \times \text{MapWidth} \times \text{MapHeight}$.
Reaching the goal state will get reward +MaxSteps/100, falling into the lava will get reward -MaxSteps/100,  otherwise there is a penalty reward -0.01 for each step.

\subsection{MinAtar}

\begin{figure*}[htbp]
    % \vspace{-0.2in}
\begin{center}
    \subfigure[Asterix]{\Description{}\includegraphics[width=0.19\textwidth]{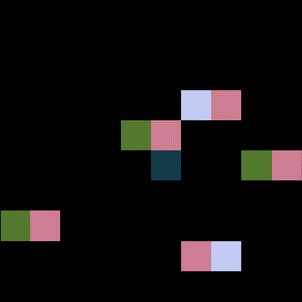}}
    \subfigure[Breakout]{\Description{}\includegraphics[width=0.191\textwidth]{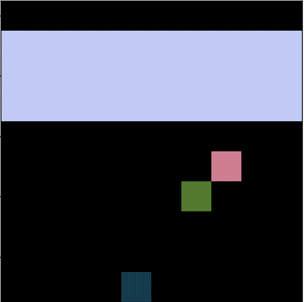}}
    \subfigure[Freeway]{\Description{}\includegraphics[width=0.19\textwidth]{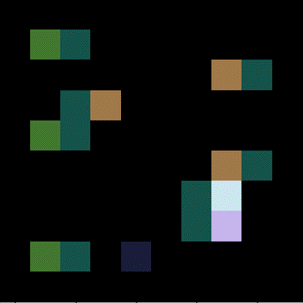}}
    \subfigure[Seaquest]{\Description{}\includegraphics[width=0.1905\textwidth]{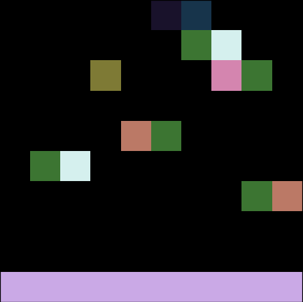}}
    \subfigure[SpaceInvaders]{\Description{}\includegraphics[width=0.1925\textwidth]{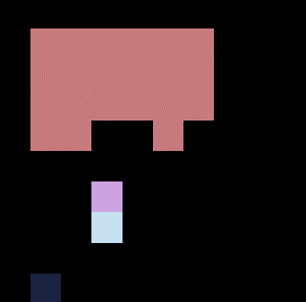}}
    % \vskip -0.1in
    \caption{Visualization of MinAtar environments.}
    \vspace{-0.1in}
    \label{fig:MinAtar visualization.}
\end{center}
\end{figure*}

MinAtar~\citep{young19minatar}~\footnote{\url{https://github.com/kenjyoung/MinAtar}} is image-based miniaturized version of Atari environments~\citep{bellemare2013arcade}, which maintains the mechanics of the original games as much as possible and is much faster than original version. 
MinAtar implements five Atari games in total, we show the visualization of each game in Figure~\ref{fig:MinAtar visualization.}.

\textbf{State Space.} Each game provides the agent with a 10 × 10 × $n$ binary state representation. 
The $n$ channels correspond to game specific objects, such as ball, paddle and brick in the game Breakout.
The objects in each game are randomly generated at different time steps. The difficulty will change as the game progresses, for examples, there will be more objects and the objects will move faster.
So these environments needs the policy to generalize across different configurations. 

\textbf{Action Space.} The action space consists of moving in the 4 cardinal directions, firing, and no-op, and omits diagonal movement as well as actions with simultaneous firing and movement. This simplification increases the difficulty for decision making. 
In addition, MinAtar games add stochasticity by incorporating sticky-actions, that the environment repeats the last action with probability 0.1 instead of executing the agent’s current action.
This can avoid deterministic behaviour that simply repeats specific sequences of actions, rather than learning policies that generalize.

\textbf{Reward Function.} The rewards in most of the MinAtar environments are either 1 or 0. The only exception is Seaquest, where a bonus reward between 0 and 10 is given proportional to remaining oxygen when surfacing with six divers.

\subsection{Wall-Clock Time Comparison}
\label{sec: wall-clock time comparison}

\begin{table*}[htbp]
    % \vskip -0.1in
    % \vskip 0.2in
    \caption{Wall-clock time comparison between different methods. We use \emph{Frames Per Second} (FPS) to measure the speed of interaction with environments during training. Our method adds mild computational overhead on DQN.}
    \label{table:wall clock time comparison details}
        \centering
        \scalebox{1.}{
            \begin{tabular}{cccc}
                \toprule[1pt]
                Method & FPS (mean $\pm$ std) & Computational Cost & Performance/Computational Cost\\ \midrule
                DQN &  870.64 $\pm$ 7.59 & 100\% & 1  \\
                Bootstrapped DQN & 445.72 $\pm$ 14.68 & 195.34\% & 0.50 \\ 
                $\epsilon$z-greedy & 923.04 $\pm$ 38.58 & 94.32\% & 0.76\\
                RND & 570.63 $\pm$ 32.31 & 152.57\% & 0.75 \\
                LESSON & 234.63 $\pm$ 12.71 & 371.0\% & 0.29\\
                $\beta$-DQN (Ours) & 627.36 $\pm$ 5.34 & 138.78\% & \textbf{1.13} \\
                \bottomrule[1pt]
            \end{tabular}
        }
        %	\vspace{1.5ex}
\end{table*}

Only reporting sample efficiency cannot tell us how long we need to train an agent for each method. We compare the wall-clock time during training in 
% Appendix~\ref{sec: wall-clock time comparison} 
Table~\ref{table:wall clock time comparison details}. We use \emph{Frames Per Second} (FPS) to measure the training speed. FPS counts the number of frames that the agent interacts with the environment per second. 

We test the speed on device with GPU NVIDIA RTX A5000 and CPU AMD EPYC 7313 16-Core Processor.
Each time we run 1 experiments and do 3 runs for each method.
We can find our method $\beta$-DQN (FPS:627) is slower than DQN (FPS: 870) and faster than methods like Bootstrapped DQN (FPS:445), RND (FPS:570) and LESSON (FPS:234). 
% Though $\beta$-DQN is only slightly faster than Bootstrapped DQN, the computation will not increases linearly with the increase of the number of polices.
In addition, since we use a simple network architecture,
% with a single convolutional layer followed by a fully connected layer, 
the computational cost is relatively low for networks comparing with the computation consumed by environments.
If we use a large network, the computation gap will be larger between our method and other methods which have more networks like RND and LESSON.
$\epsilon$z-greedy run a little faster than DQN because it sample a random action and act for a random duration. This will consume less inference from $Q$ network.

\section{Additional Experimental Results}
\label{appendix:Additional Experimental Results}

\subsection{Toy example}
\begin{figure*}[htbp]
  \centering
    \vspace{0.05in}
    \subfigure{\raisebox{0.15\height}{\Description{}\includegraphics[width=0.39\textwidth]{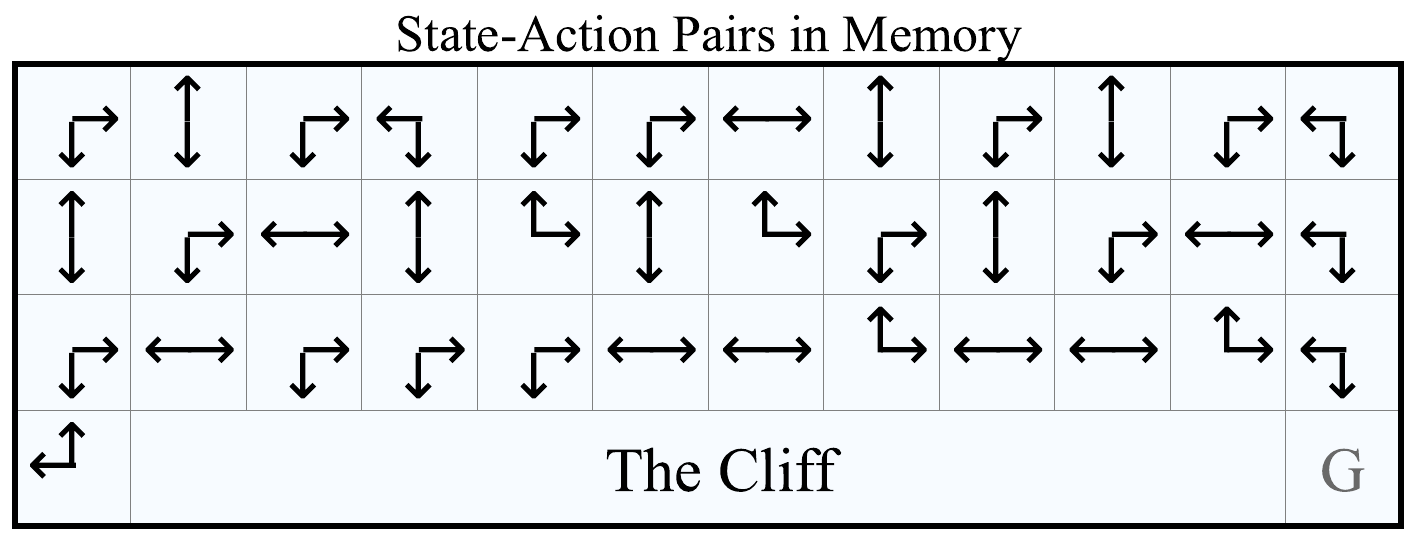}}}
    \subfigure{\Description{}\includegraphics[width=0.41\textwidth]{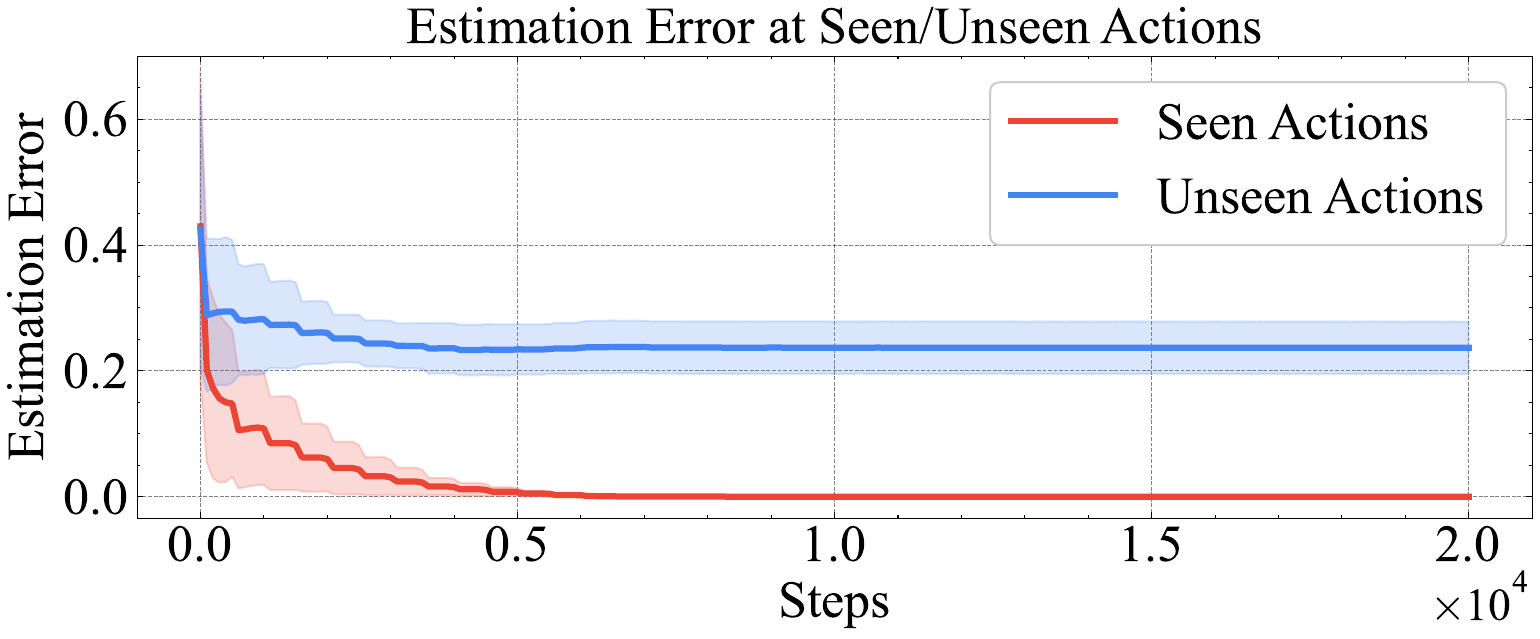}}
    \vskip -0.15in
    \subfigure{\Description{}\includegraphics[width=0.38\textwidth]{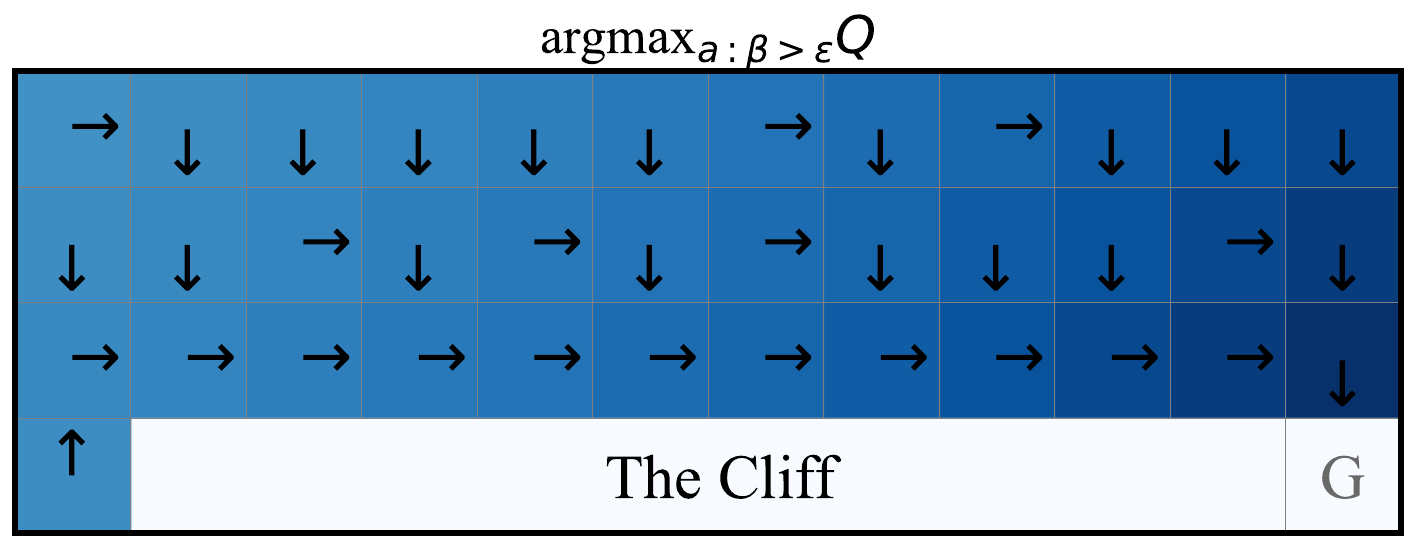}}
    \subfigure{\Description{}\includegraphics[width=0.38\textwidth]{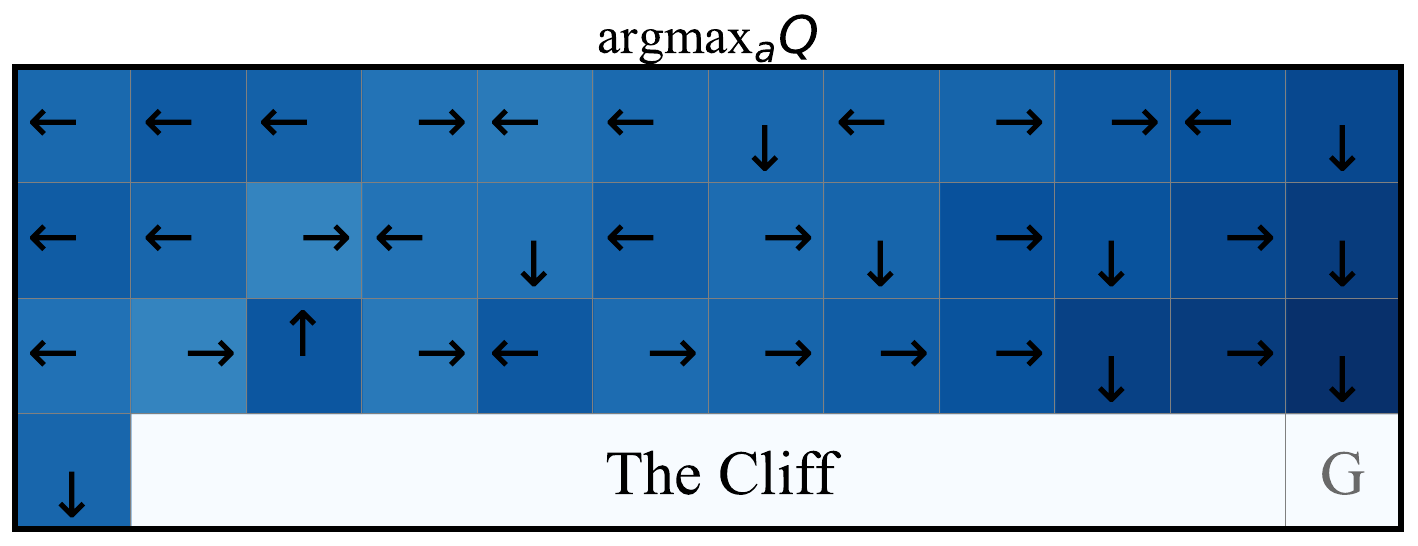}}
    % \vskip -0.05in
    \subfigure{\Description{}\includegraphics[width=0.04\textwidth, height=0.135\textwidth]{pic/toy_example/colorbar_horizontal.pdf}}
    \subfigure{\Description{}\includegraphics[width=0.44\textwidth]{pic/toy_example/horizontal_legend.pdf}}
    \vskip -0.1in
    \subfigure{\Description{}\includegraphics[width=0.4\textwidth]{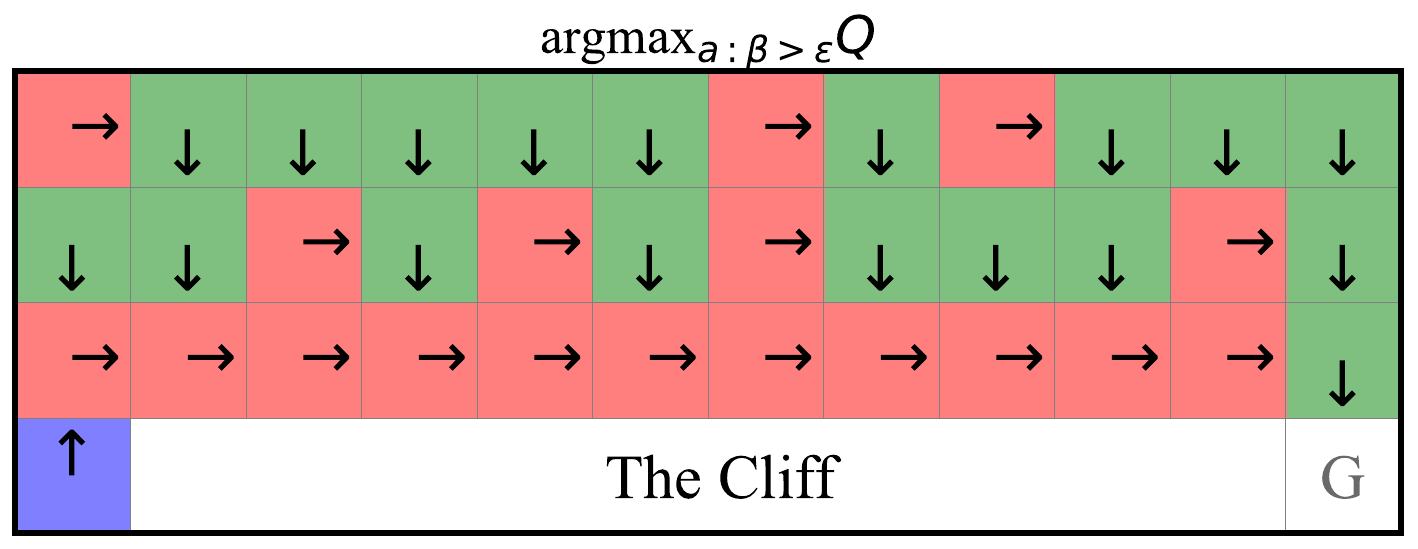}}
    \subfigure{\Description{}\includegraphics[width=0.4\textwidth]{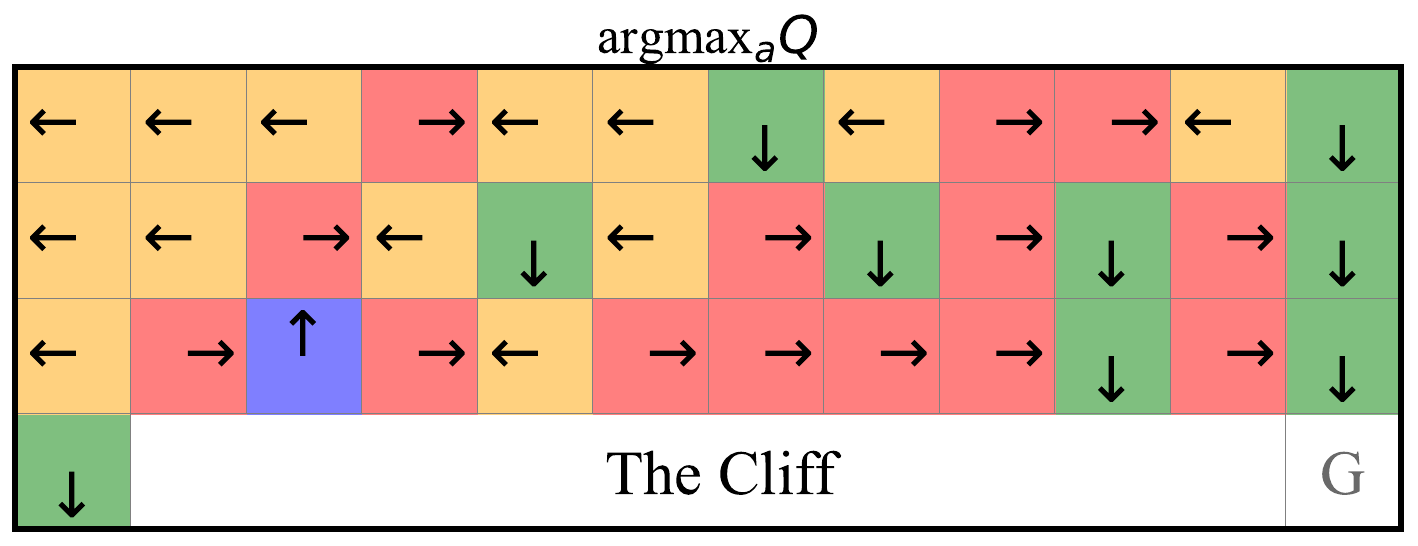}}
    \vskip -0.1in
    \subfigure{\Description{}\includegraphics[width=0.4\textwidth]{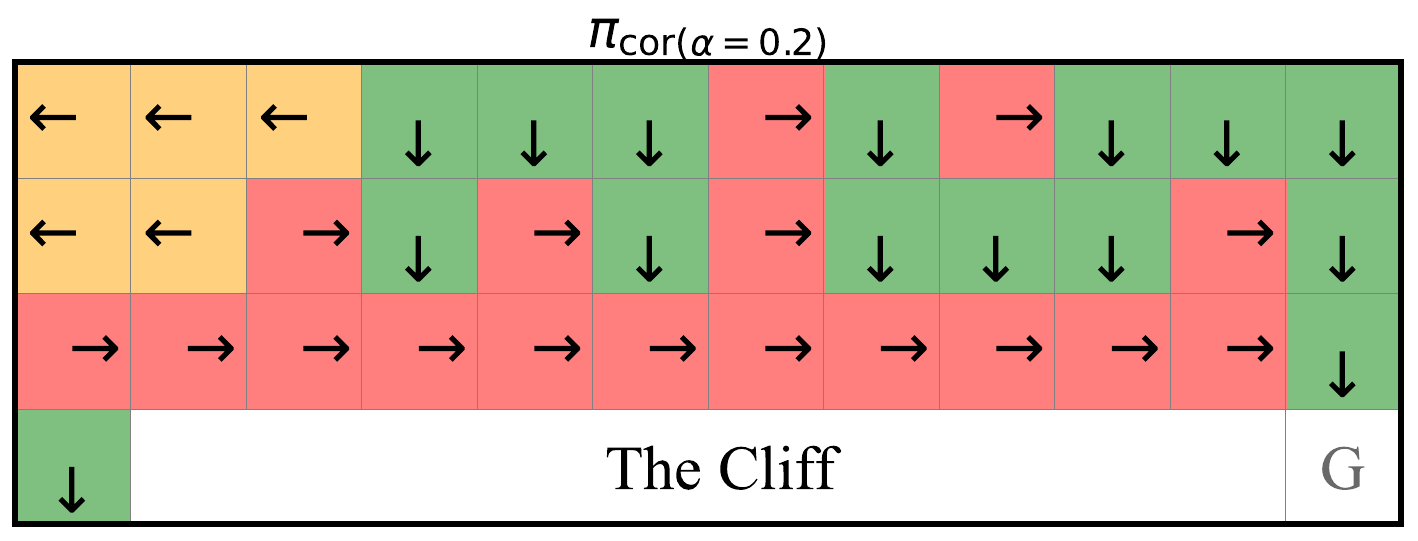}}
    \subfigure{\Description{}\includegraphics[width=0.4\textwidth]{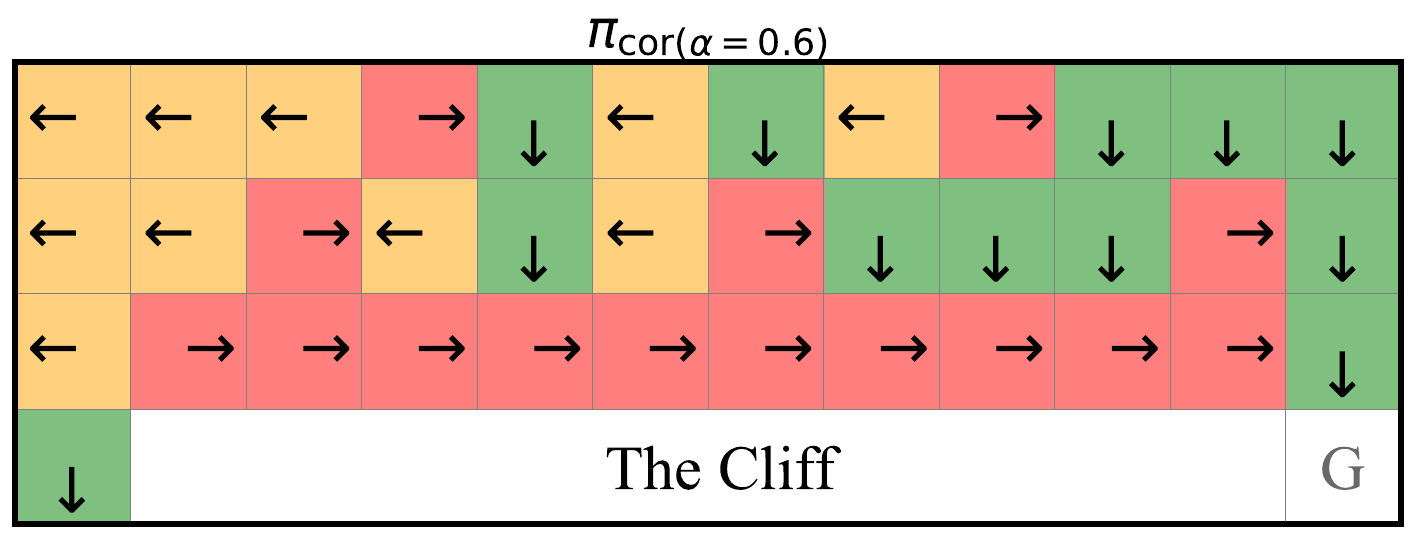}}
    \vskip -0.1in
    \subfigure{\Description{}\includegraphics[width=0.4\textwidth]{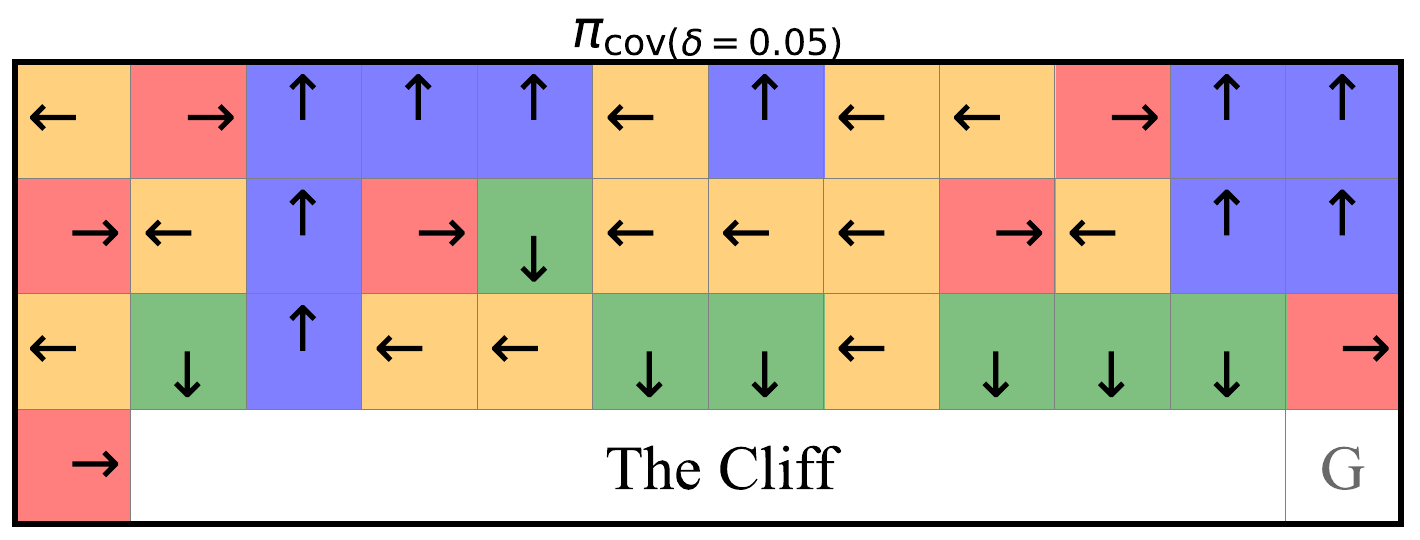}}
    \subfigure{\Description{}\includegraphics[width=0.4\textwidth]{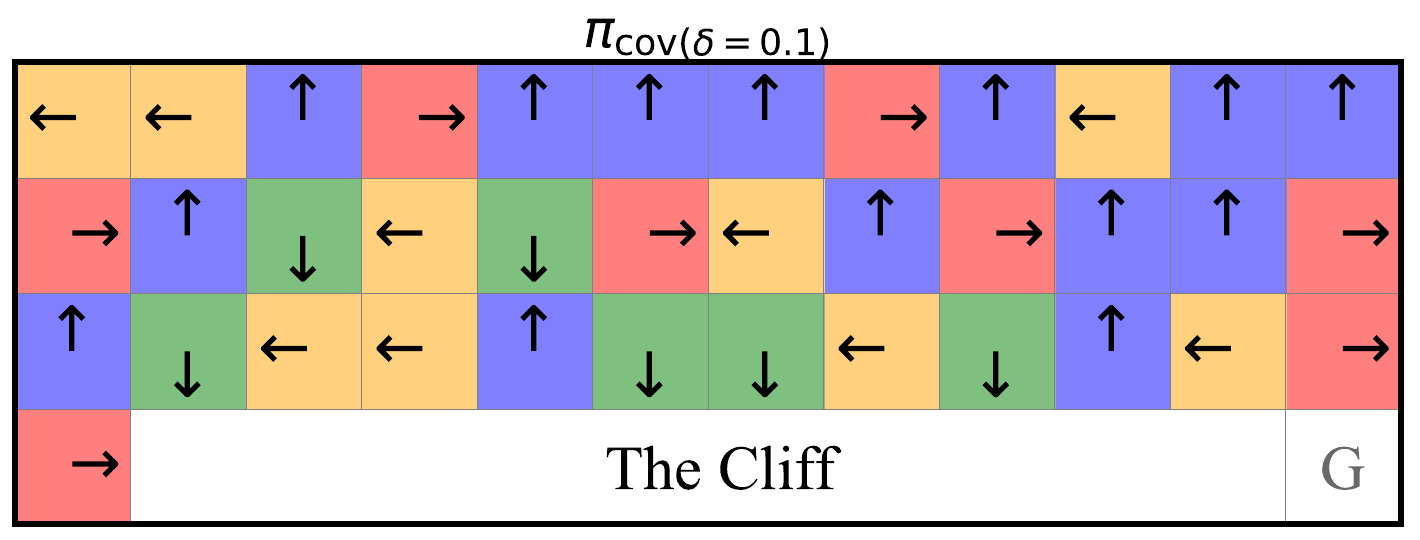}}
    \vskip -0.1in
    \caption{
    Policy diversity at a specific case. The top left figure shows the state-action pairs in the current memory. All states have been visited and each state has at least one optimal action in the replay memory, but there are still actions have not been tried. The top right figure shows $Q$ value errors at seen and unseen actions. And the second rows explicitly show the action values for the greedy actions of $Q_{\textit{mask}}$ and $Q$, which indicates $Q$ may overestimate at some states and $Q_{\textit{mask}}$ can yield a best possible policy, implying different roles of $Q$ and $Q_{\textit{mask}}$.
    In the remaining rows, we show diverse polices derived from $\beta,Q$ and $Q_{\textit{mask}}$. These polices take different actions at these states, which benefits the learning. 
    }
    % \vskip -0.1in
    \label{fig:another toy example}
\end{figure*}

Besides the example given in~\cref{fig:toy example}, we give another example with full state coverage as shown in~\cref{fig:another toy example}. 
The top left shows all states have been visited. 
Each state has at least one optimal action in the replay memory, but there are still actions have not been tried.
The function $\beta$ learned by \cref{eq:behavior policy learning} will assign probability uniformly to existing actions at these states and other actions as 0.
The top right figure shows the estimation error of function $Q$ learned by~\cref{eq:in distribution td learning} at seen and unseen actions.
We can find $Q$ learns accurate estimates at seen actions but inaccurate at unseen actions, which indicates $Q$ may overestimate at some states and $Q_{\textit{mask}}$ can yield a best possible policy following current data in the memory.
A detailed illustration is given at the second row. 
The blue shading indicates the action values for the greedy actions of $Q_{\textit{mask}}$ and $Q$.
We can find $Q_{\textit{mask}}$ learn the accurate estimate, but $Q$ overestimates at some state-action pairs and take wrong actions.
When combining $\beta,Q$ and $Q_{\textit{mask}}$, we obtain a group of diverse policies as shown in the remaining rows. 
These polices take different actions and explore the whole state action space.

\subsection{Overall Performance}
\begin{figure*}[htbp]
    % \vskip -0.1in
\begin{center}
    \subfigure{\includegraphics[width=0.9\textwidth]{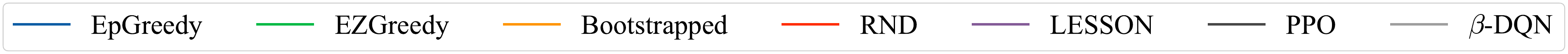}}
    \vskip -0.03in
    \subfigure{\Description{}\includegraphics[width=0.24\textwidth]{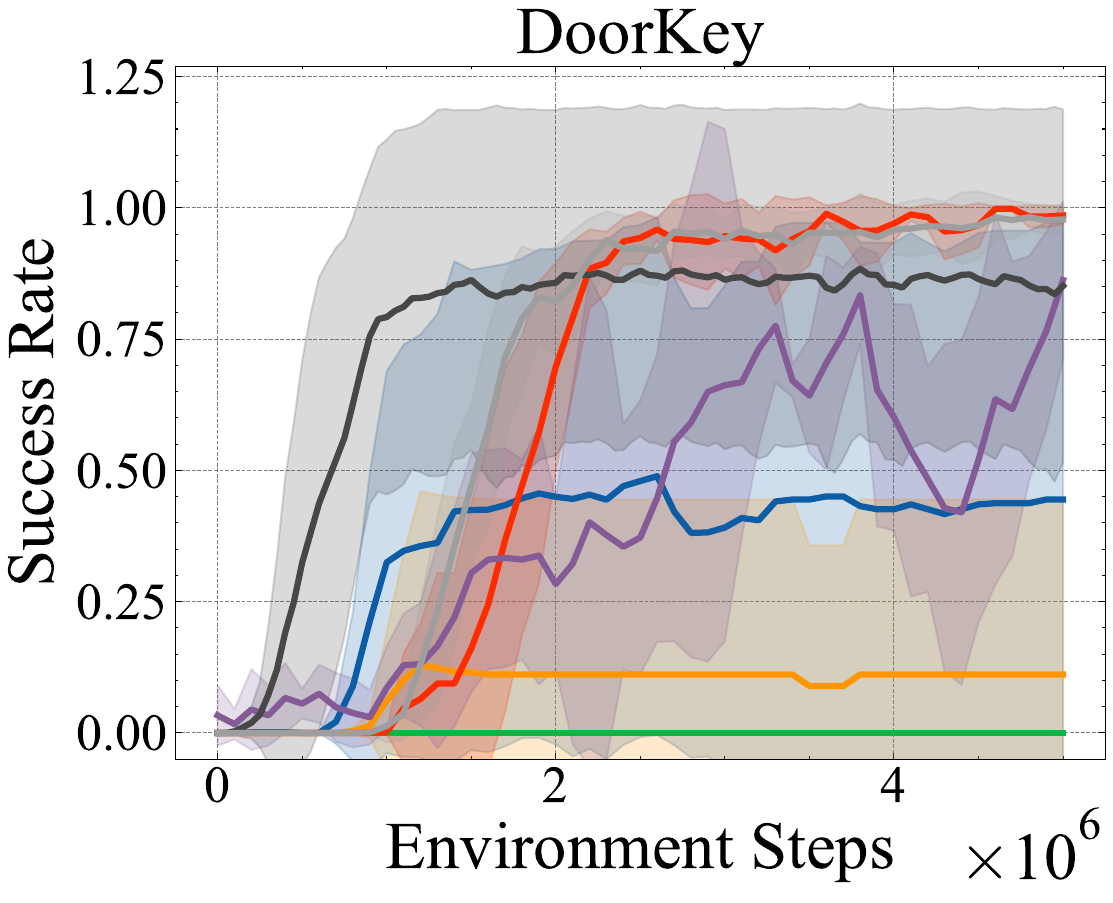}}
    \subfigure{\Description{}\includegraphics[width=0.24\textwidth]{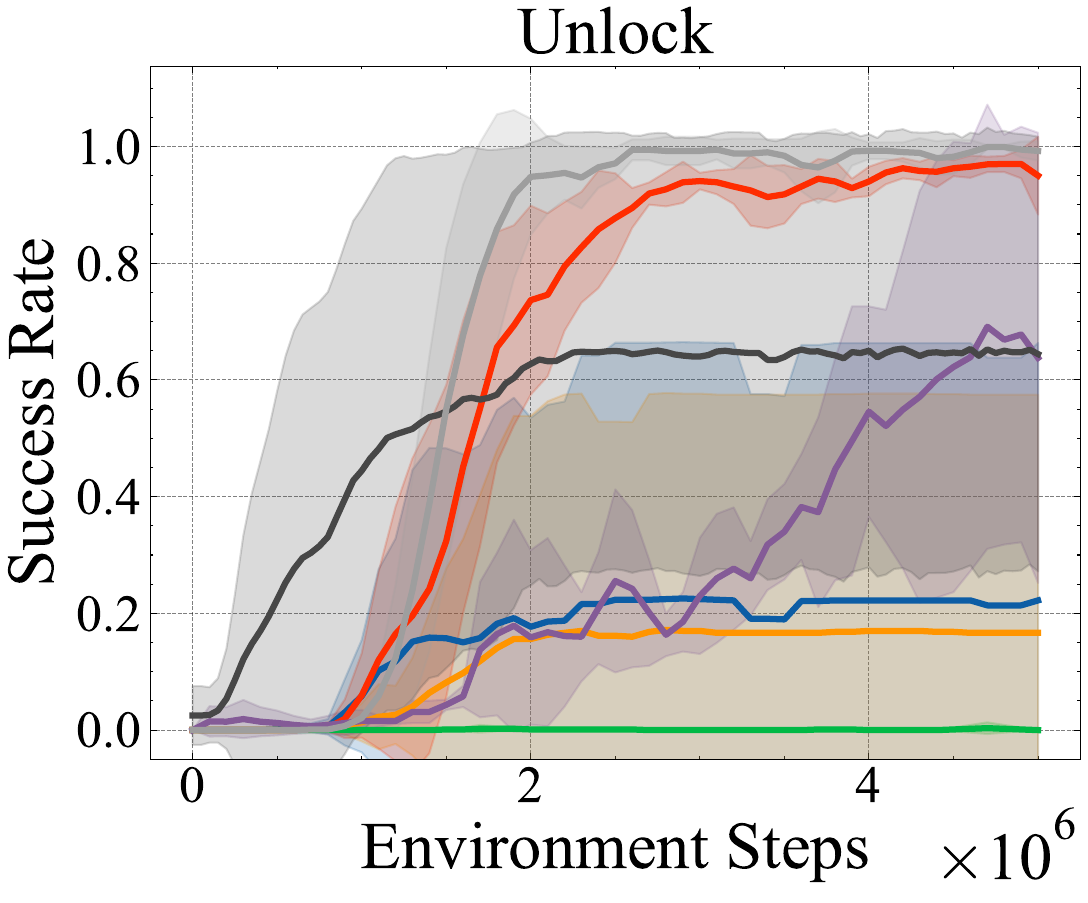}}
    \subfigure{\Description{}\includegraphics[width=0.24\textwidth]{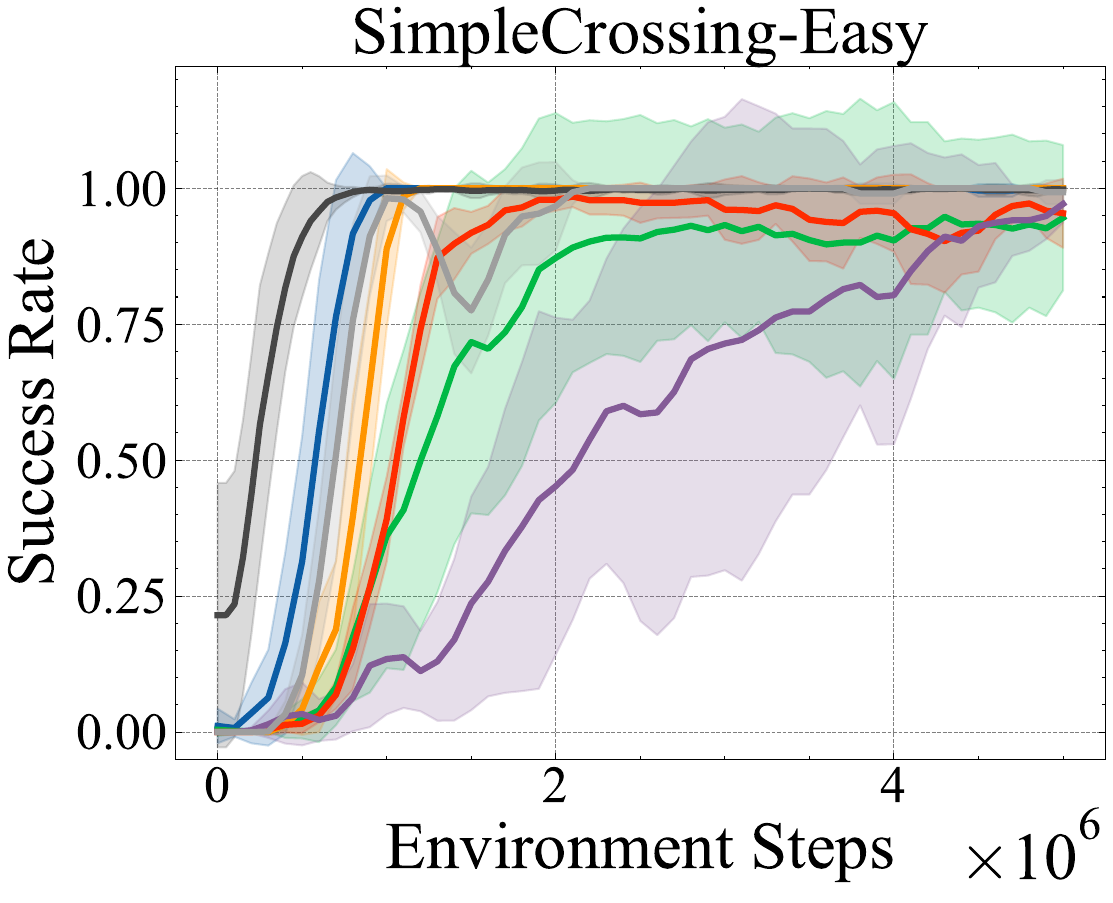}}
    \subfigure{\Description{}\includegraphics[width=0.24\textwidth]{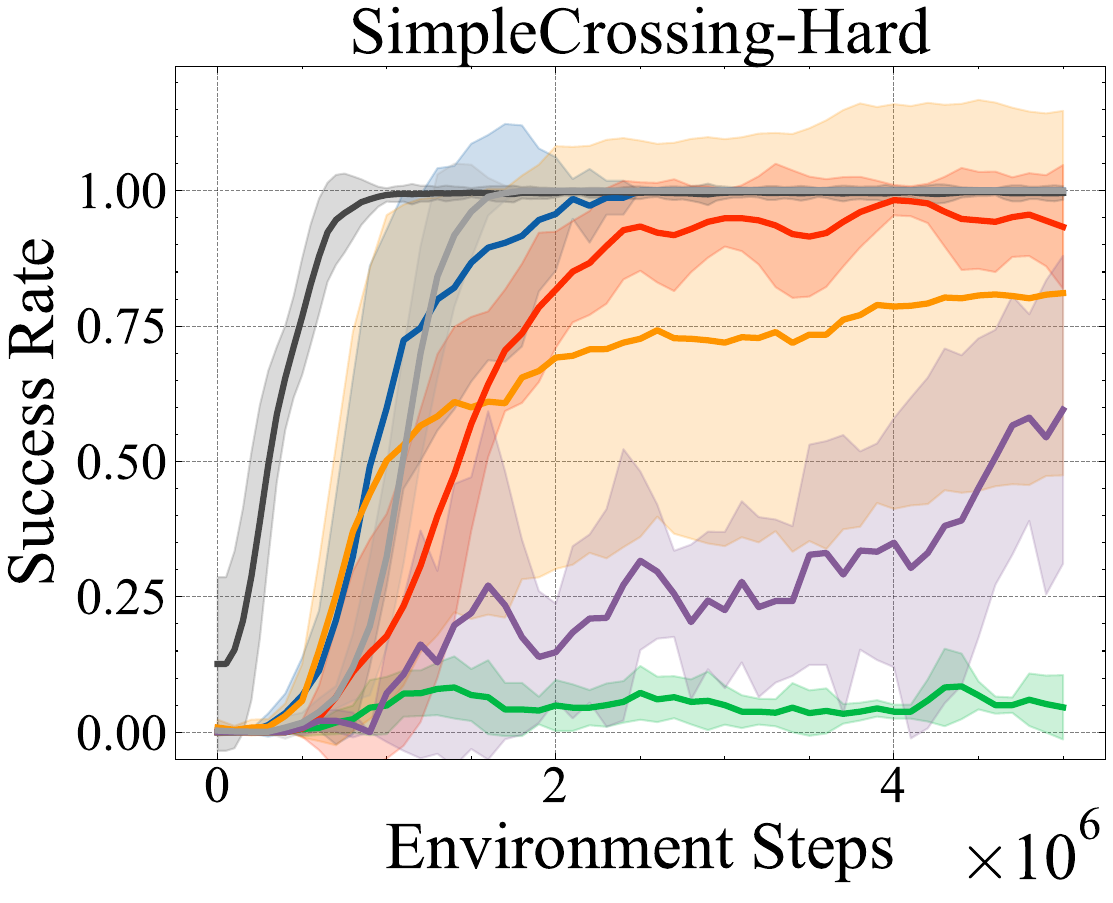}}
    \subfigure{\Description{}\includegraphics[width=0.24\textwidth]{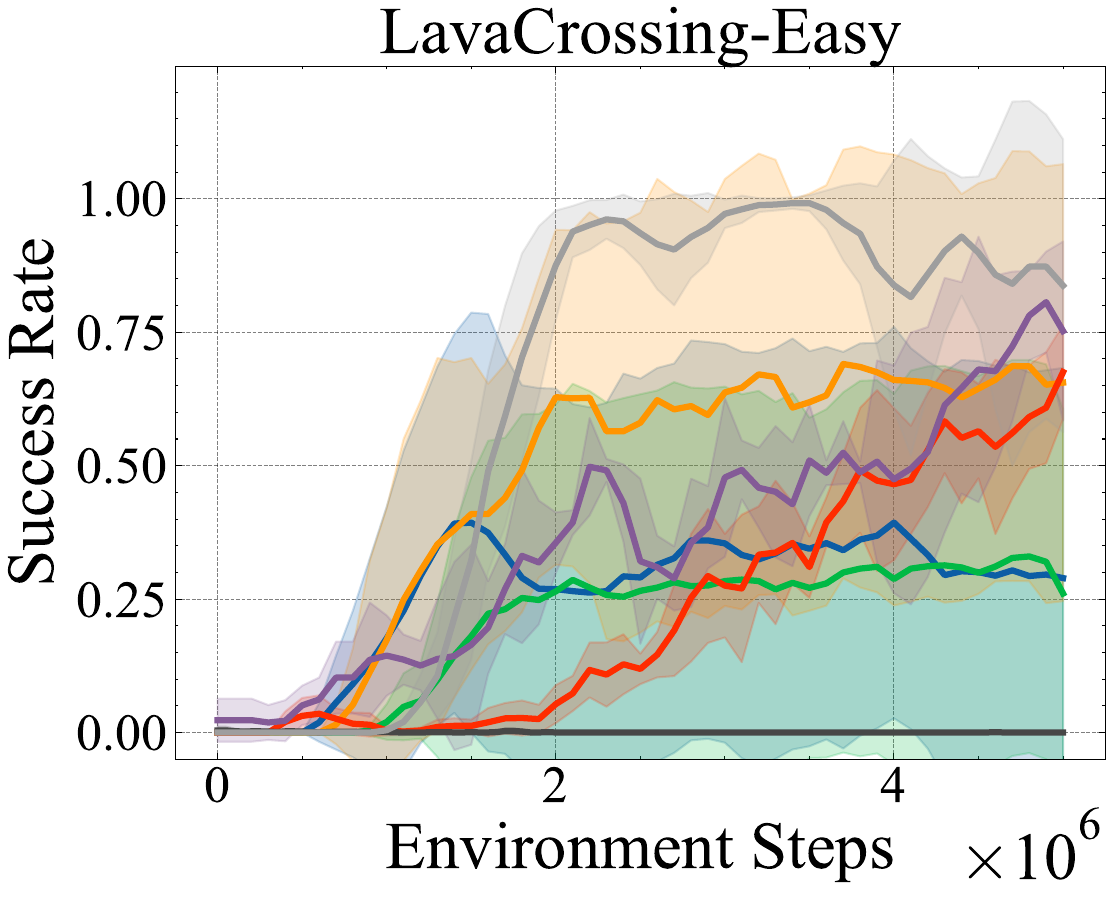}}
    \subfigure{\Description{}\includegraphics[width=0.24\textwidth]{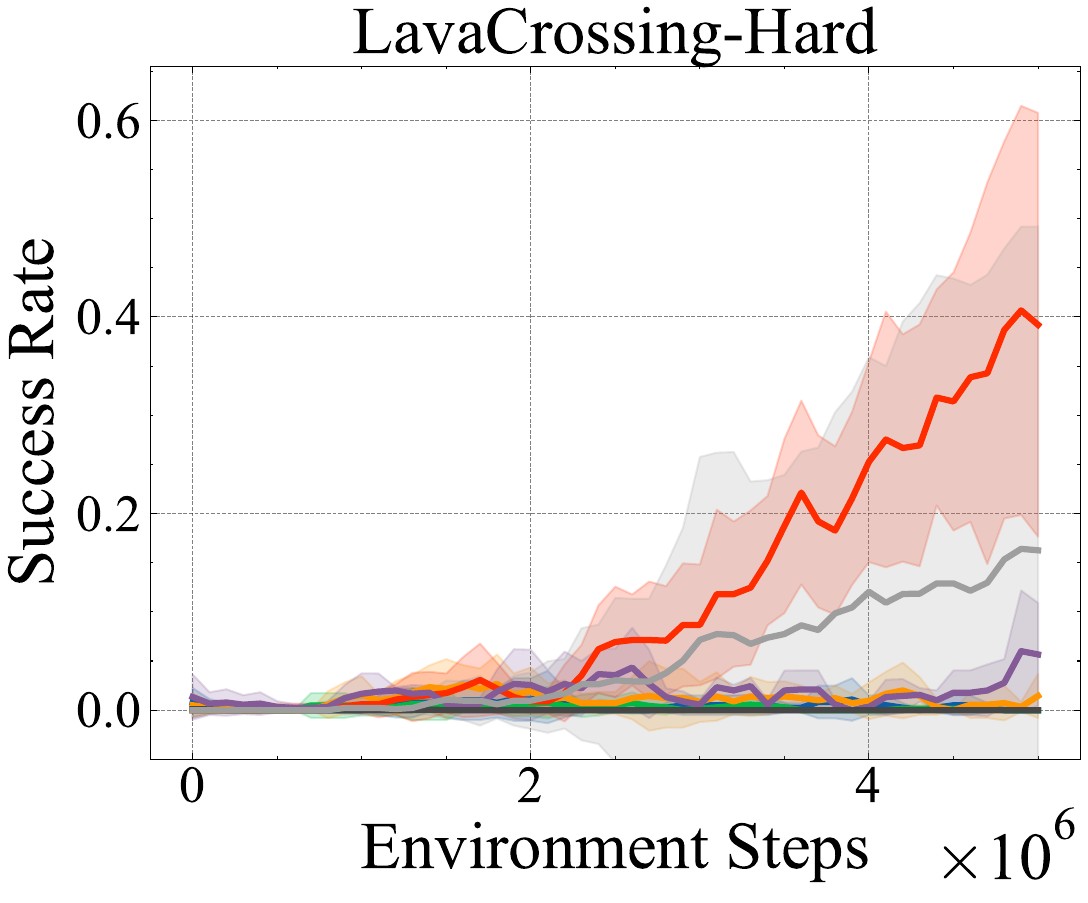}}
    \subfigure{\Description{}\includegraphics[width=0.24\textwidth]{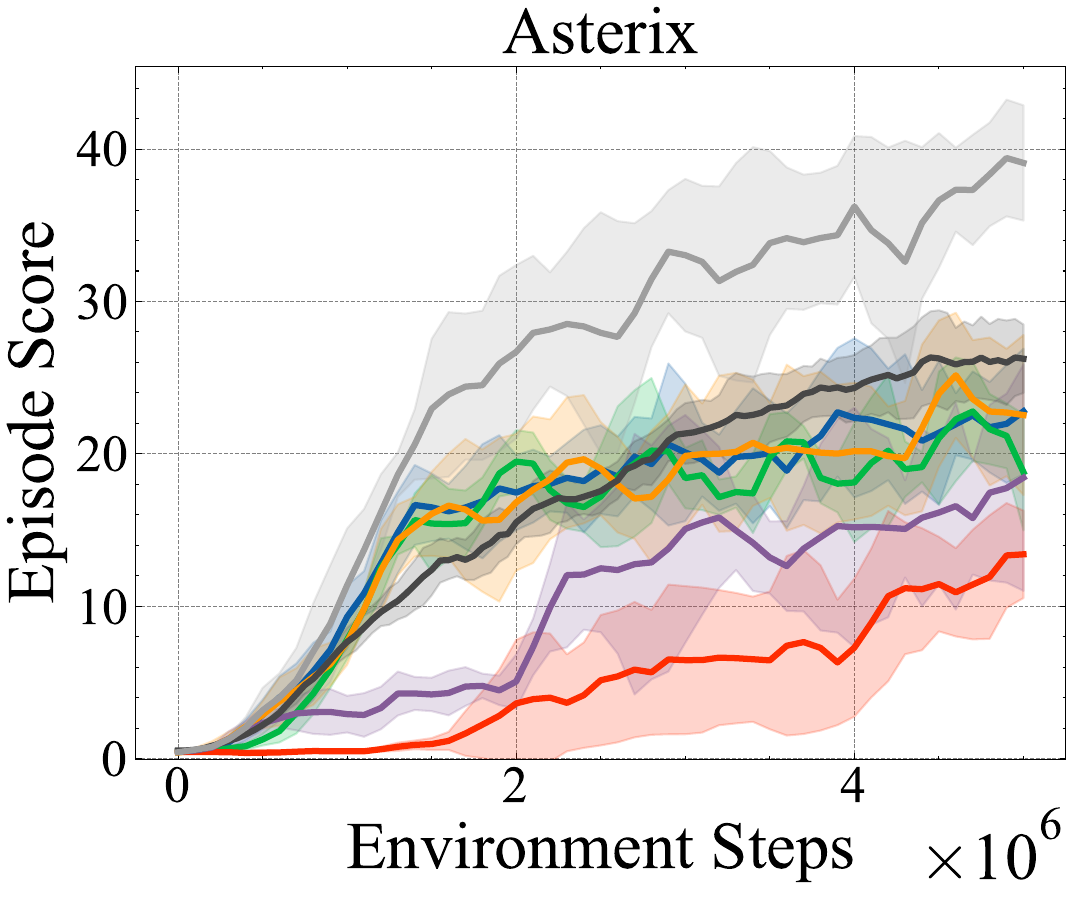}}
    \subfigure{\Description{}\includegraphics[width=0.24\textwidth]{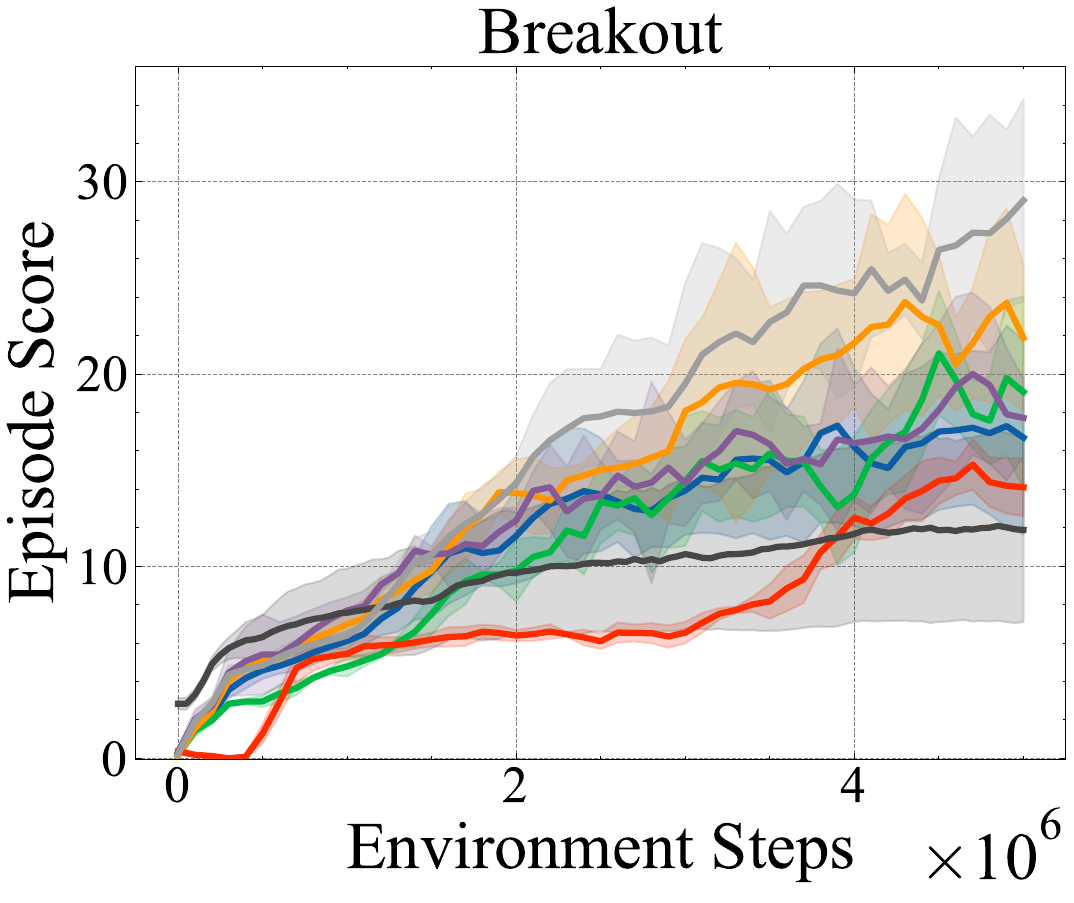}}
    \subfigure{\Description{}\includegraphics[width=0.24\textwidth]{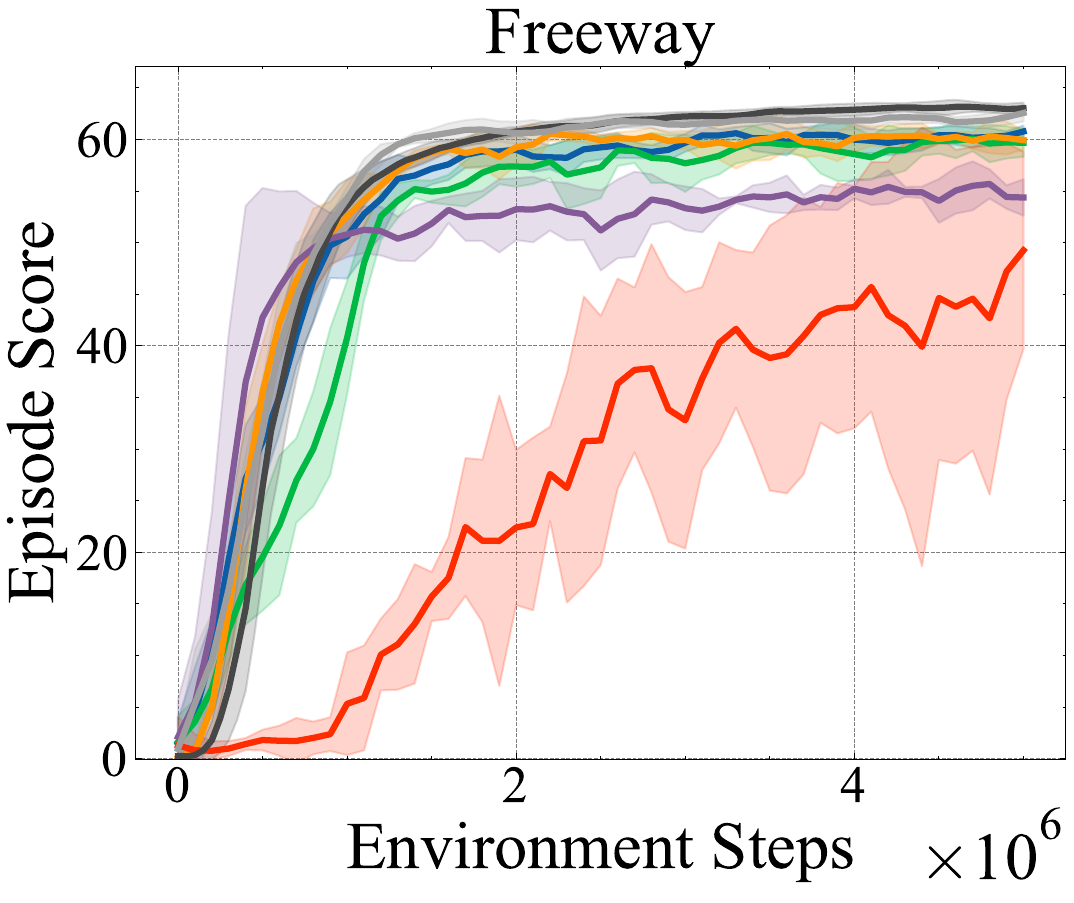}}
    \subfigure{\Description{}\includegraphics[width=0.24\textwidth]{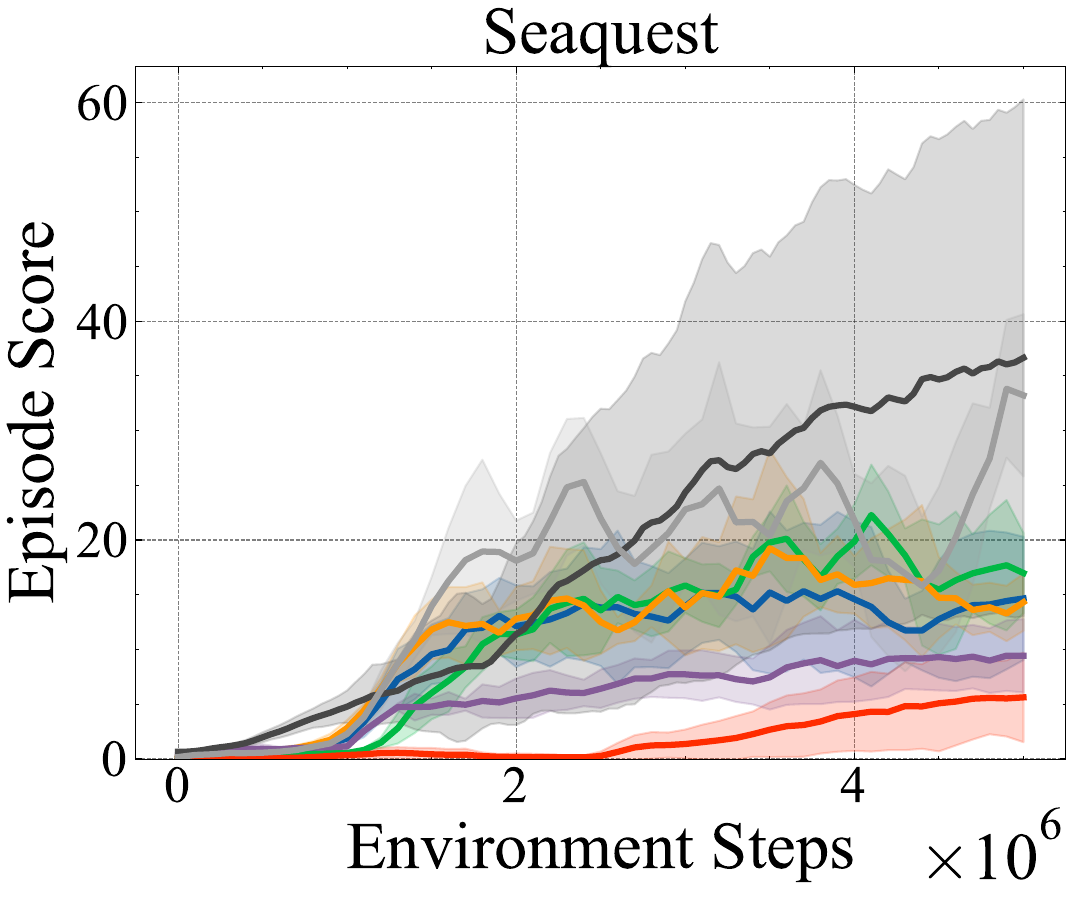}}
    \subfigure{\Description{}\includegraphics[width=0.24\textwidth]{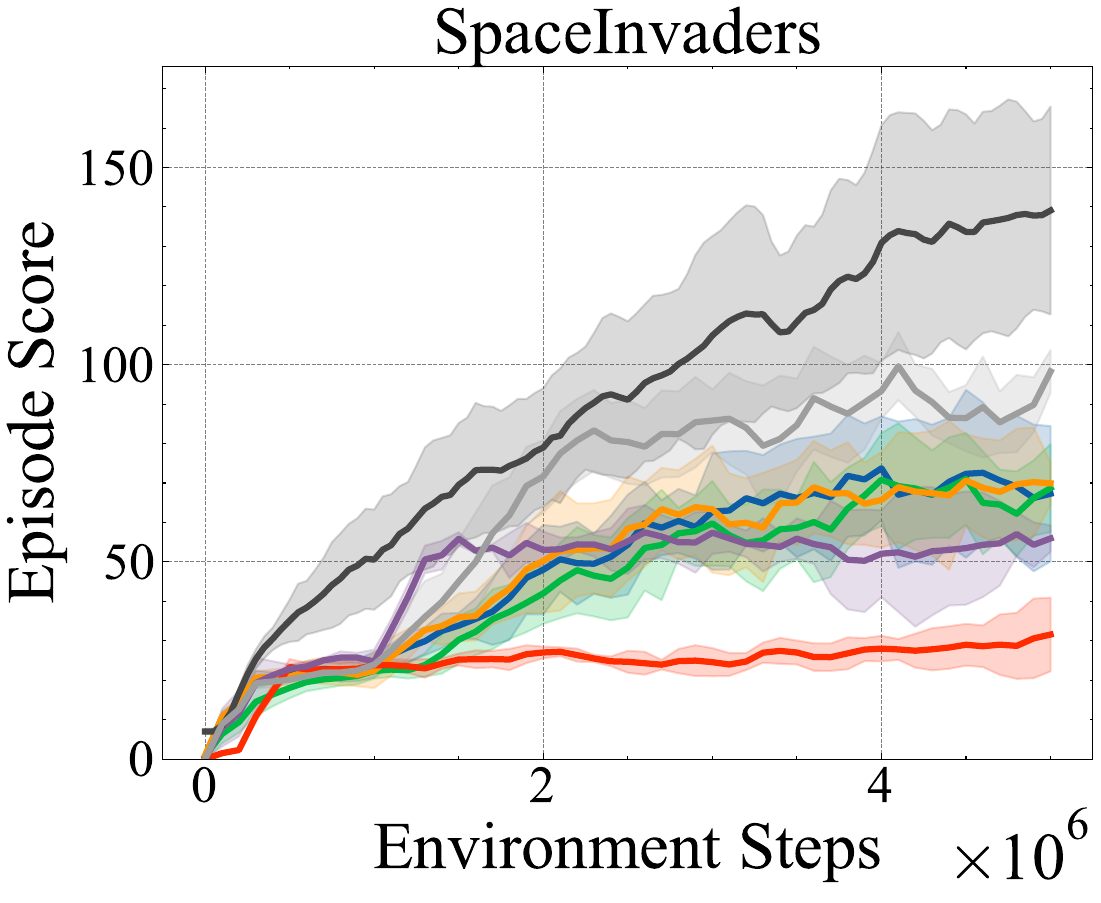}}
    \vskip -0.1in
    \caption{
    Learning curves on all environments. Our method achieves the best performance on most of environments across easy and hard exploration domains, which indicates our method is general and effective.
    }
    % \vskip -0.2in
    \label{fig:performance}
\end{center}
\end{figure*}

We show the learning curves of each method on MiniGrid and MinAtar in~\cref{fig:performance}.
Each line is the average of running 10 different random seeds. 
The solid line shows the mean success rate for MiniGrid and the mean return for MinAtar. The shaded area shows the standard error.
We can find our method $\beta$-DQN achieves the best performance on most of environments across easy and hard exploration domains, which indicates our method achieves diverse exploration and helps the learning.
This highlights that our methods is general and suitable for a
ll kinds of tasks.

\subsection{Additional Analysis}
\label{appendix:additional analysis}

\subsubsection{The role of the two kinds of exploration policies}

We construct two kind of exploration polices in our policy set, $\pi_{\textit{cov}}$ for state space coverage and $\pi_{\textit{cor}}$ for bias correction. We can get many of these polices with different $\delta$ and $\alpha$. 
To show the role of the two kinds of exploration policies, we count them together for clear illustration. 
For example our main result is based on policy set $\Pi = \{\pi_{\text{cov}(0.05)},\pi_{\text{cov}(0.1)}, \pi_{\text{cor}(0)},\pi_{\text{cor}(0.1)},\pi_{\text{cor}(0.2)},\cdots,\pi_{\text{cor}(1)} \}$, and sliding-window length $L=1000$.
We count $\{\pi_{\text{cov}(0.05)},\pi_{\text{cov}(0.1)}\}$ together as $\pi_{\textit{cov}}$ for state space coverage, and $\{\pi_{\text{cor}(0.1)},$ $\pi_{\text{cor}(0.2)},$ $\cdots,\pi_{\text{cor}(1)} \}$ together as $\pi_{\textit{cor}}$ for bias correction. 
We also plot $\pi_{\text{cor}(0)}$ separately, since it is a pure exploitation policy.

\begin{figure*}[htbp]
    % \vskip -0.1in
\begin{center}
    \subfigure{\includegraphics[width=0.36\textwidth]{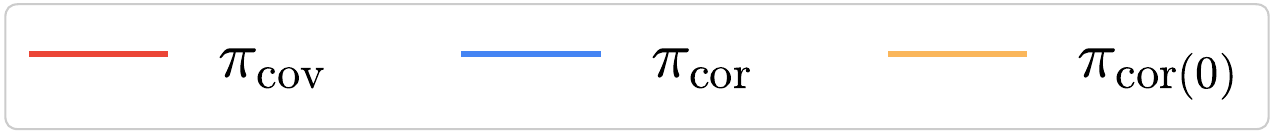}}
    \vskip -0.03in
    \subfigure{\Description{}\includegraphics[width=0.24\textwidth]{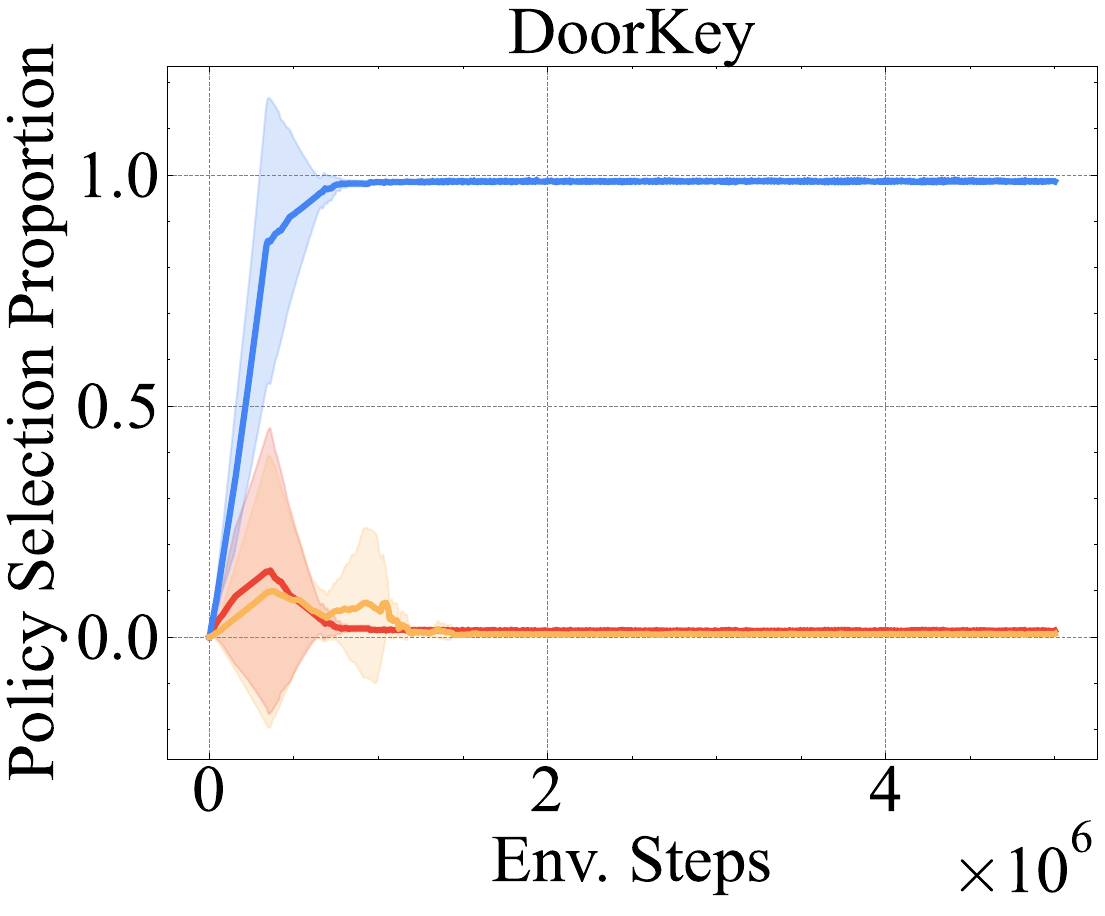}}
    \subfigure{\Description{}\includegraphics[width=0.24\textwidth]{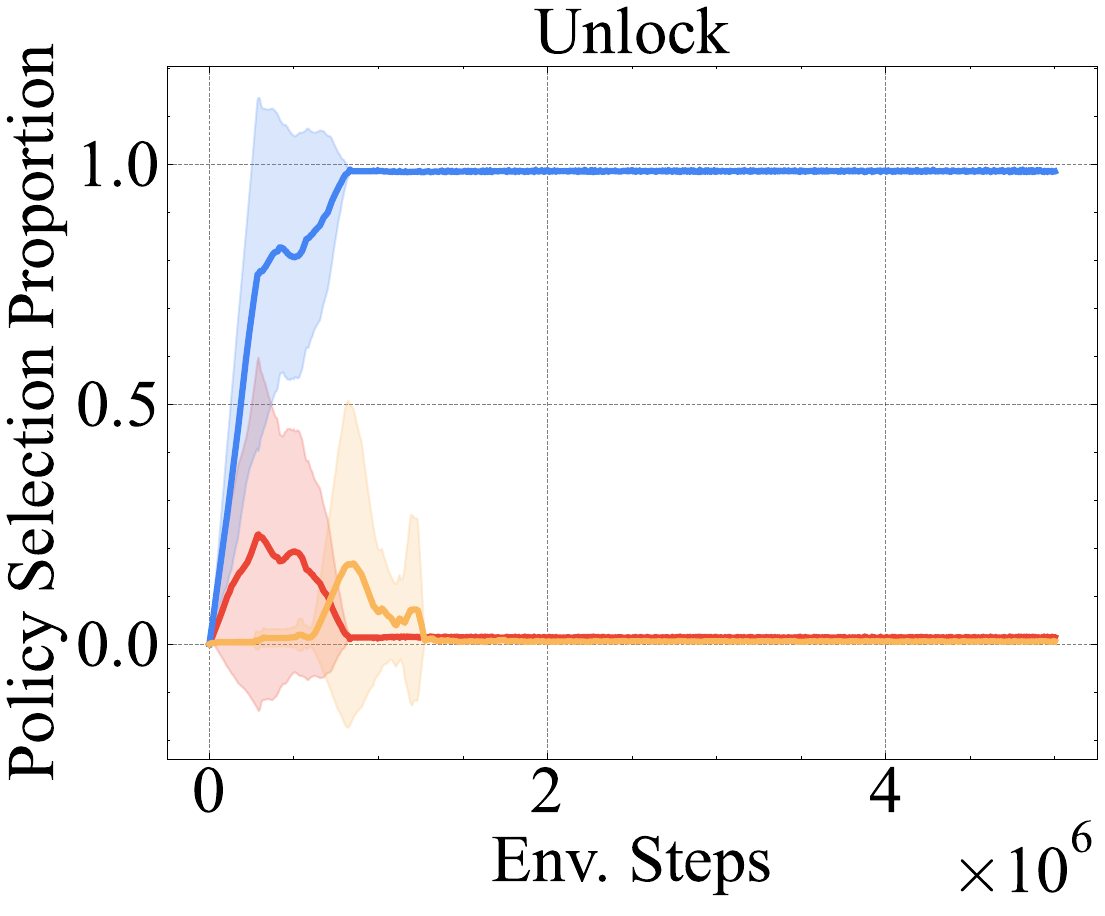}}
    \subfigure{\Description{}\includegraphics[width=0.24\textwidth]{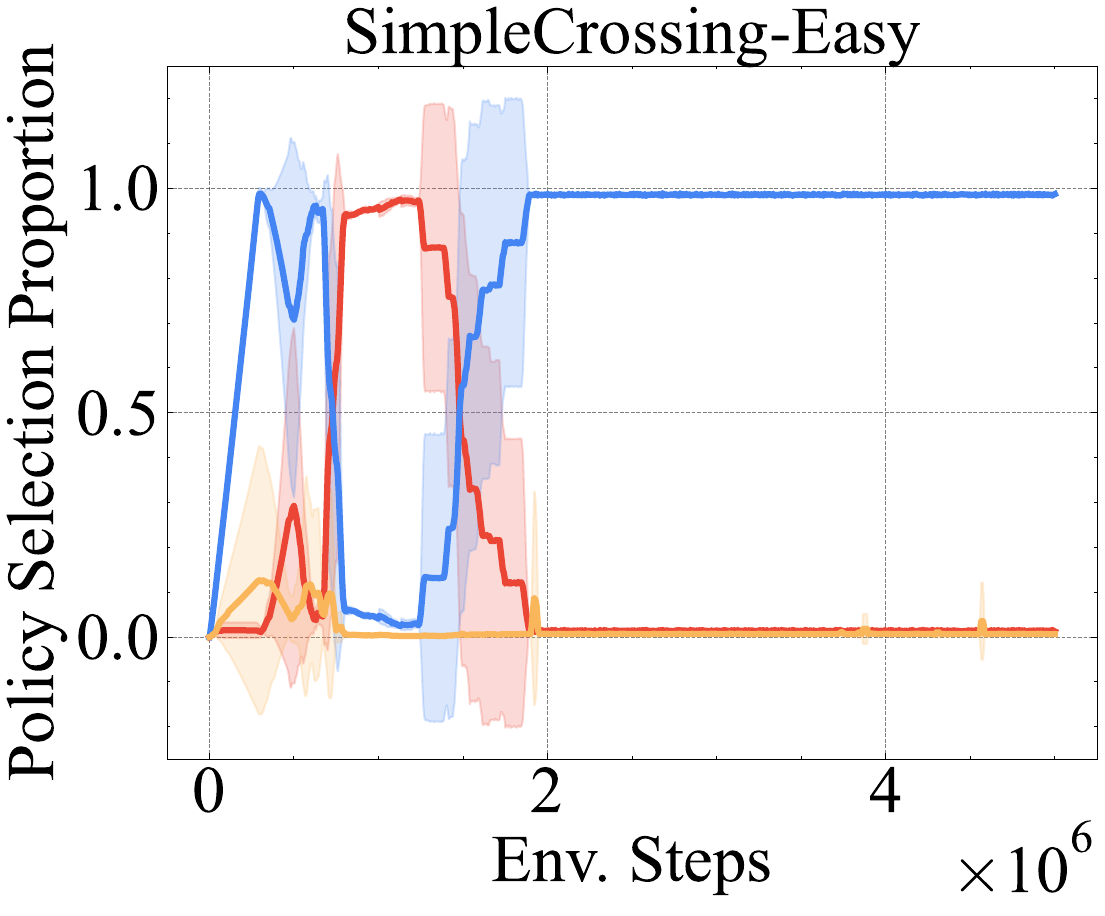}}
    \subfigure{\Description{}\includegraphics[width=0.24\textwidth]{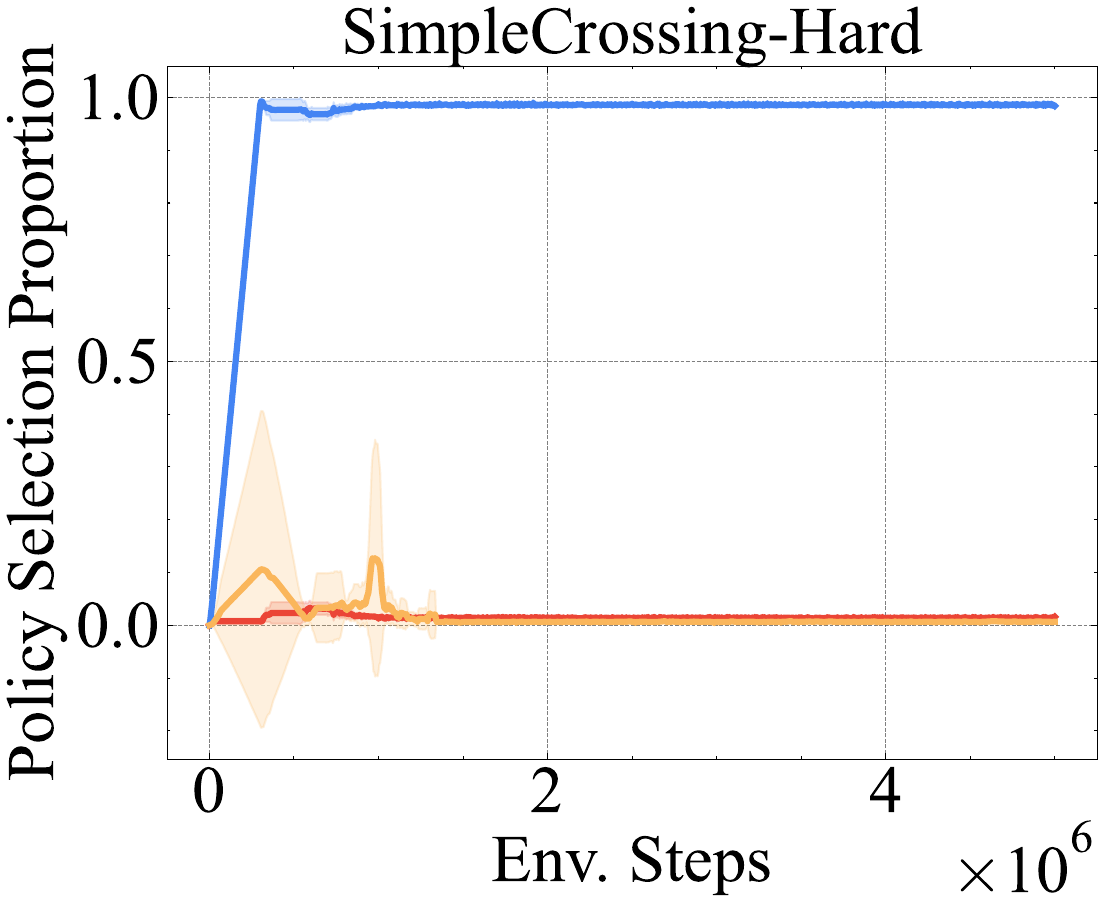}}
    \subfigure{\Description{}\includegraphics[width=0.24\textwidth]{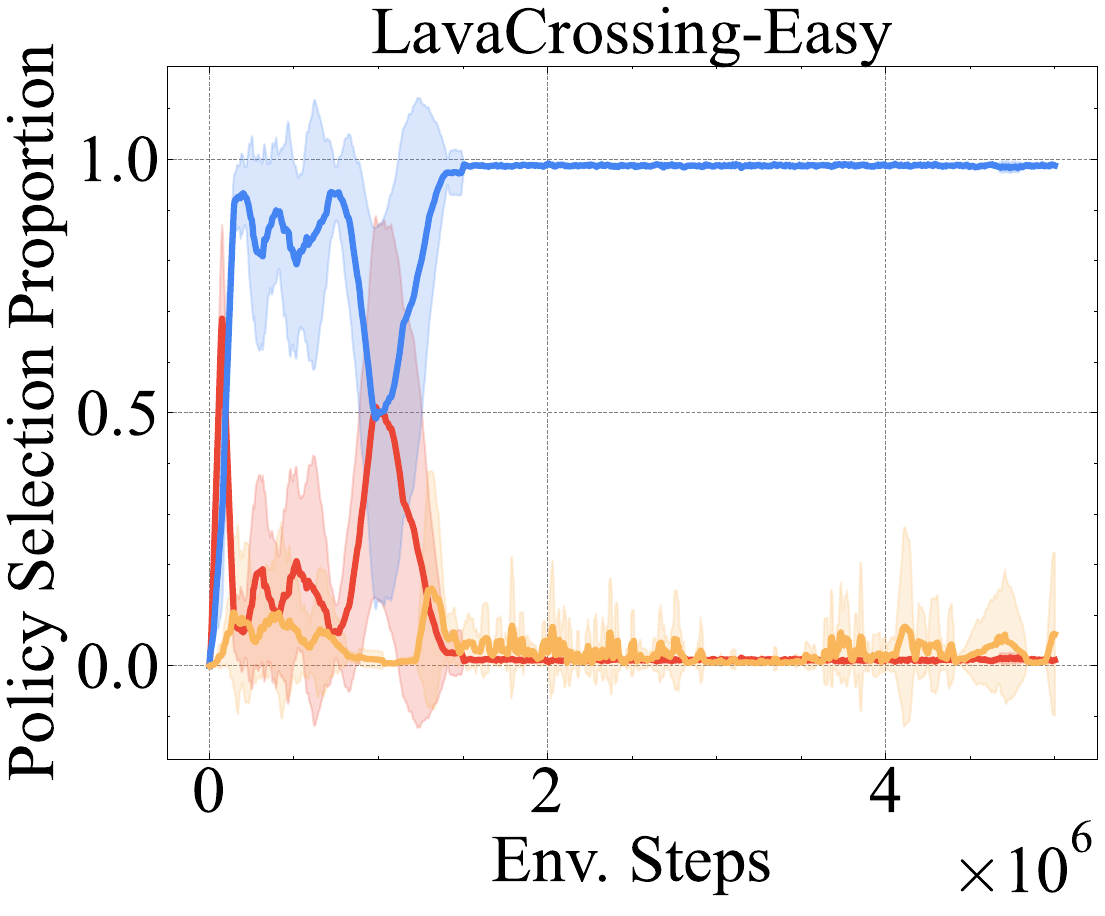}}
    \subfigure{\Description{}\includegraphics[width=0.24\textwidth]{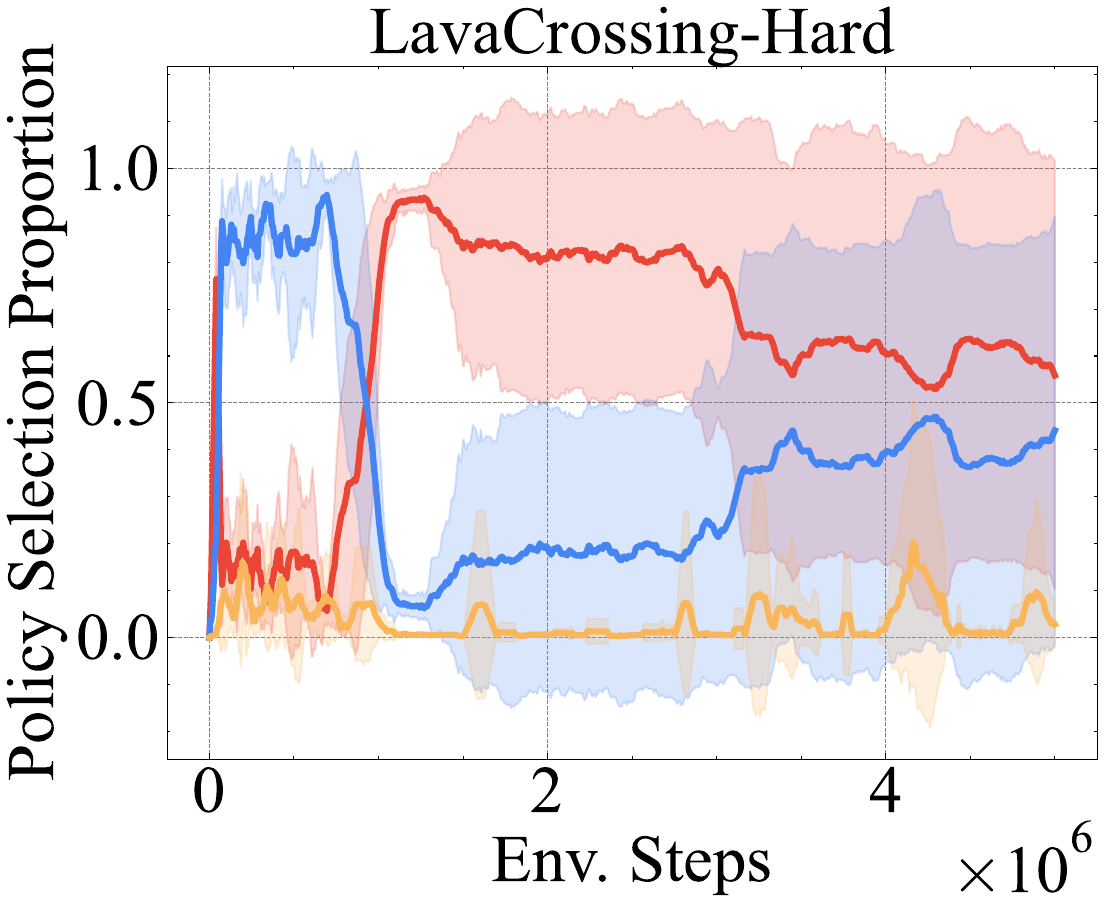}}
    \subfigure{\Description{}\includegraphics[width=0.24\textwidth]{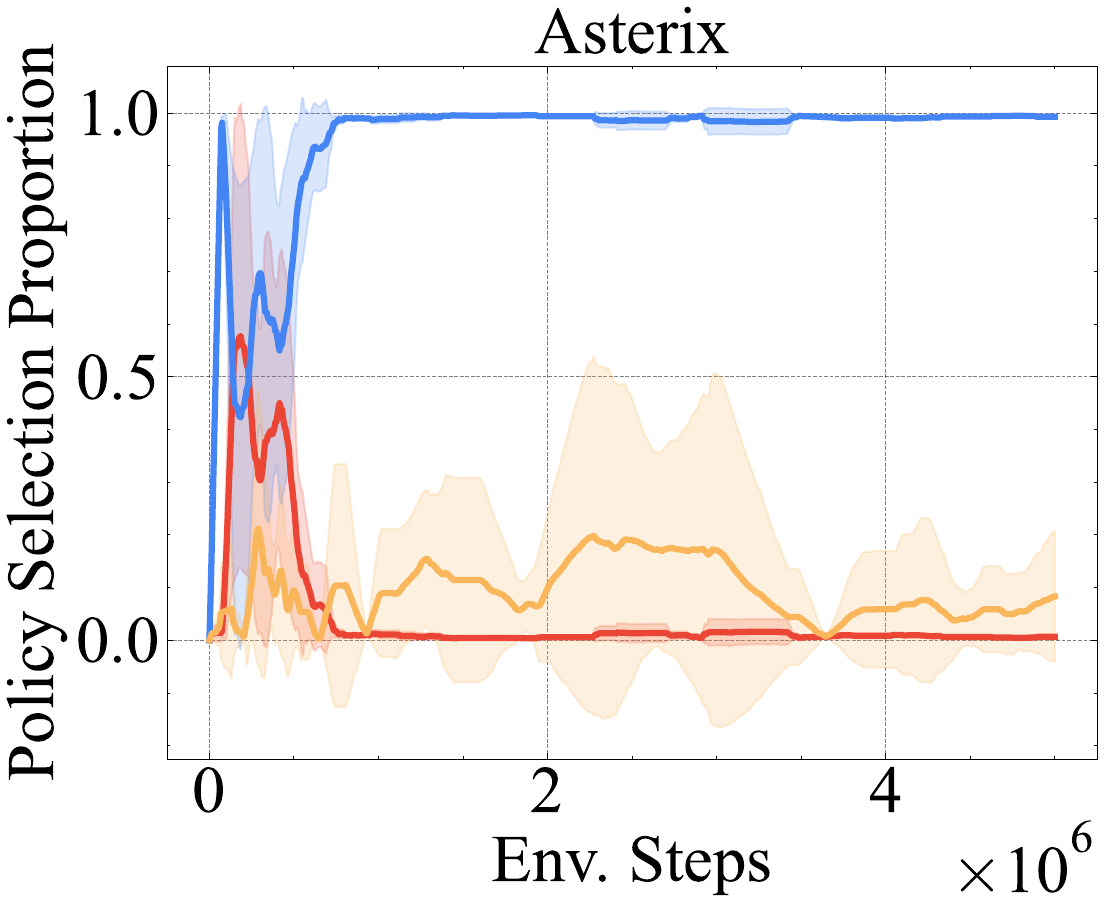}}
    \subfigure{\Description{}\includegraphics[width=0.24\textwidth]{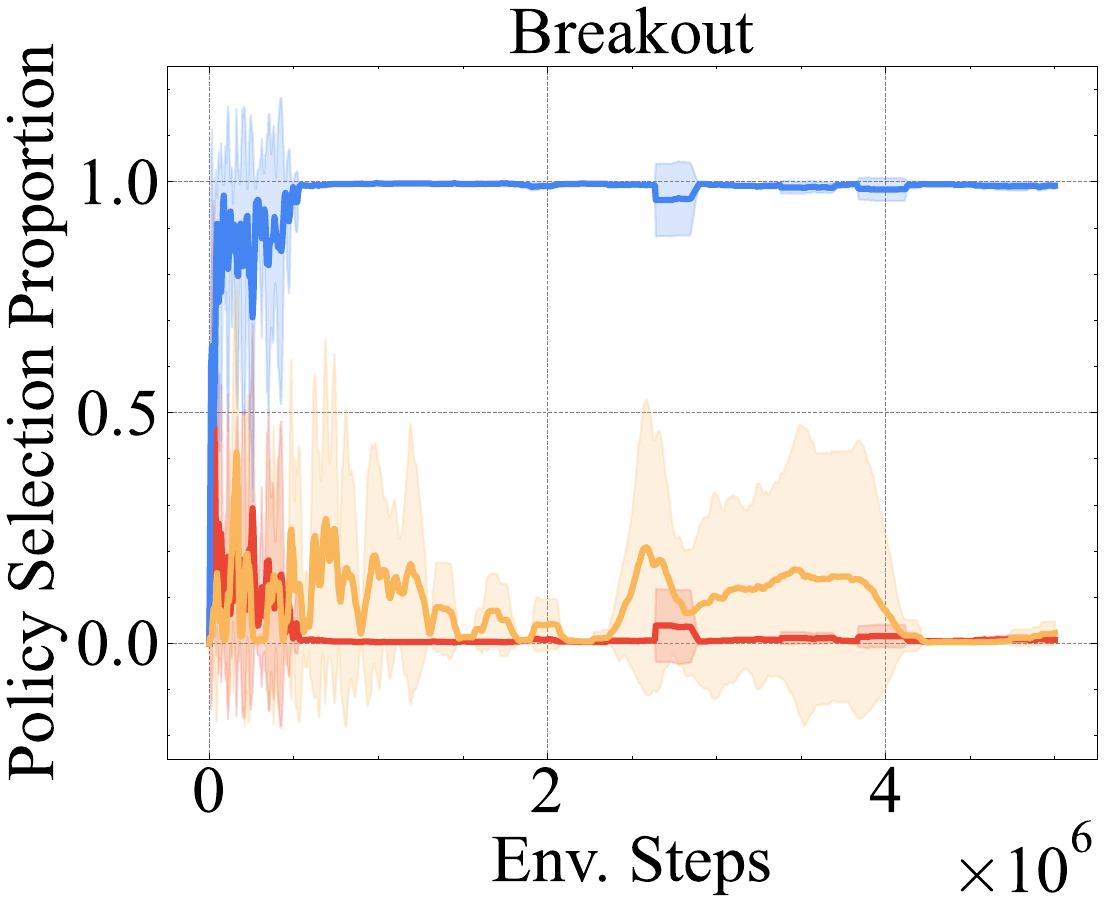}}
    \subfigure{\Description{}\includegraphics[width=0.24\textwidth]{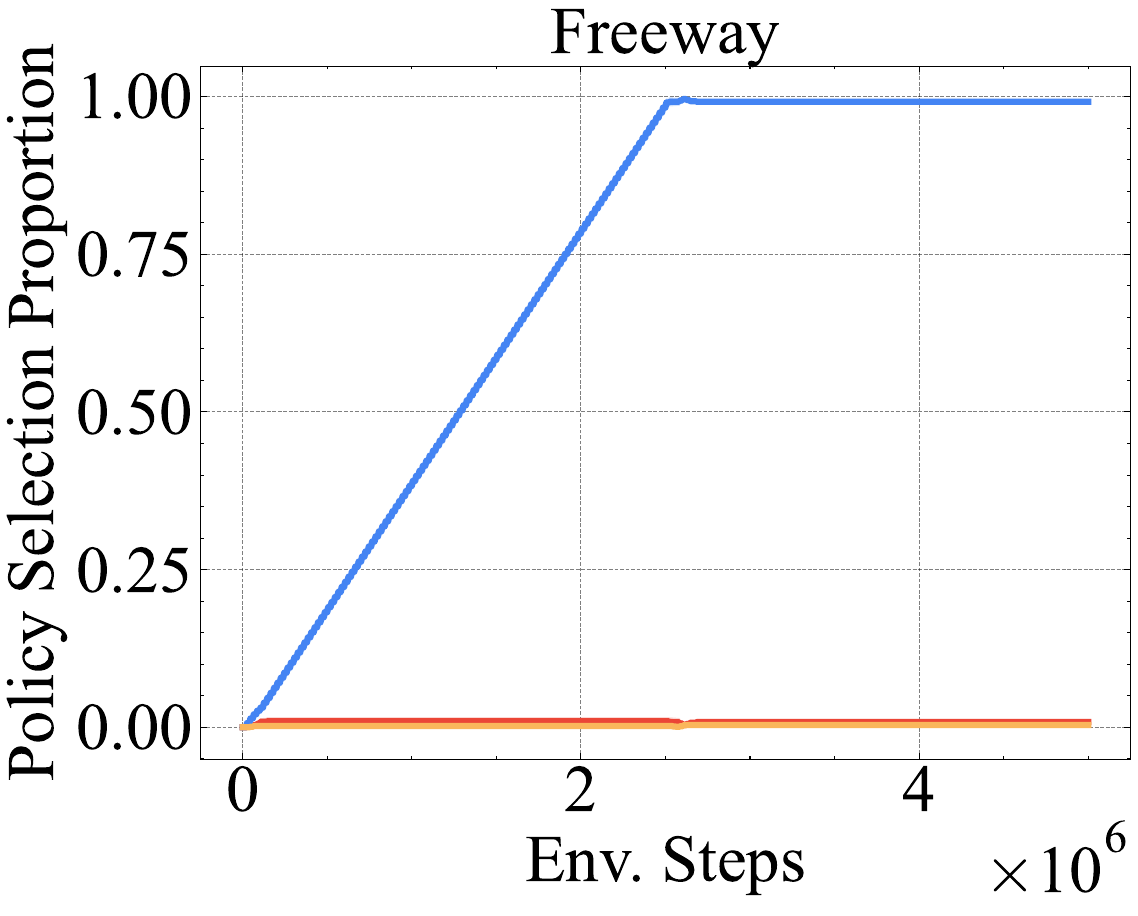}}
    \subfigure{\Description{}\includegraphics[width=0.24\textwidth]{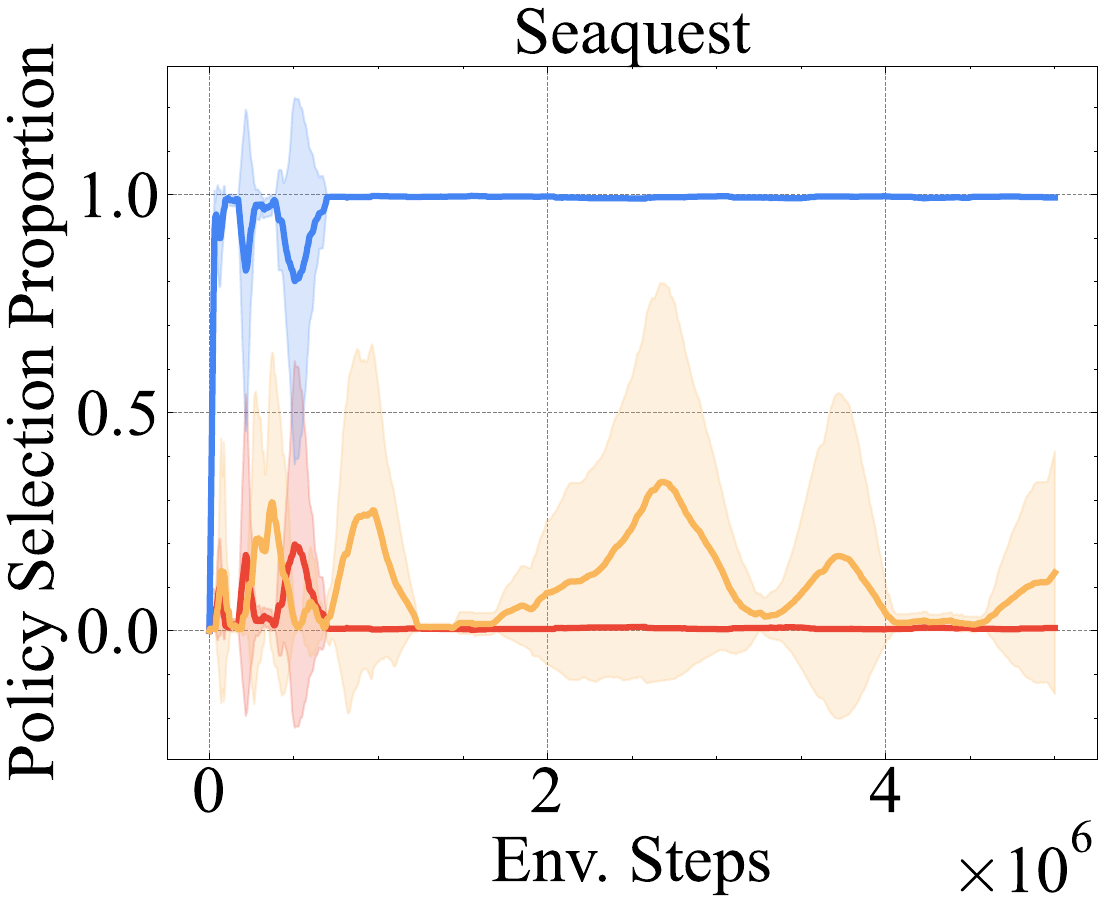}}
    \subfigure{\Description{}\includegraphics[width=0.24\textwidth]{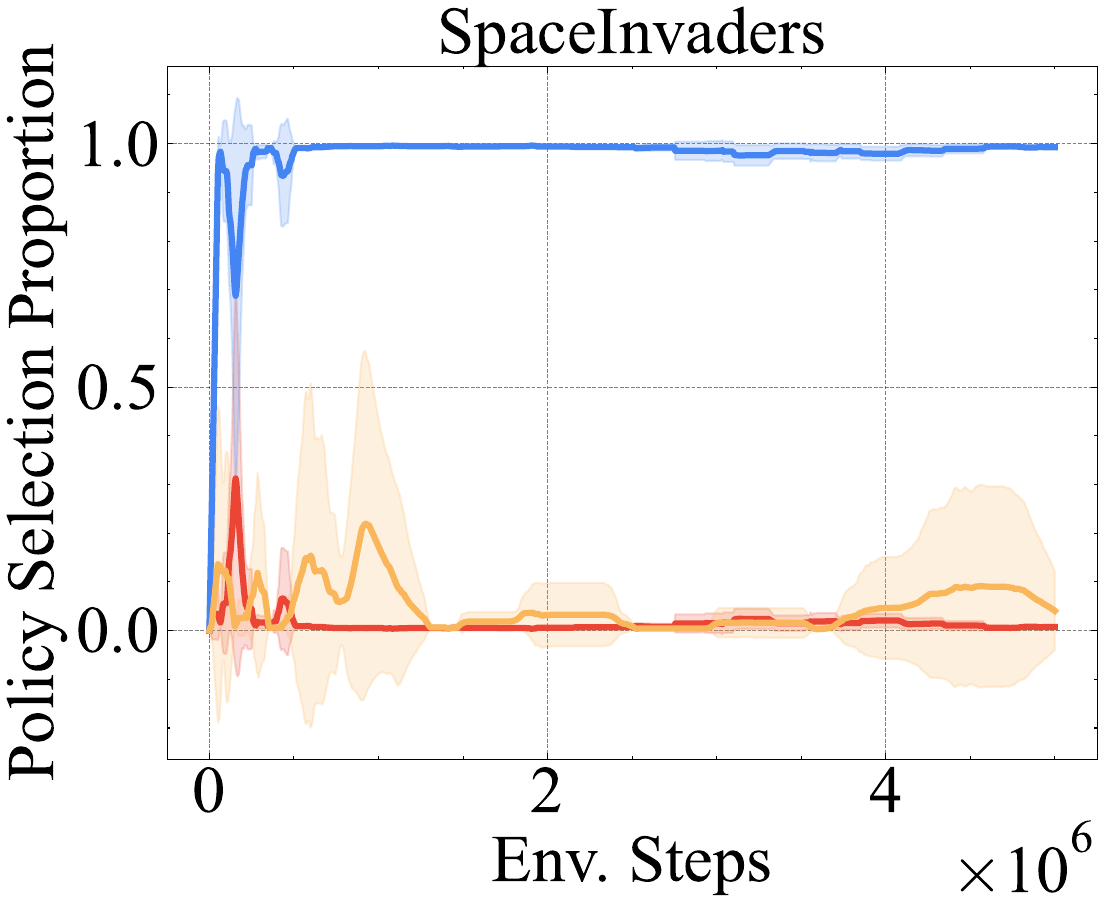}}
    \vskip -0.1in
    \caption{
    Policy selection proportions during learning on all environments. Policy $\pi_{\textit{cor}}$ play more important role in simple and dense reward environments to get corrective feedback and correct biased estimation. In hard exploration environments, the two kinds of polices $\pi_{cor}$ and $\pi_{cov}$ interleave and result in a more intricate selection pattern.
    }
    % \vskip -0.2in
    \label{fig:Policy selection proportions}
\end{center}
\end{figure*}

In~\cref{fig:Policy selection proportions}, We show the selection proportions of the two kinds of polices in the sliding-window during the learning process. 
We can find, in all MinAtar environment, the most frequently selected policy are always $\pi_{\textit{cor}}$.
It means following the exploit mode at some states and exploring some overestimated action is enough to get good performance.
This indicates, in dense reward environment, there is no need to put much effort to discover hard explored rewards, it is usually more efficient to find more low-hanging-fruit rewards.
In MiniGrid environments, the policy selection pattern is more complicated.
The state-novelty exploration plays more important role in some environments such as LavaCrossing-Hard.
The two types of policies interleave, resulting in a more intricate selection pattern.
Our meta-controller parallels the principles of depth-first search (DFS) and breadth-first search (BFS). When encountering positive rewards, our approach adopts a depth-first exploration, delving deeper into the discovered areas for further exploration. Conversely, in the absence of immediate positive feedback, we shift towards a breadth-first strategy, exploring widely in search of promising areas.

\subsubsection{The performance of polices in the policy set}
\begin{figure*}[htbp]
    % \vskip -0.1in
\begin{center}
    \subfigure{\Description{}\includegraphics[width=0.5\textwidth]{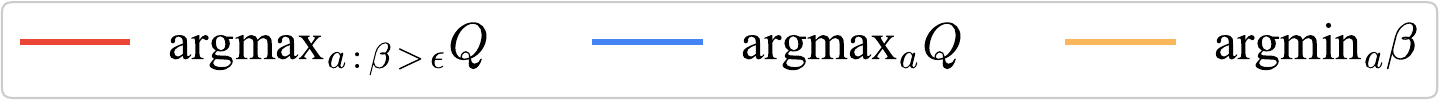}}
    \vskip -0.03in
    \subfigure{\Description{}\includegraphics[width=0.24\textwidth]{pic/3_basic_policies/DoorKey_ablation_3_basic_policies.pdf}}
    \subfigure{\Description{}\includegraphics[width=0.24\textwidth]{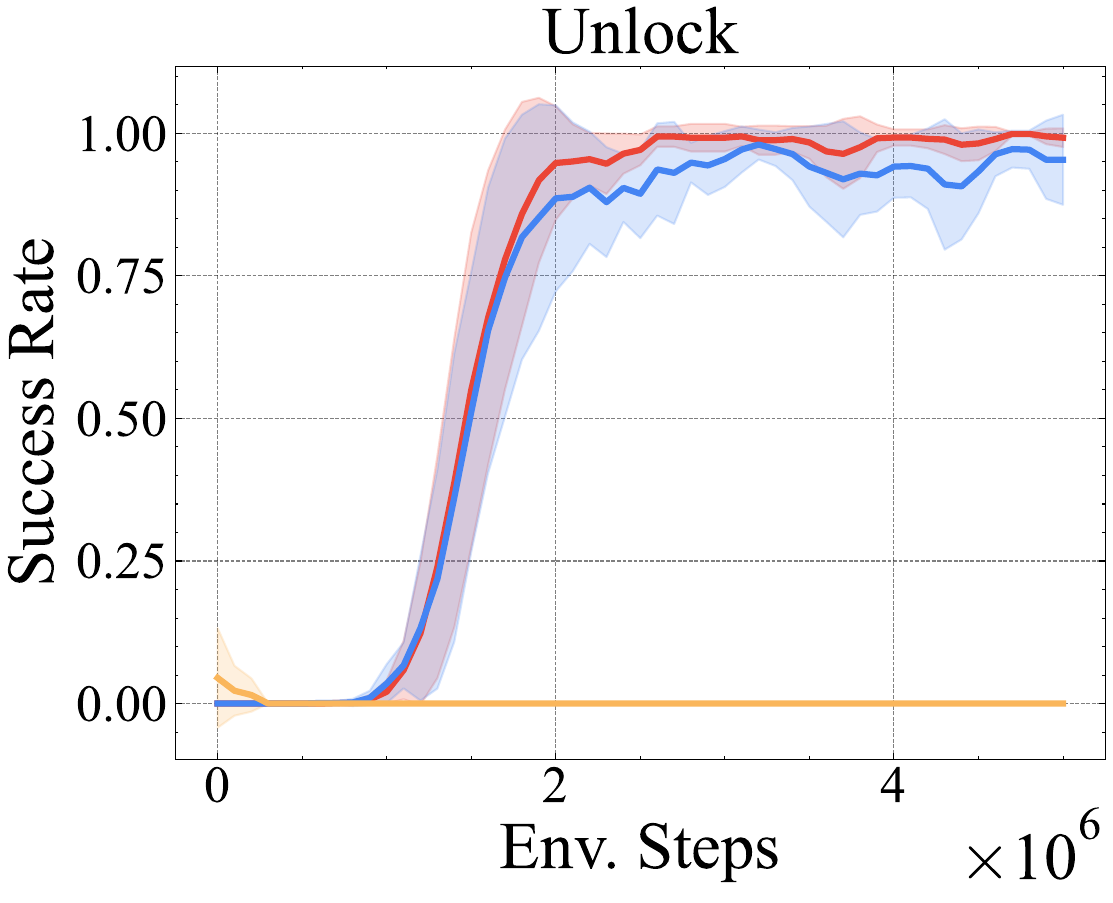}}
    \subfigure{\Description{}\includegraphics[width=0.24\textwidth]{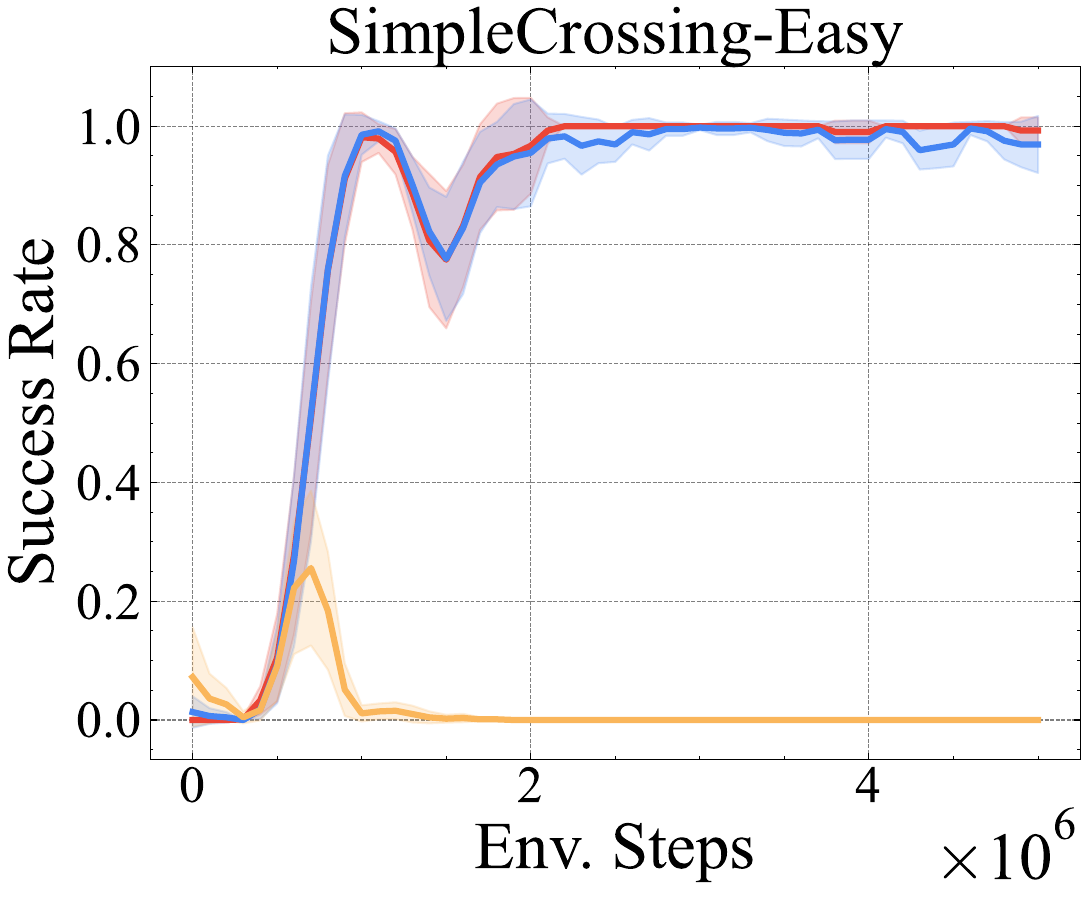}}
    \subfigure{\Description{}\includegraphics[width=0.24\textwidth]{pic/3_basic_policies/SimpleCrossing-Hard_ablation_3_basic_policies.pdf}}
    \subfigure{\Description{}\includegraphics[width=0.24\textwidth]{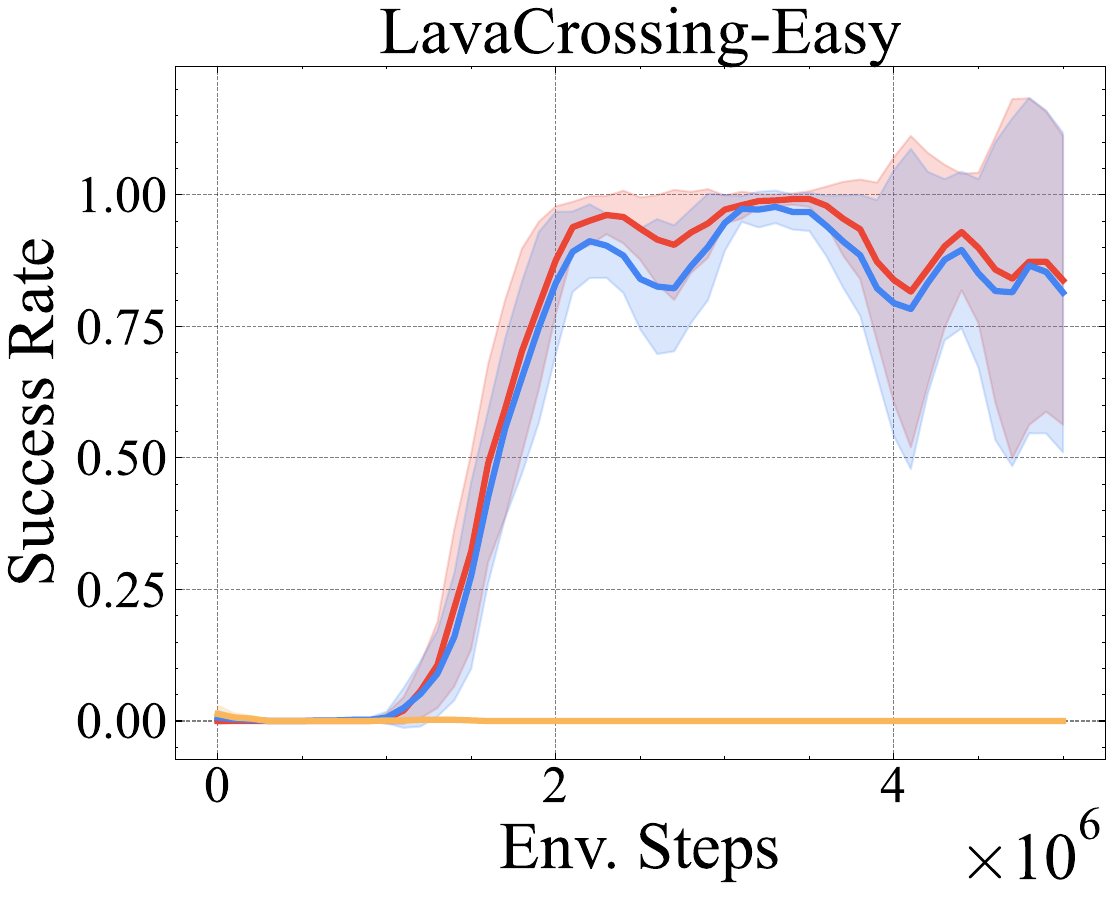}}
    \subfigure{\Description{}\includegraphics[width=0.24\textwidth]{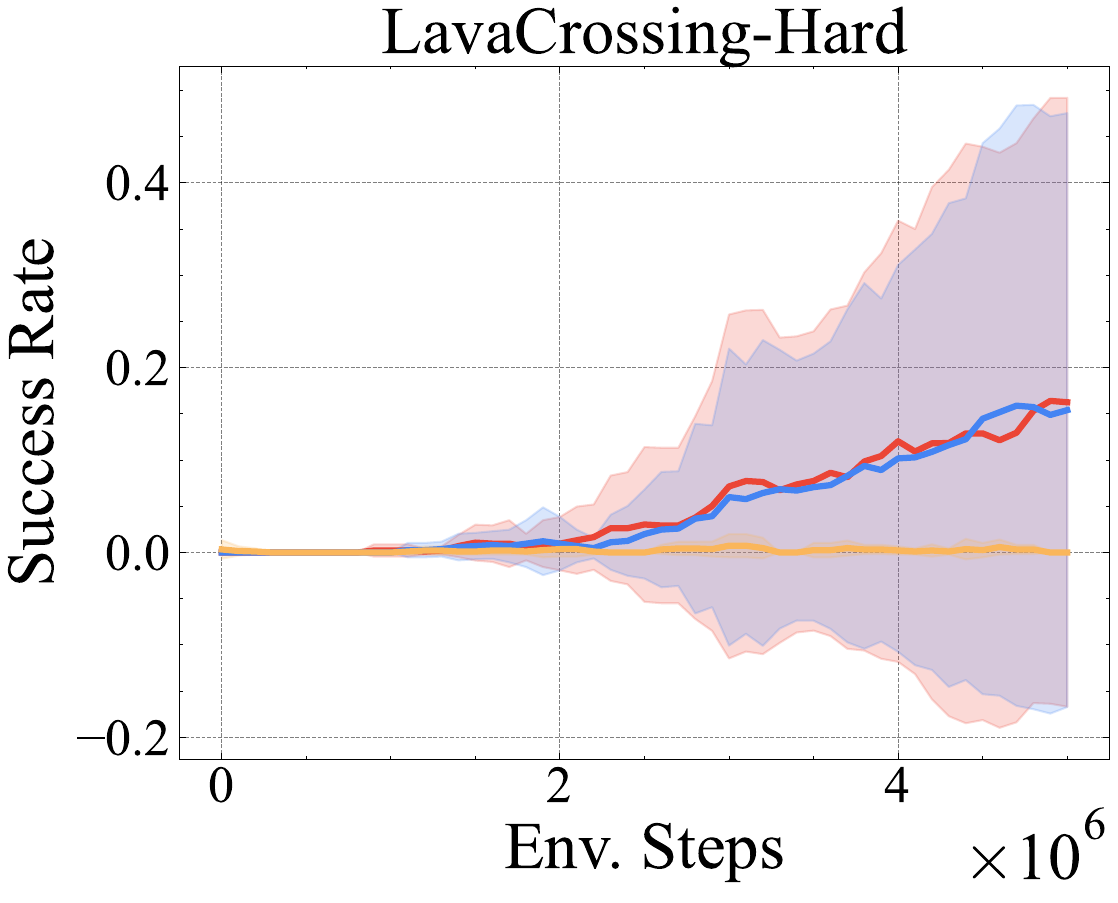}}
    \subfigure{\Description{}\includegraphics[width=0.24\textwidth]{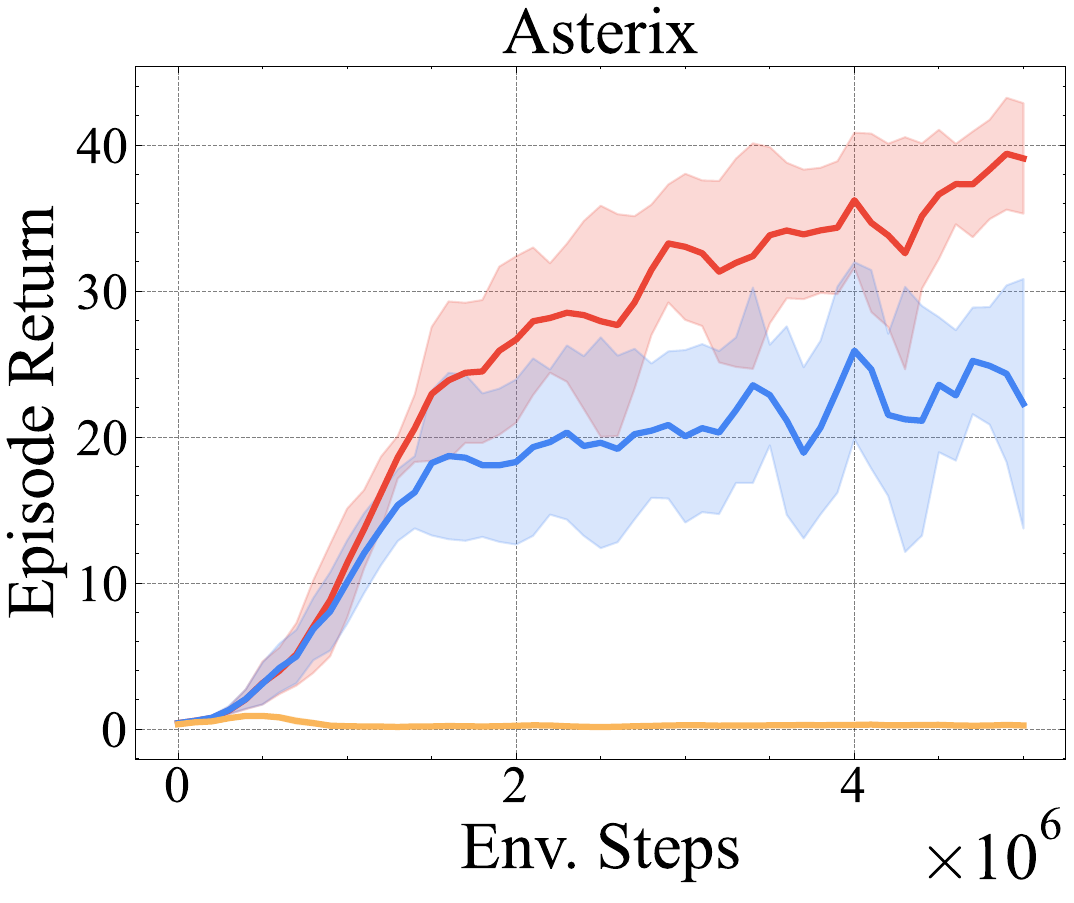}}
    \subfigure{\Description{}\includegraphics[width=0.24\textwidth]{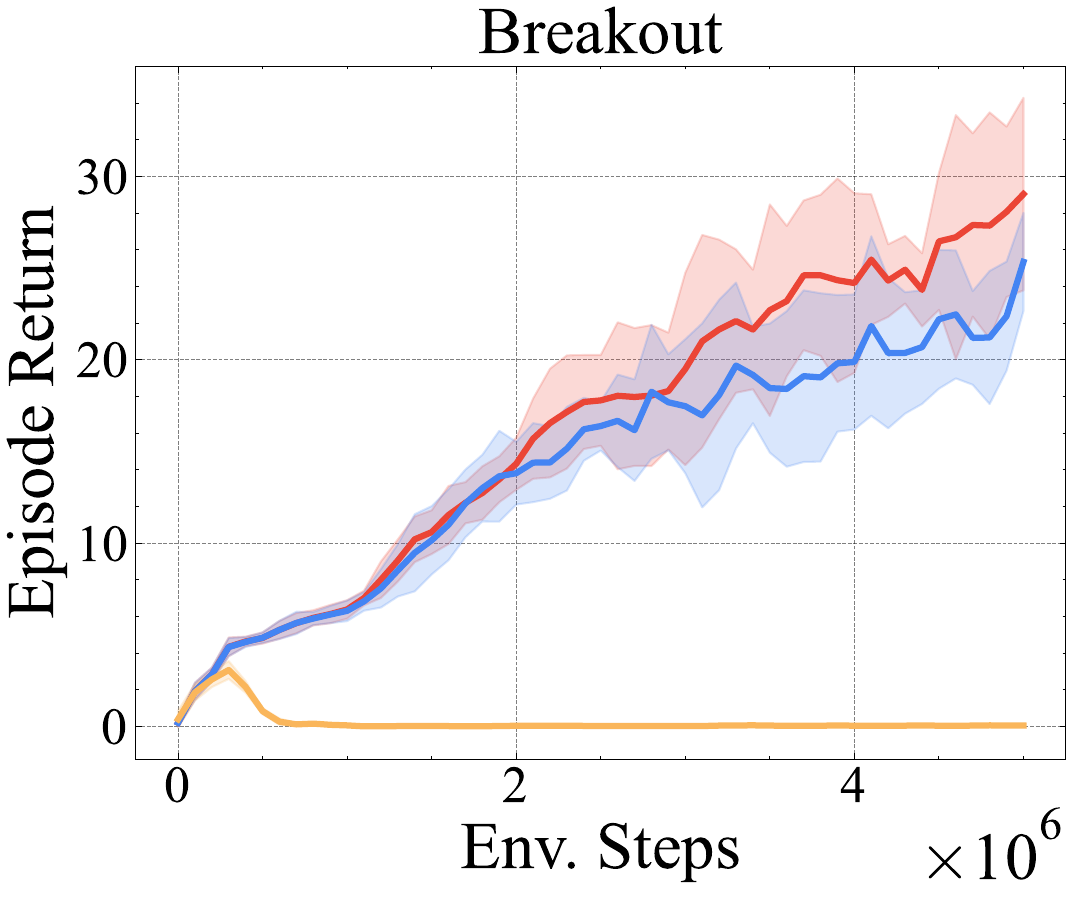}}
    \subfigure{\Description{}\includegraphics[width=0.24\textwidth]{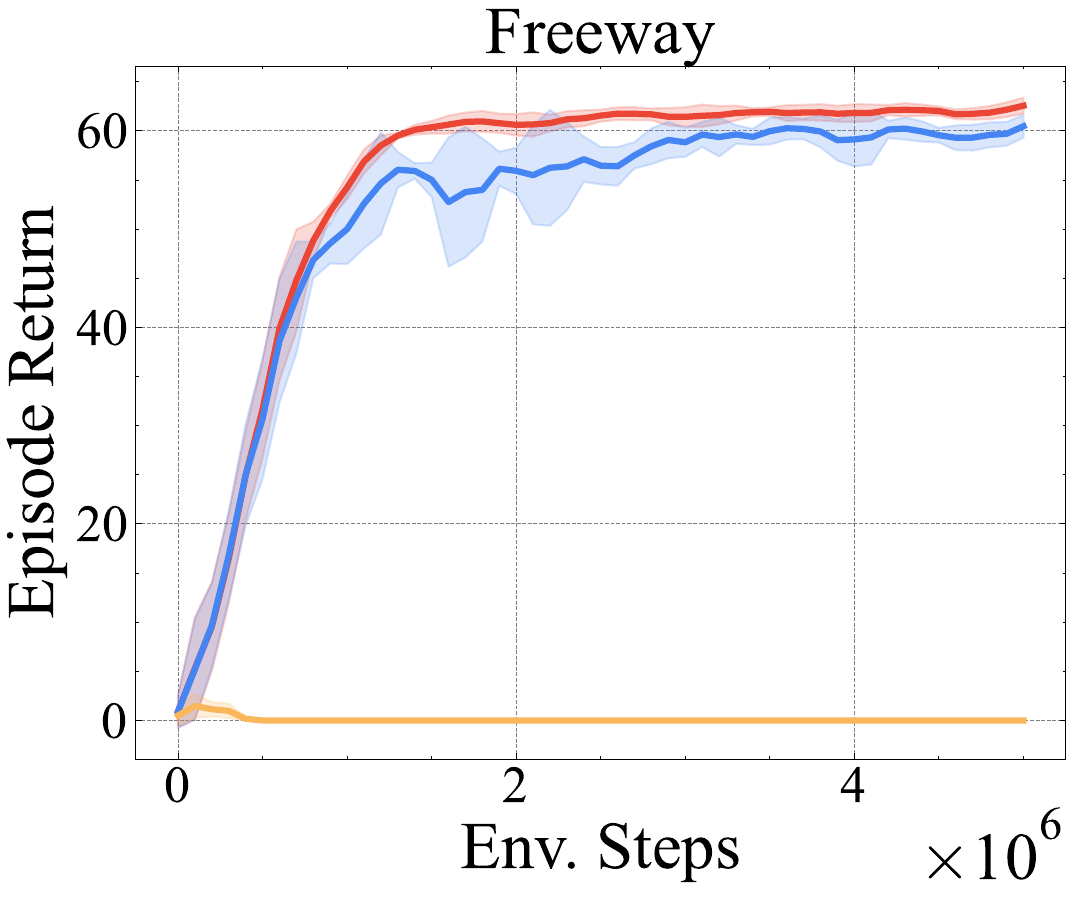}}
    \subfigure{\Description{}\includegraphics[width=0.24\textwidth]{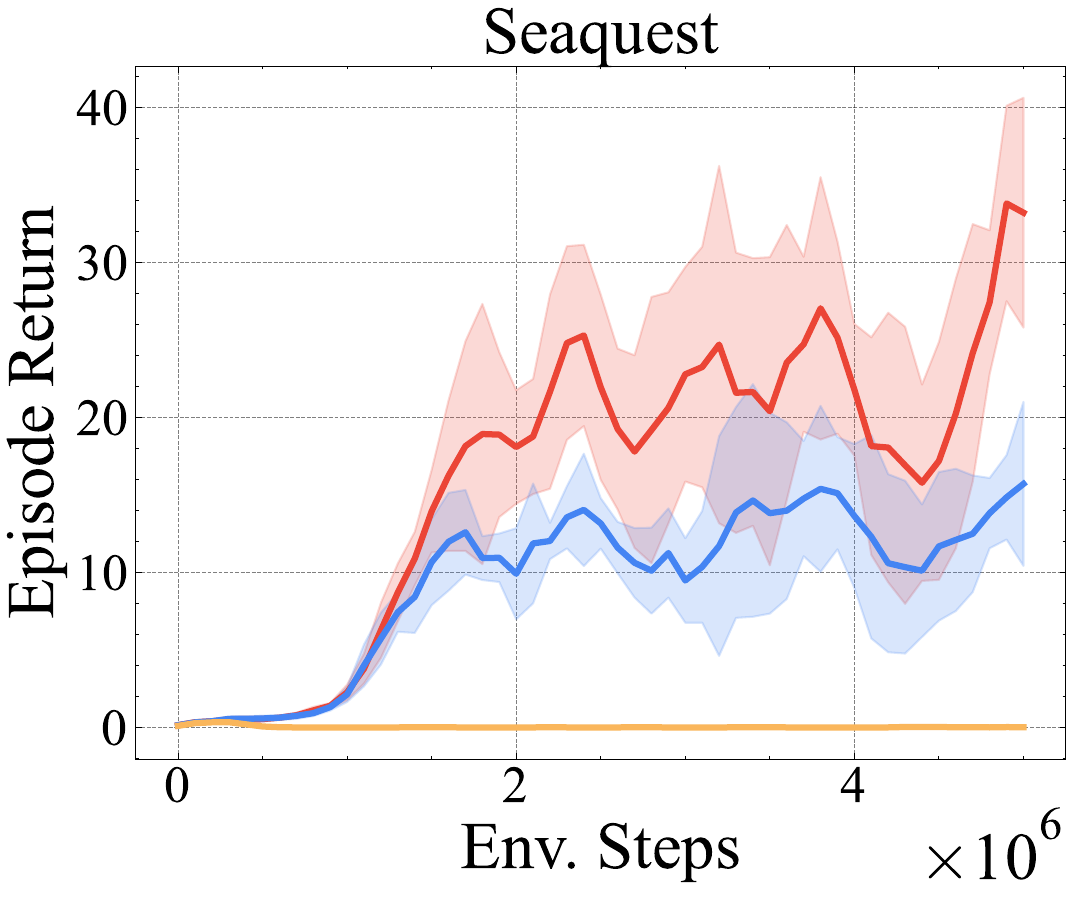}}
    \subfigure{\Description{}\includegraphics[width=0.24\textwidth]{pic/3_basic_policies/SpaceInvaders_ablation_3_basic_policies.pdf}}
    \vskip -0.15in
    \caption{
    The performance of the three basic polices during learning. $\arg\min_a \beta$ learns nothing since it does not care about performance. $\arg\max_a Q_{\textit{mask}}$ chooses in-sample greedy actions and performs the best. $\arg\max_a Q$ take greedy actions among the whole action space and may take overestimated actions. Its performance is closed to but a little worse than $\arg\max_a Q_{\textit{mask}}$.
    }
    % \vskip -0.2in
    \label{fig:3 basic polices}
\end{center}
\end{figure*}

We have three basic functions in our method, one may curious about the performance of the polices
derived from the three functions.
We show the performance of them in~\cref{fig:3 basic polices}. 
The policy $\arg\min_a \beta$ always takes actions with the least probability, it does not care about the performance thus learns nothing. 
The policy $\arg\max_a Q_{\textit{mask}}$ chooses greedy actions that is well-supported in the replay memory and performs the best, which is what we expected. 
The policy $\arg\max_a Q$ takes greedy actions among the whole action space. It may take overestimated actions at some states thus the performance is not as stable as $\arg\max_a Q_{\textit{mask}}$.

We can also find in some environments like SimpleCrossing-Easy and SimpleCrossing-Hard, $\arg\max_a Q$ performs similar as $\arg\max_a Q_{\textit{mask}}$. 
This indicates the space has been fully explored and there is little estimation bias in function $Q$.
In contrast, in some other environments like Asterix and SpaceInvaders, there is a large gap between $\arg\max_a Q$ and $\arg\max_a Q_{\textit{mask}}$. This indicates there is a lot of underexplored overestimated actions waiting for correcting.

\subsubsection{Learning two separate Q functions}
\begin{figure*}[htbp]
    % \vskip -0.1in
\begin{center}
    \subfigure{\Description{}\includegraphics[width=0.25\textwidth]{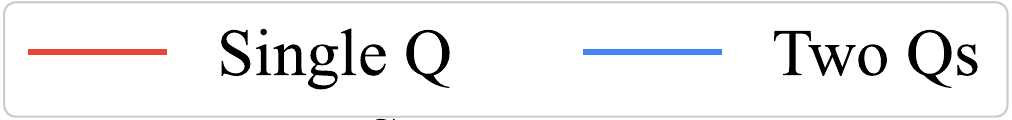}}
    \vskip -0.03in
    \subfigure{\Description{}\includegraphics[width=0.24\textwidth]{pic/learn2Q/DoorKey_ablation_learn2Q.pdf}}
    \subfigure{\Description{}\includegraphics[width=0.24\textwidth]{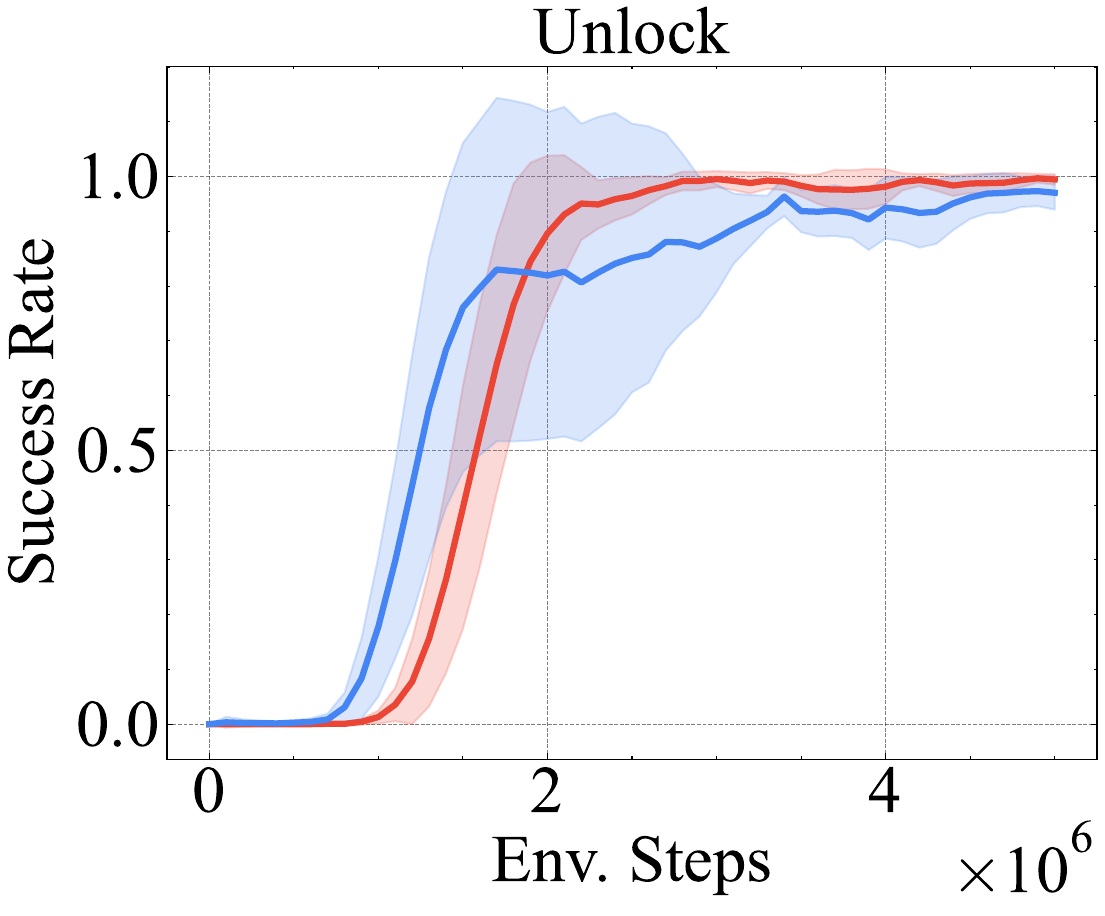}}
    \subfigure{\Description{}\includegraphics[width=0.24\textwidth]{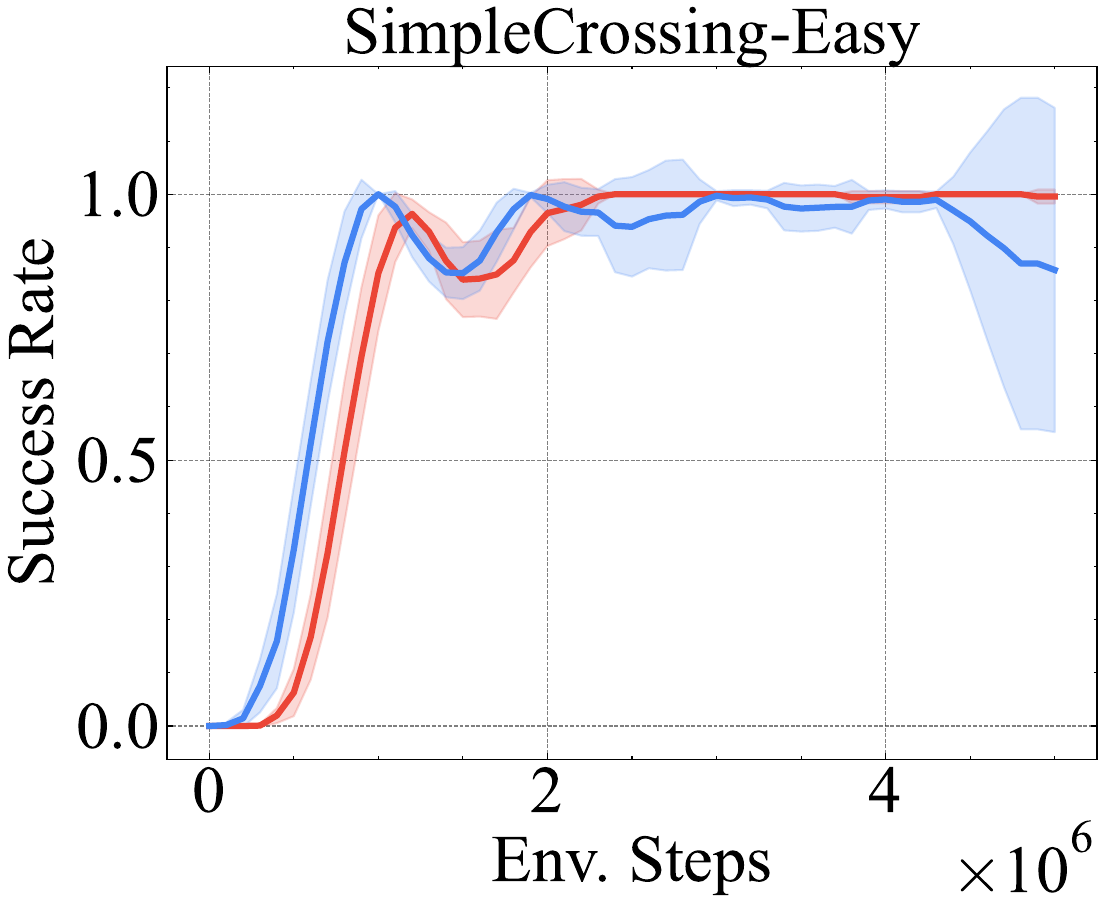}}
    \subfigure{\Description{}\includegraphics[width=0.24\textwidth]{pic/learn2Q/SimpleCrossing-Hard_ablation_learn2Q.pdf}}
    \subfigure{\Description{}\includegraphics[width=0.24\textwidth]{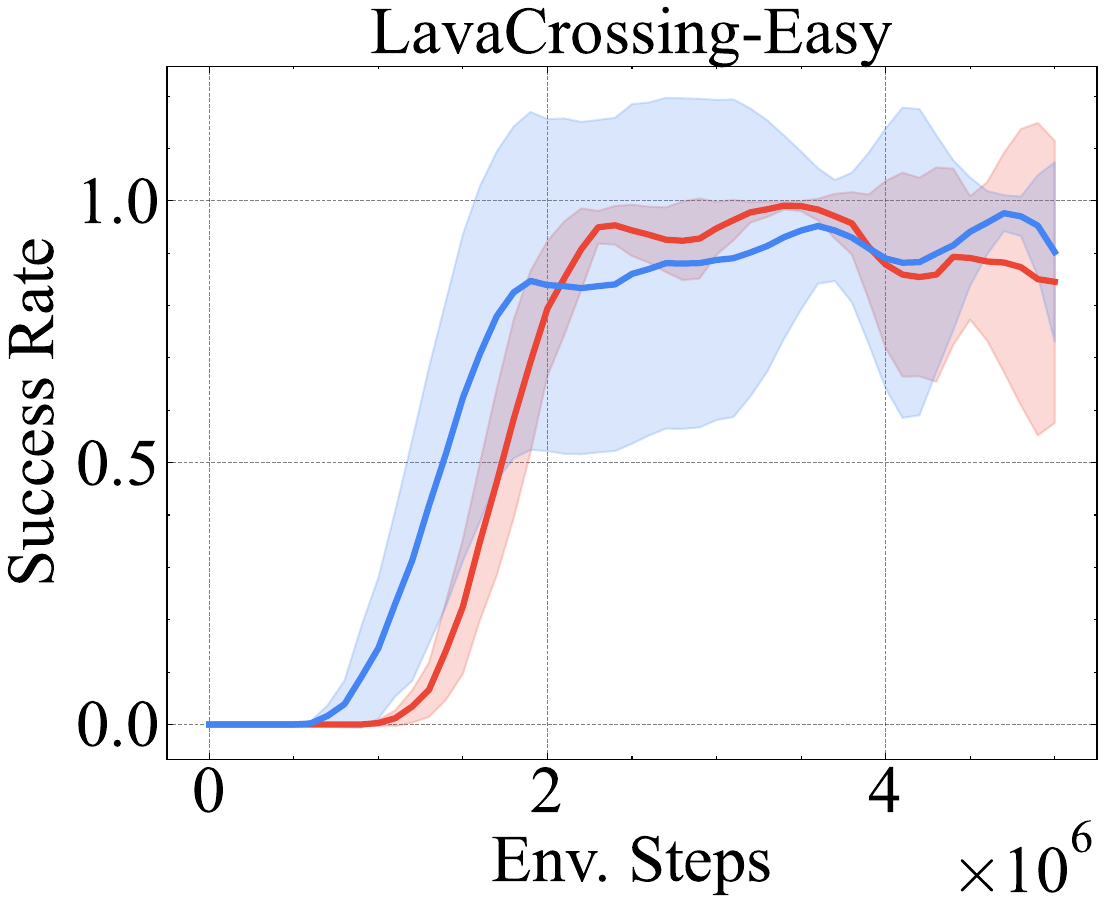}}
    \subfigure{\Description{}\includegraphics[width=0.24\textwidth]{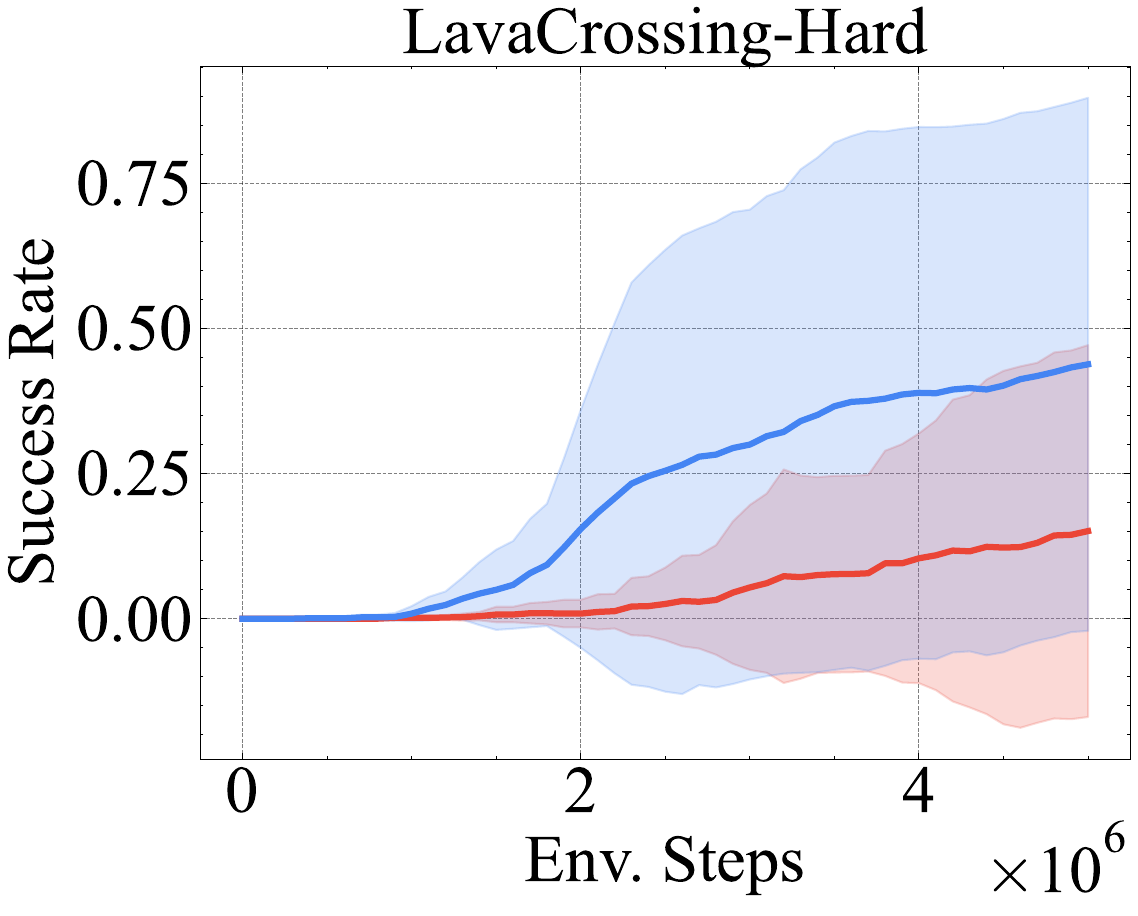}}
    \subfigure{\Description{}\includegraphics[width=0.24\textwidth]{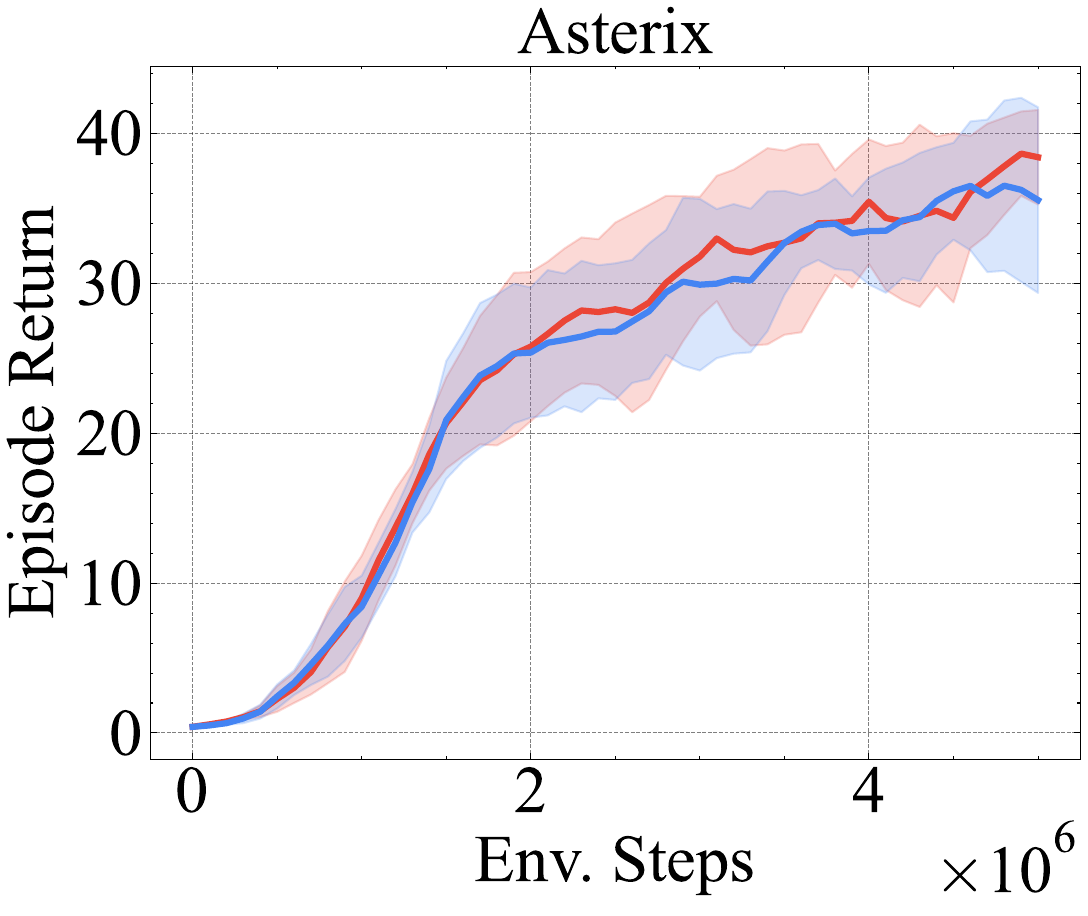}}
    \subfigure{\Description{}\includegraphics[width=0.24\textwidth]{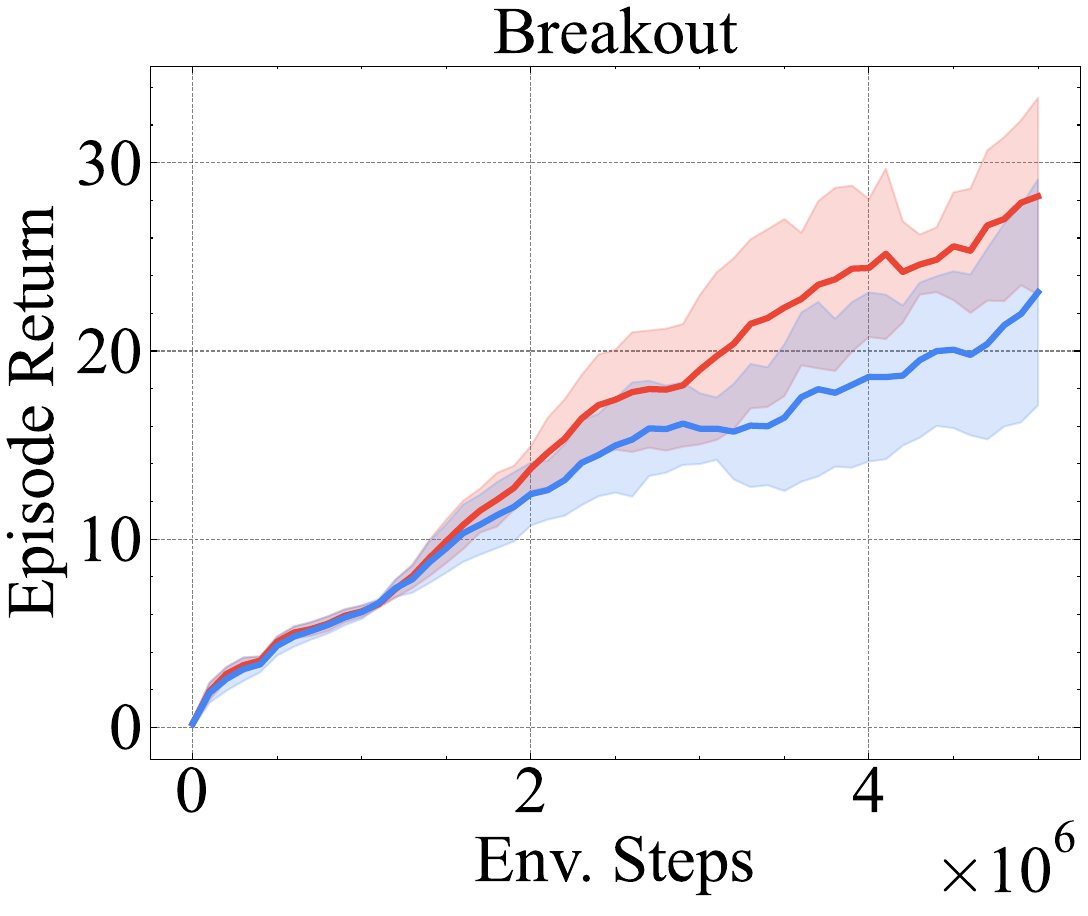}}
    \subfigure{\Description{}\includegraphics[width=0.24\textwidth]{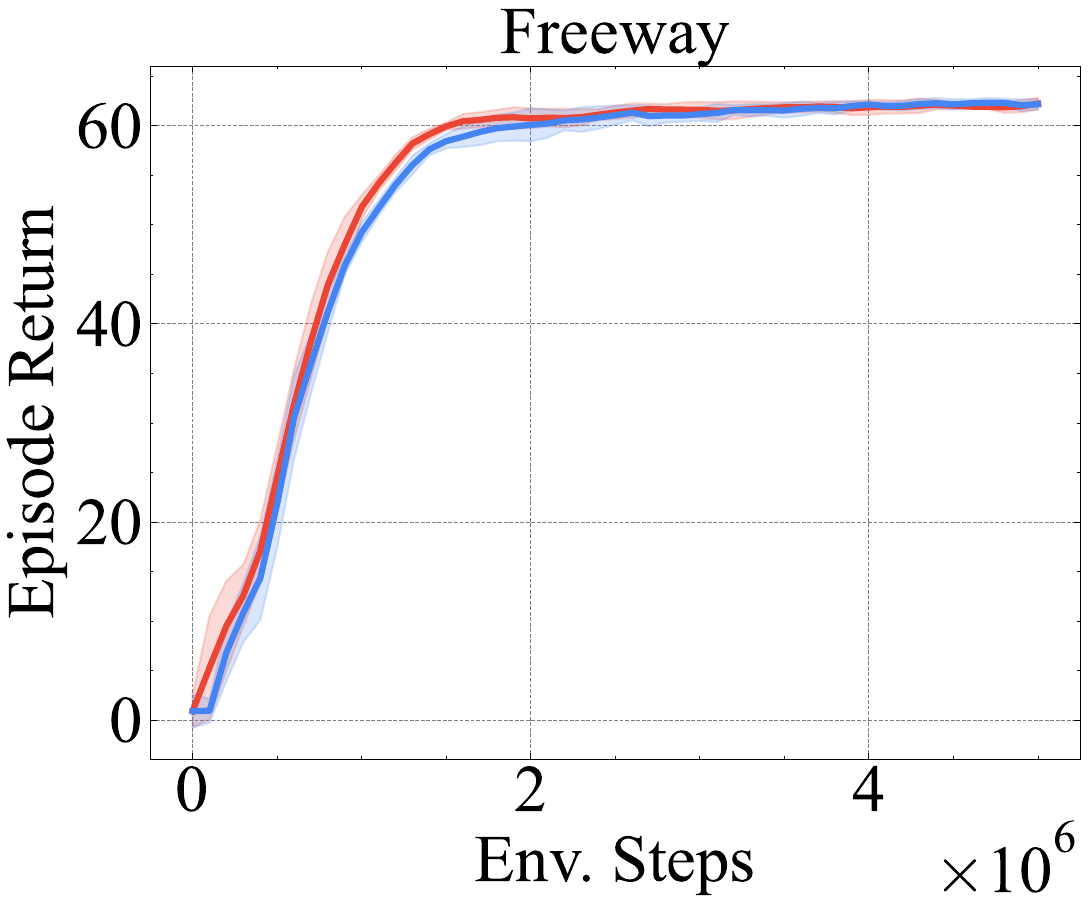}}
    \subfigure{\Description{}\includegraphics[width=0.24\textwidth]{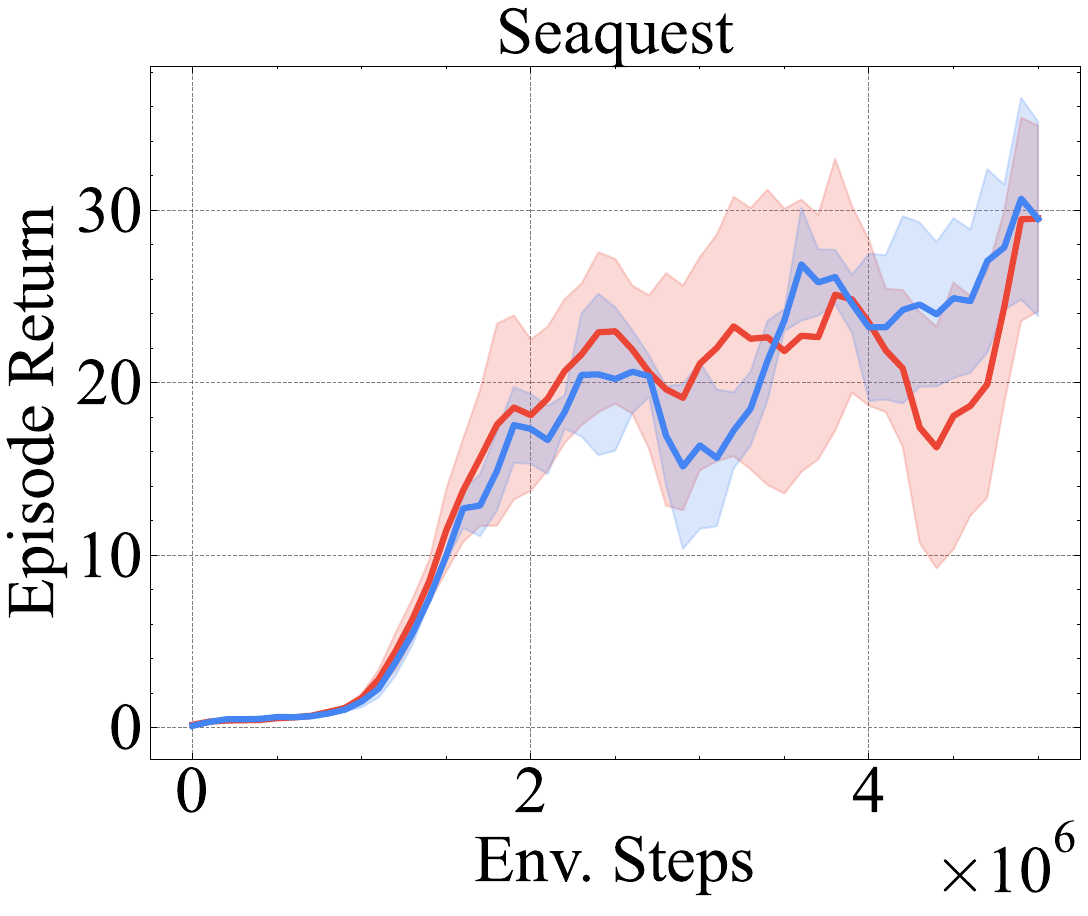}}
    \subfigure{\Description{}\includegraphics[width=0.24\textwidth]{pic/learn2Q/SpaceInvaders_ablation_learn2Q.pdf}}
    \vskip -0.1in
    \caption{
    Ablation study on learning two separate Q functions. We learn the $Q$ function with TD learning, and learning $Q_{\textit{mask}}$ with constrained TD learning and behavior function $\beta$. There is no obvious difference between the two methods, which means learning one $Q$ with constrained TD learning is enough to derive two $Q$ functions. 
    }
    % \vskip -0.2in
    \label{fig:learn 2 Q}
\end{center}
\end{figure*}

% \noindent\textbf{Learn two separate $Q$s.} 
Our method learns one $Q$ function with~\cref{eq:in distribution td learning} and obtain $Q$ and $Q_{\textit{mask}}$ from the single function.
The intuition is that though~\cref{eq:in distribution td learning} gives us a conservative estimate based on in-distribution data, it may still overestimate at unseen state-action pairs as shown in the toy example~\cref{fig:toy example,fig:another toy example}.
A more natural way is that we can maintain the update rule in DQN unchanged.
And additionally learn another $Q$ following~\cref{eq:in distribution td learning}.
In this way, we learn $Q$ function with TD learning and $Q_{\textit{mask}}$ with in-sample TD learning.
We show the ablation in~\cref{fig:learn 2 Q}.
On most of these environment, we find no big difference, which means learning one $Q$ function with~\cref{eq:in distribution td learning} is enough to get both conservative and optimistic estimation and adds less computational overhead.

\subsubsection{The influence of policy set size.}
\label{sec:influence of policy set size}
\begin{figure*}[htbp]
    % \vskip -0.1in
\begin{center}
    \subfigure{\includegraphics[width=0.9\textwidth]{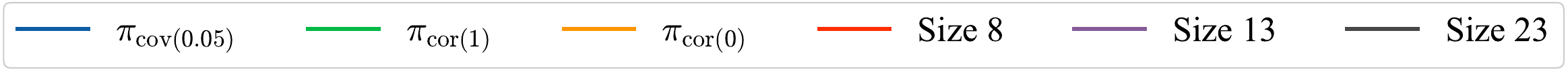}}
    \vskip -0.03in
    \subfigure{\Description{}\includegraphics[width=0.24\textwidth]{pic/policy_set_size/DoorKey_ablation_policy_set_size.pdf}}
    \subfigure{\Description{}\includegraphics[width=0.24\textwidth]{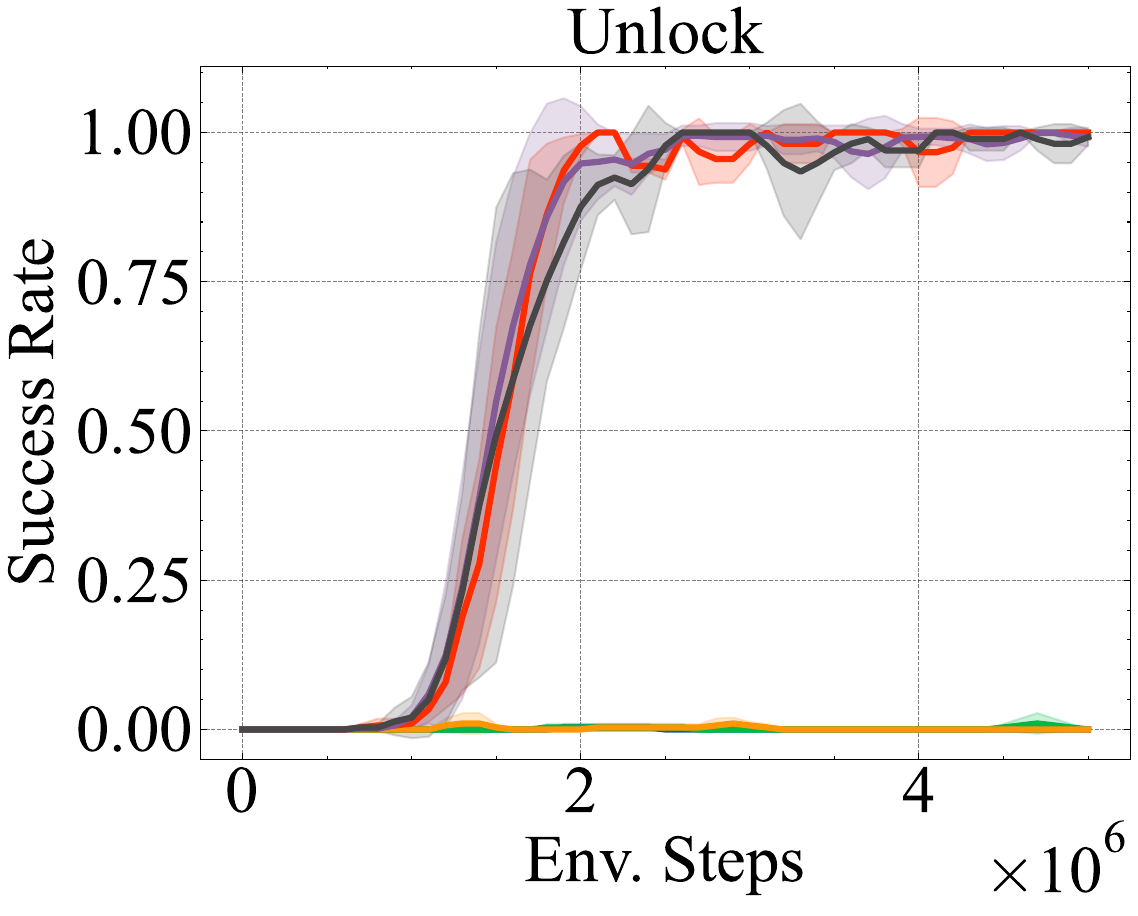}}
    \subfigure{\Description{}\includegraphics[width=0.24\textwidth]{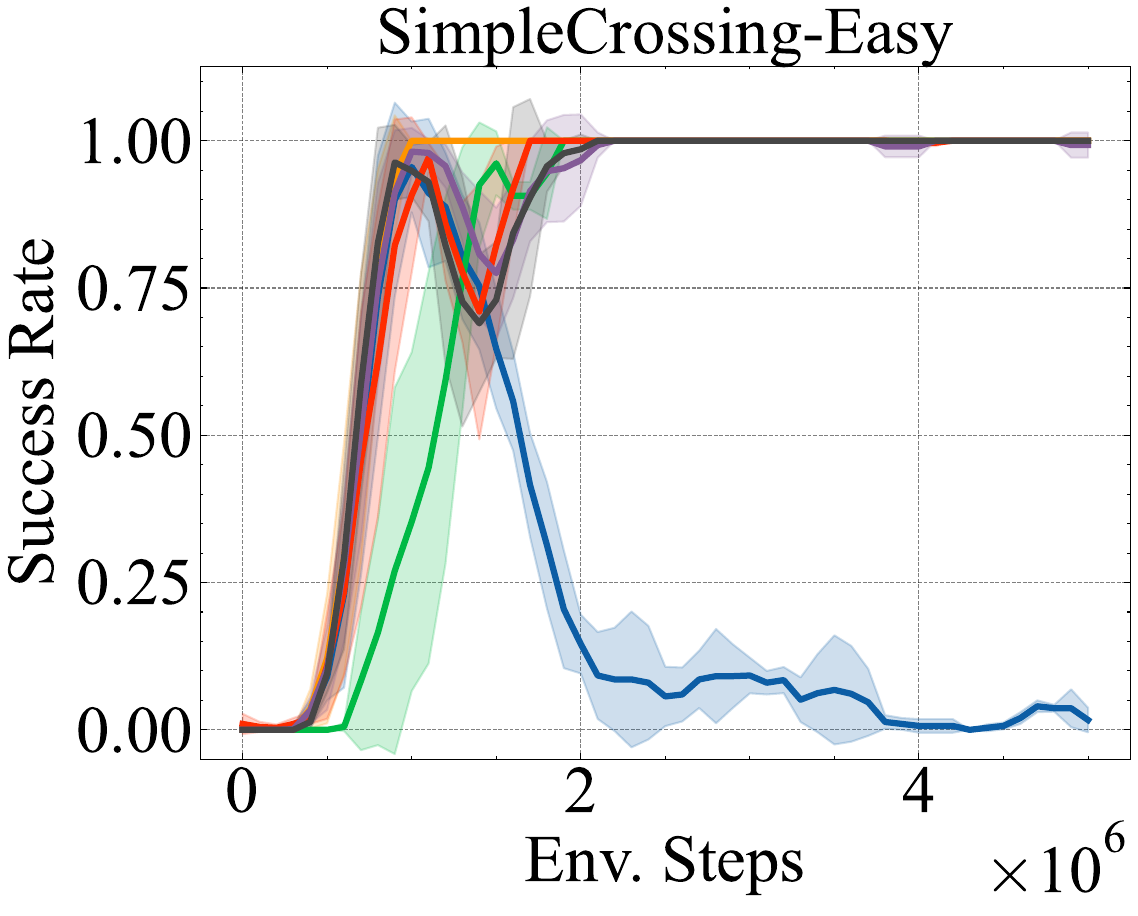}}
    \subfigure{\Description{}\includegraphics[width=0.24\textwidth]{pic/policy_set_size/SimpleCrossing-Hard_ablation_policy_set_size.pdf}}
    \subfigure{\Description{}\includegraphics[width=0.24\textwidth]{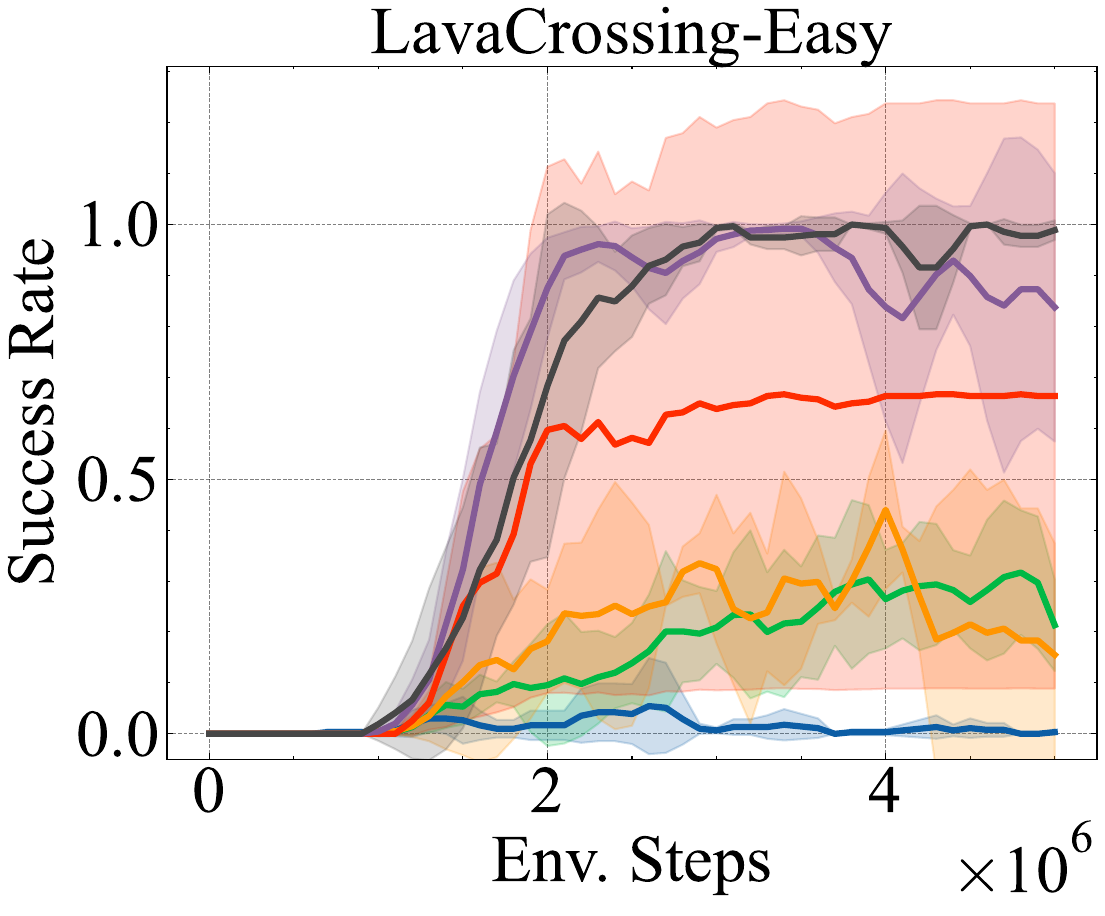}}
    \subfigure{\Description{}\includegraphics[width=0.24\textwidth]{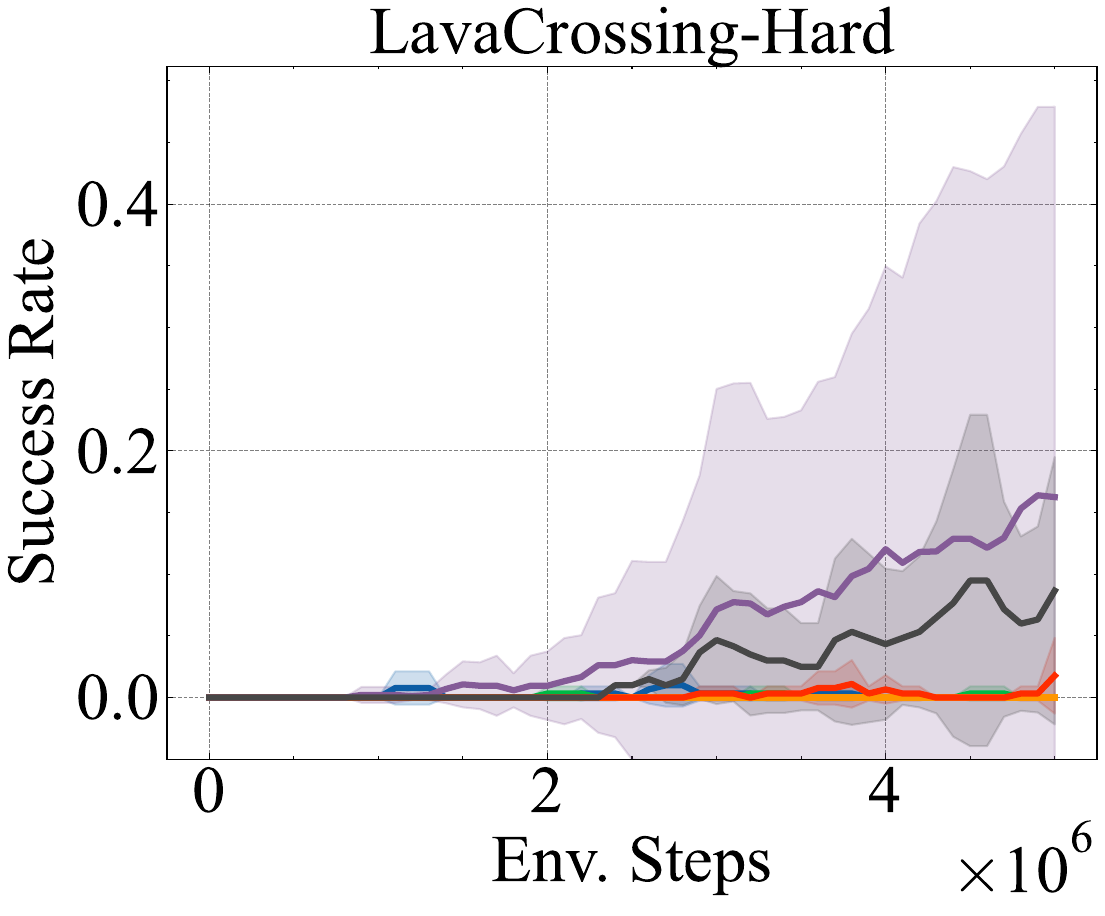}}
    \subfigure{\Description{}\includegraphics[width=0.24\textwidth]{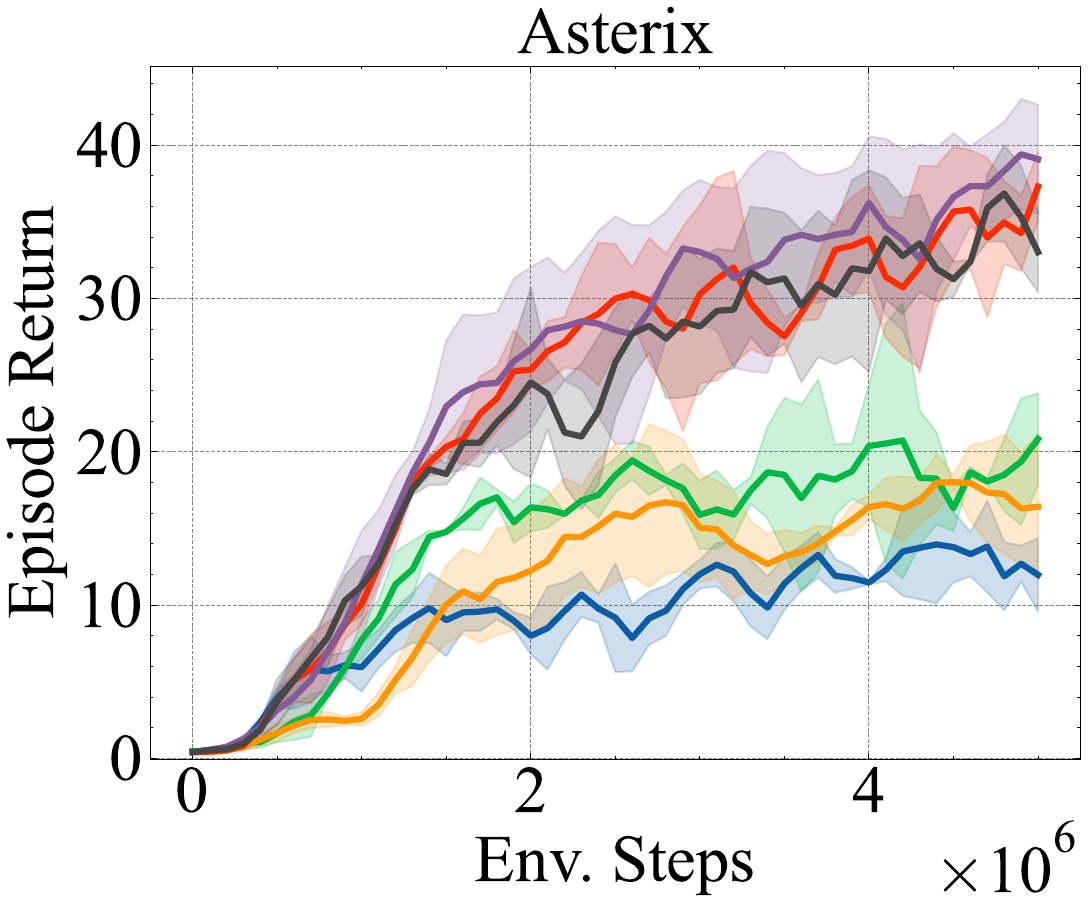}}
    \subfigure{\Description{}\includegraphics[width=0.24\textwidth]{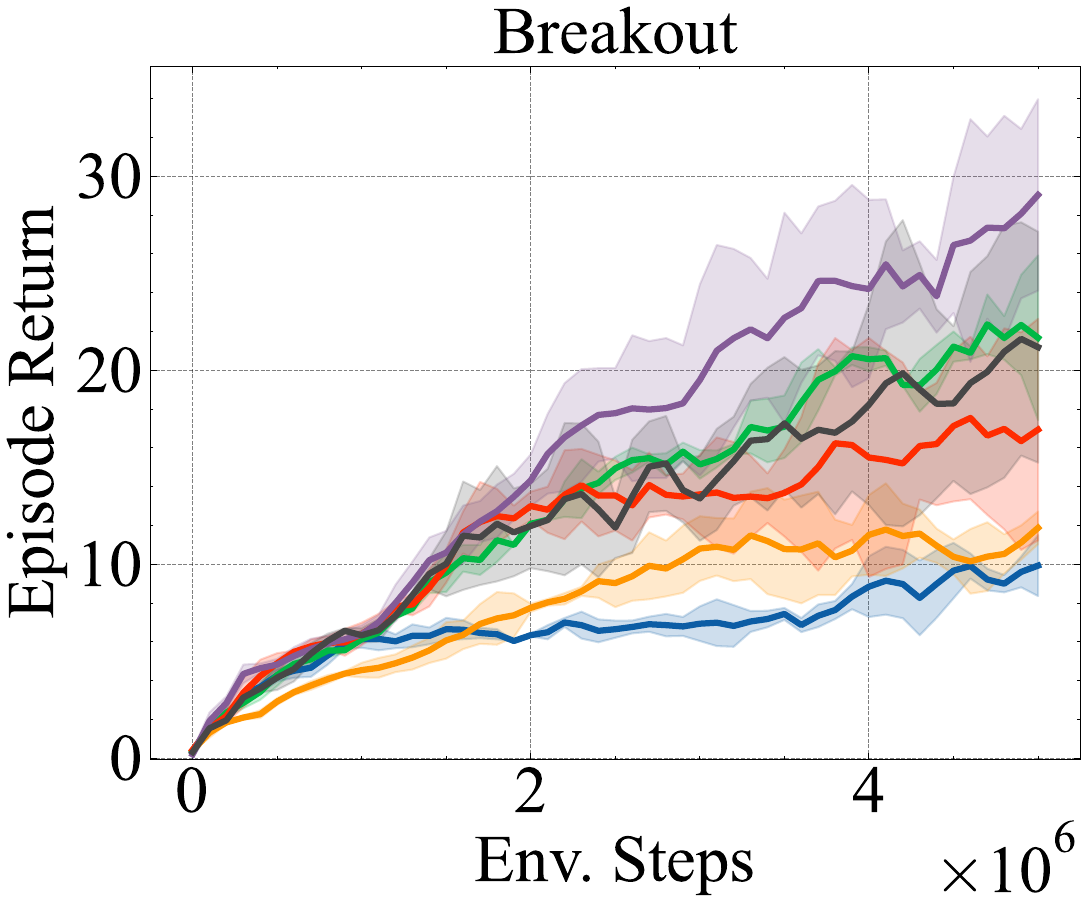}}
    \subfigure{\Description{}\includegraphics[width=0.24\textwidth]{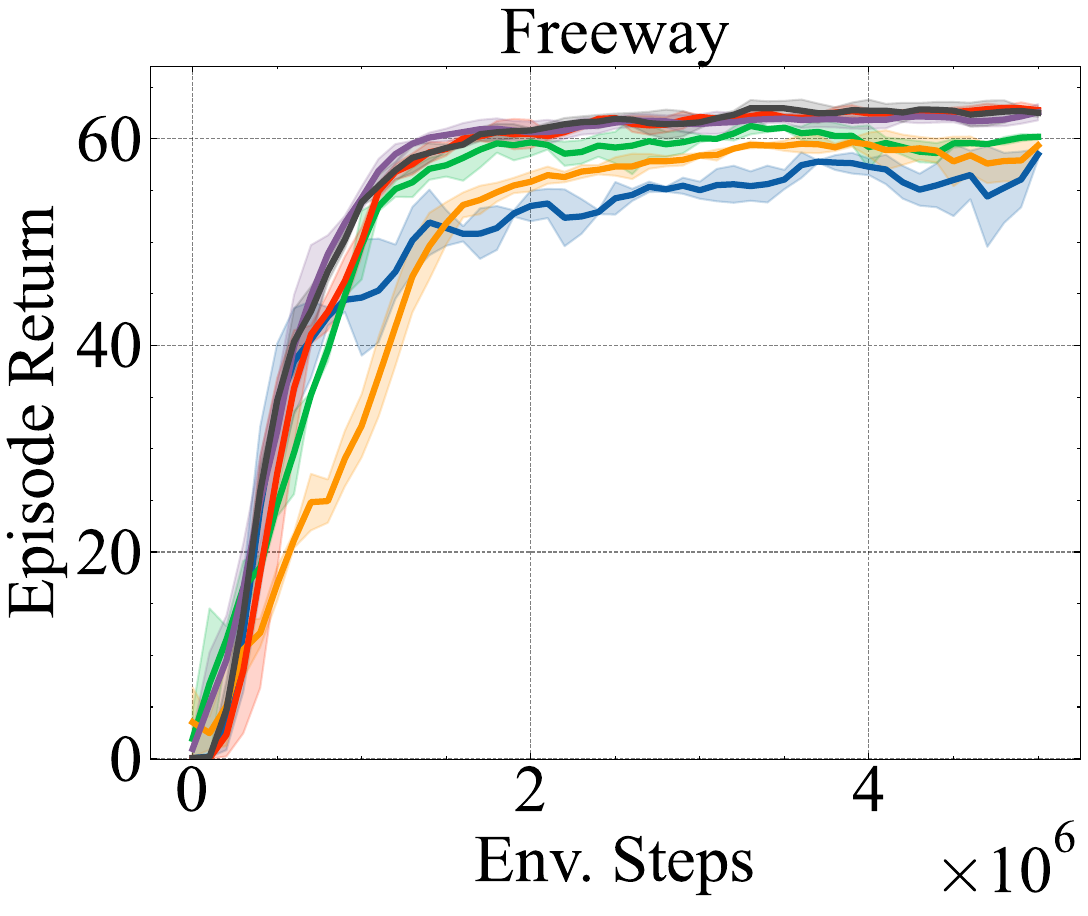}}
    \subfigure{\Description{}\includegraphics[width=0.24\textwidth]{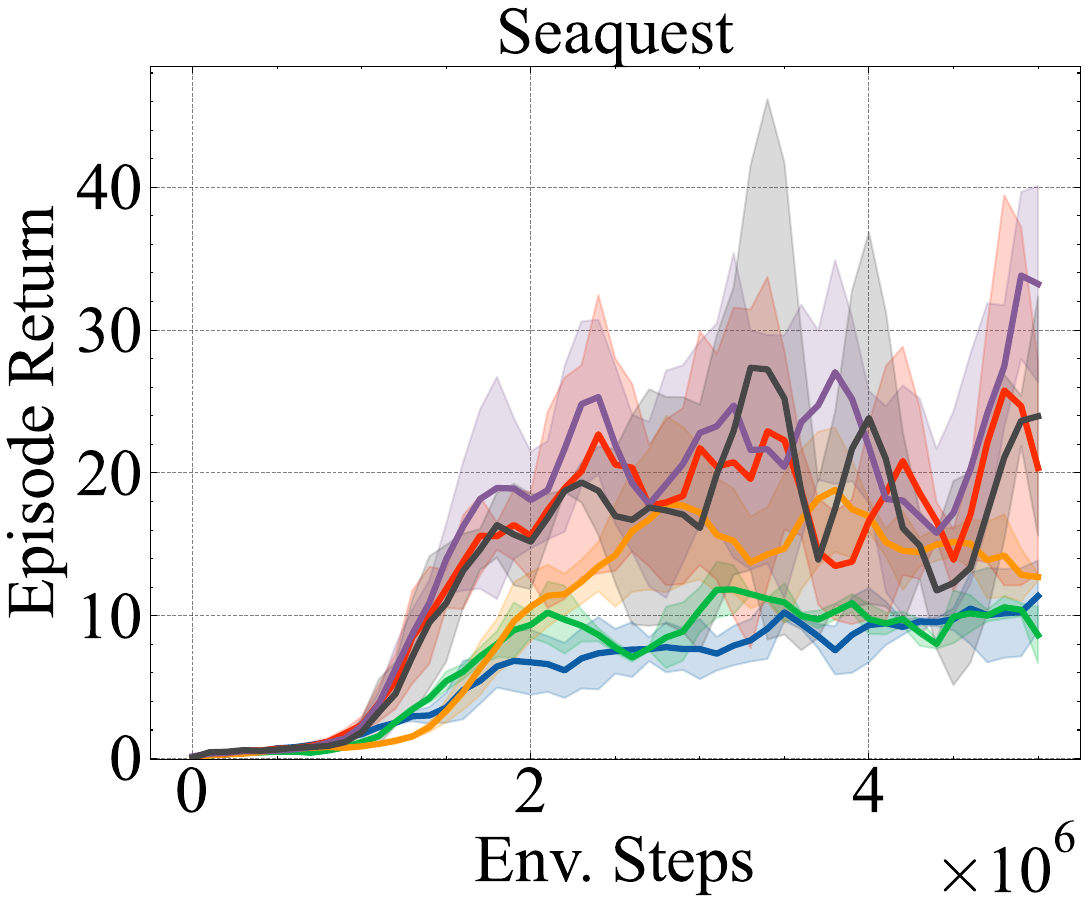}}
    \subfigure{\Description{}\includegraphics[width=0.24\textwidth]{pic/policy_set_size/SpaceInvaders_ablation_policy_set_size.pdf}}
    \vskip -0.1in
    \caption{
    The influence of the policy set size. We construct the policy set with different sizes. The performance improves with increasing policy size and obtain obvious improvement if we contain all the three basic functions in the policy set.
    }
    % \vskip -0.2in
    \label{fig:policy set size}
\end{center}
\end{figure*}

% \noindent\textbf{Size of policy set.} 
One benefit of our method is that we can construct policy sets with different sizes without increasing computational overhead.
By adding different $\delta$ and $\alpha$, we get larger policy set.
We construct different sizes of policy sets as shown in~\cref{fig:policy set size}.

The $\pi_{\textit{cov}(0.05)},\pi_{\textit{cor}(0)},\pi_{\textit{cor}(1)}$ in the figure means there is only one policy.
And others show the size of the policy set that combining all three basic functions like~\cref{eq:policy set}.
\textbf{Size 8} denotes the policy set $\Pi = \{\pi_{\textit{cov}(0.05)},\pi_{\textit{cov}(0.1)}, \pi_{\textit{cor}(0)},\pi_{\textit{cor}(0.2)},\pi_{\textit{cor}(0.4)},\cdots,\pi_{\textit{cor}(1)} \}$.
\textbf{Size 13} denotes the policy set $\Pi = \{\pi_{\textit{cov}(0.05)},$ $\pi_{\textit{cov}(0.1)},$ $\pi_{\textit{cor}(0)},$ $\pi_{\textit{cor}(0.1)},$ $\pi_{\textit{cor}(0.2)},$ $\cdots,\pi_{\textit{cor}(1)} \}$, which we used in our main results.
\textbf{Size 23} denote the policy set $\Pi = \{\pi_{\textit{cov}(0.05)},$ $\pi_{\textit{cov}(0.1)},$ $\pi_{\textit{cor}(0)},$ $\pi_{\textit{cor}(0.05)},$ $\pi_{\textit{cor}(0.1)},$ $\cdots,\pi_{\textit{cor}(1)} \}$.

% $\alpha$ interval 0.2,0.1,0.05,  and threshold: (0.2, (0.05)), (0.1,(0.05,0.1)). 
We can find $\pi_{\textit{cov}(0.05)}$ does not learn anything in most of environments, which indicates only focusing on space coverage does no benefit the learning. 
This may because novel states may not correlate with improved rewards~\citep{bellemare2016unifying,simmons2021reward}.  
Though $\pi_{\textit{cor}(0)}$ and $\pi_{\textit{cor}(1)}$ both learns something, they perform poorer than a big policy set, which indicates a single policy itself is not enough to get good performance due to the lack of diverse exploration.
In contrast, when we combine the three of the basic functions, we get obvious performance gain with larger set sizes, which emphasize the importance of diverse exploration. 
And when we increase the policy size to 13 and 23, there is no big difference within 5 million environmental steps, which may indicate the diversity is similar in this two policy sets.

\begin{figure*}[htbp]
    % \vskip -0.1in
\begin{center}
    \subfigure{\Description{}\includegraphics[width=0.24\textwidth]{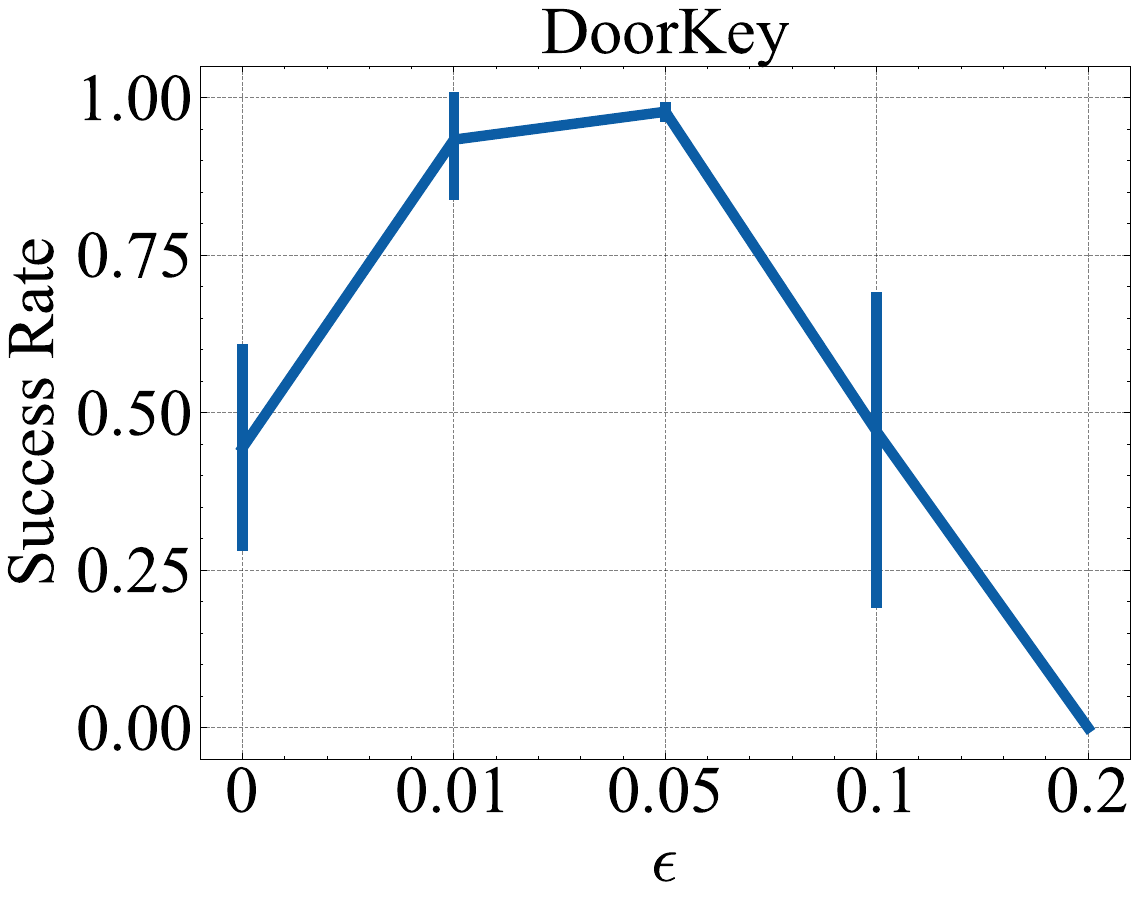}}
    \subfigure{\Description{}\includegraphics[width=0.24\textwidth]{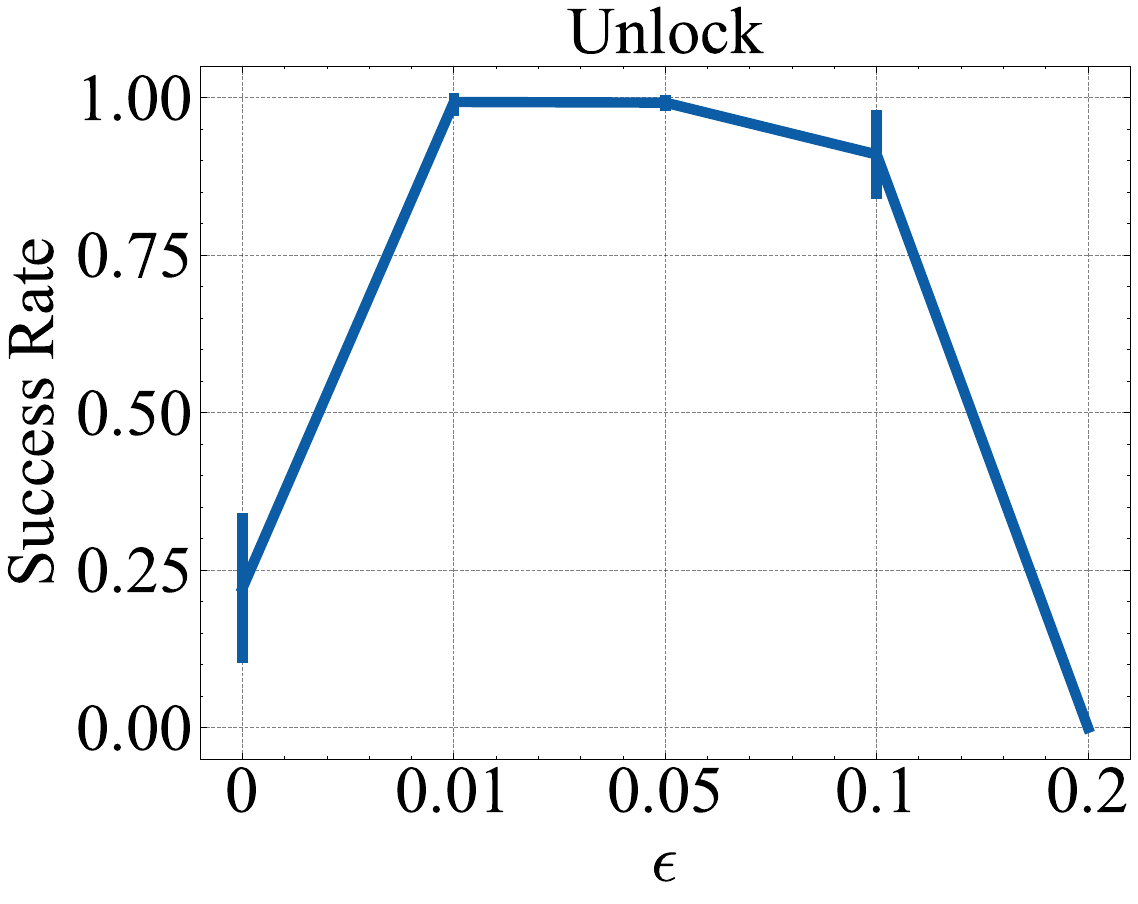}}
    \subfigure{\Description{}\includegraphics[width=0.24\textwidth]{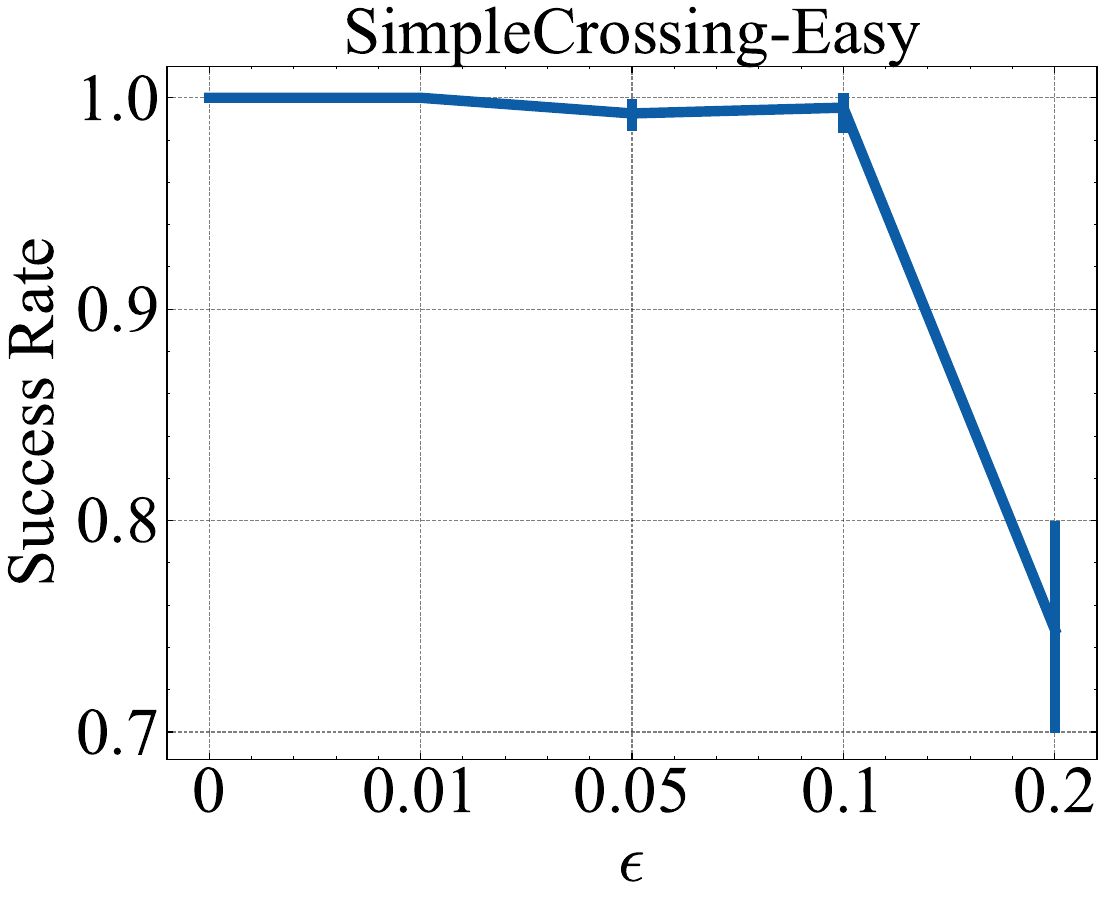}}
    \subfigure{\Description{}\includegraphics[width=0.24\textwidth]{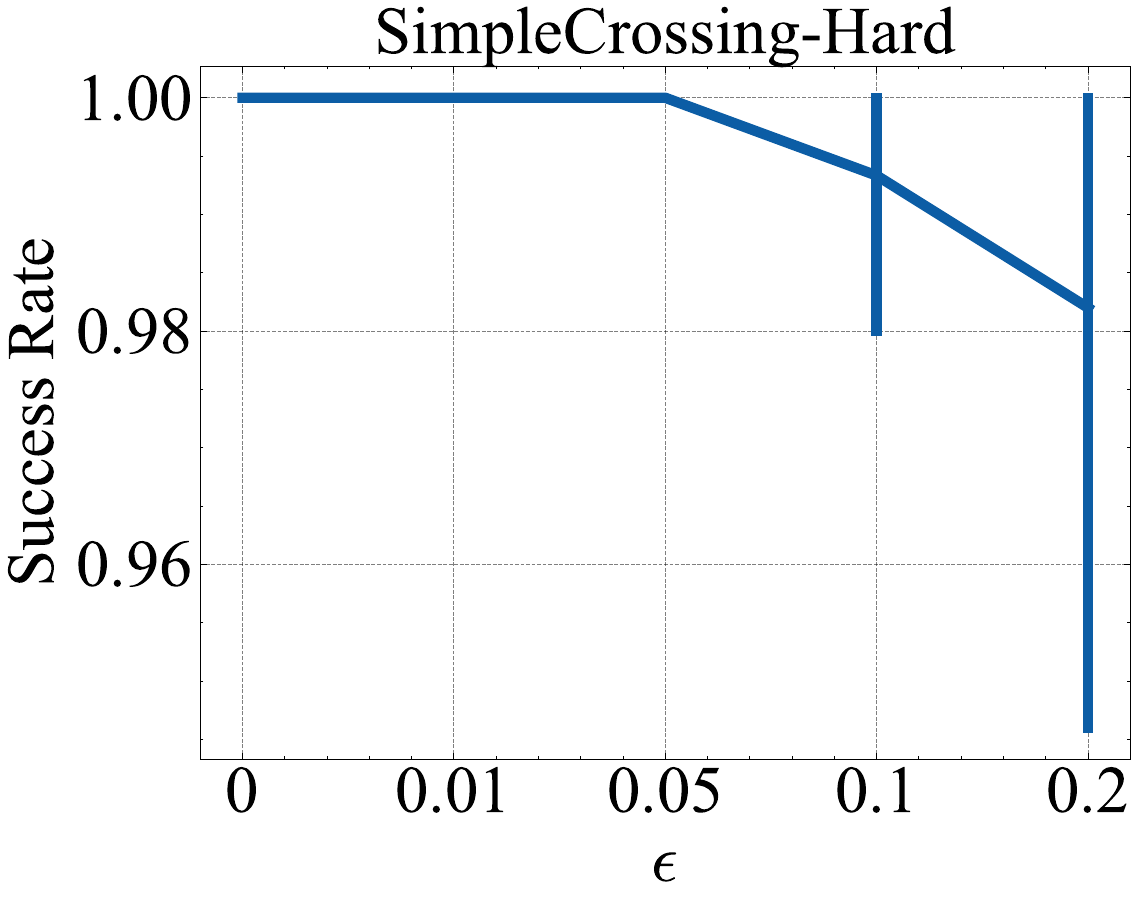}}
    \subfigure{\Description{}\includegraphics[width=0.24\textwidth]{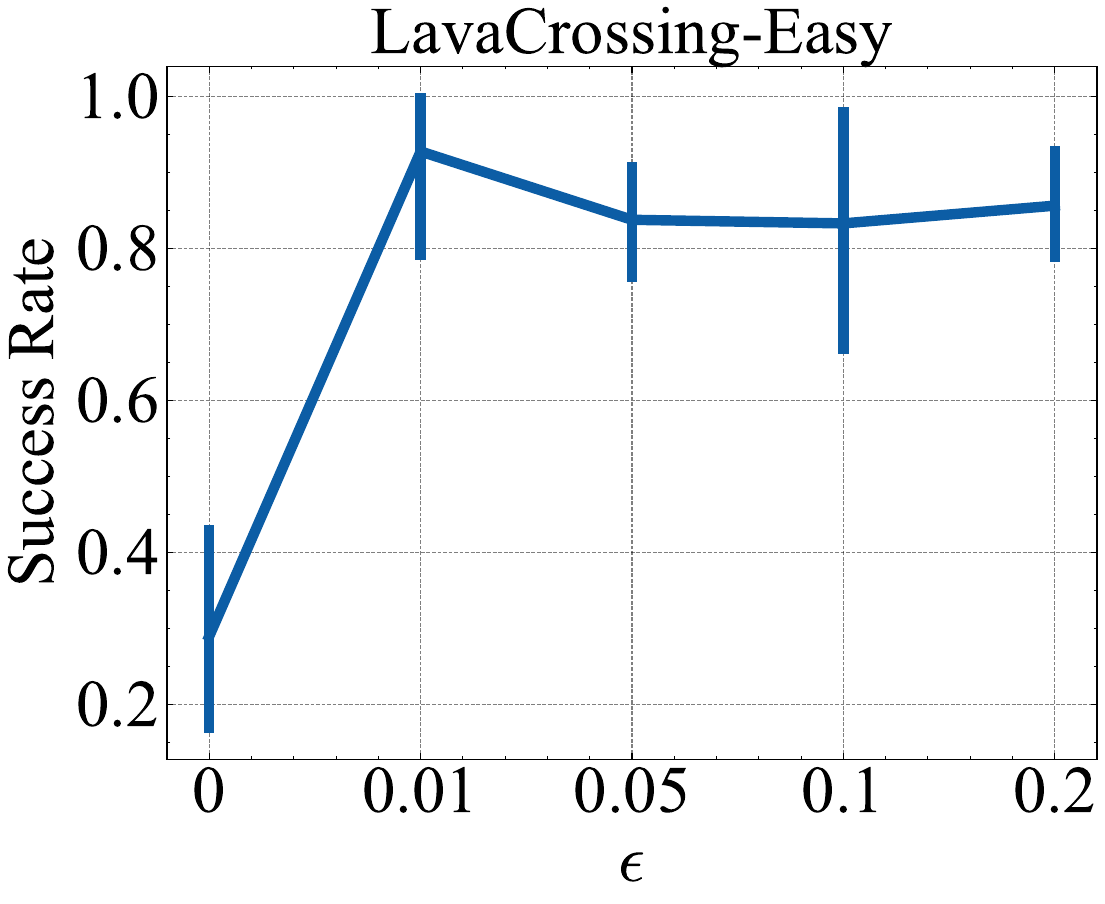}}
    \subfigure{\Description{}\includegraphics[width=0.24\textwidth]{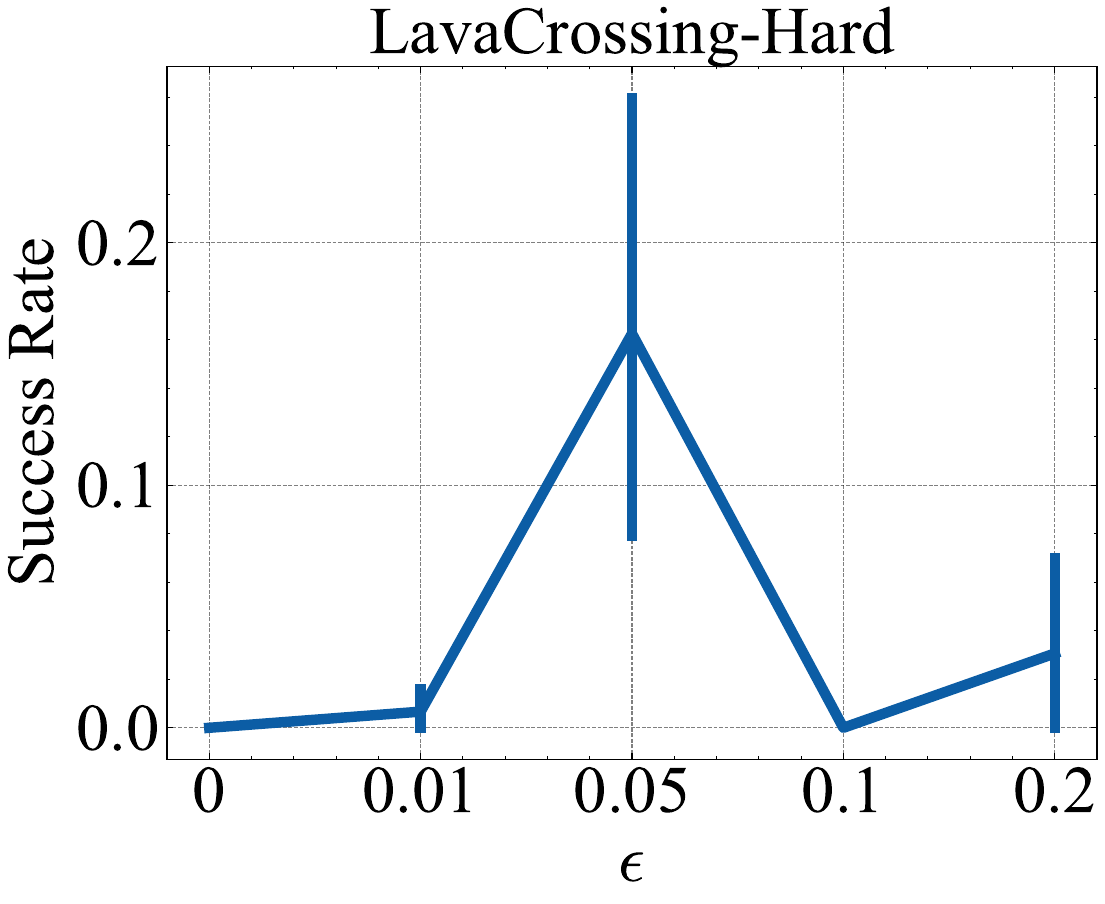}}
    \subfigure{\Description{}\includegraphics[width=0.24\textwidth]{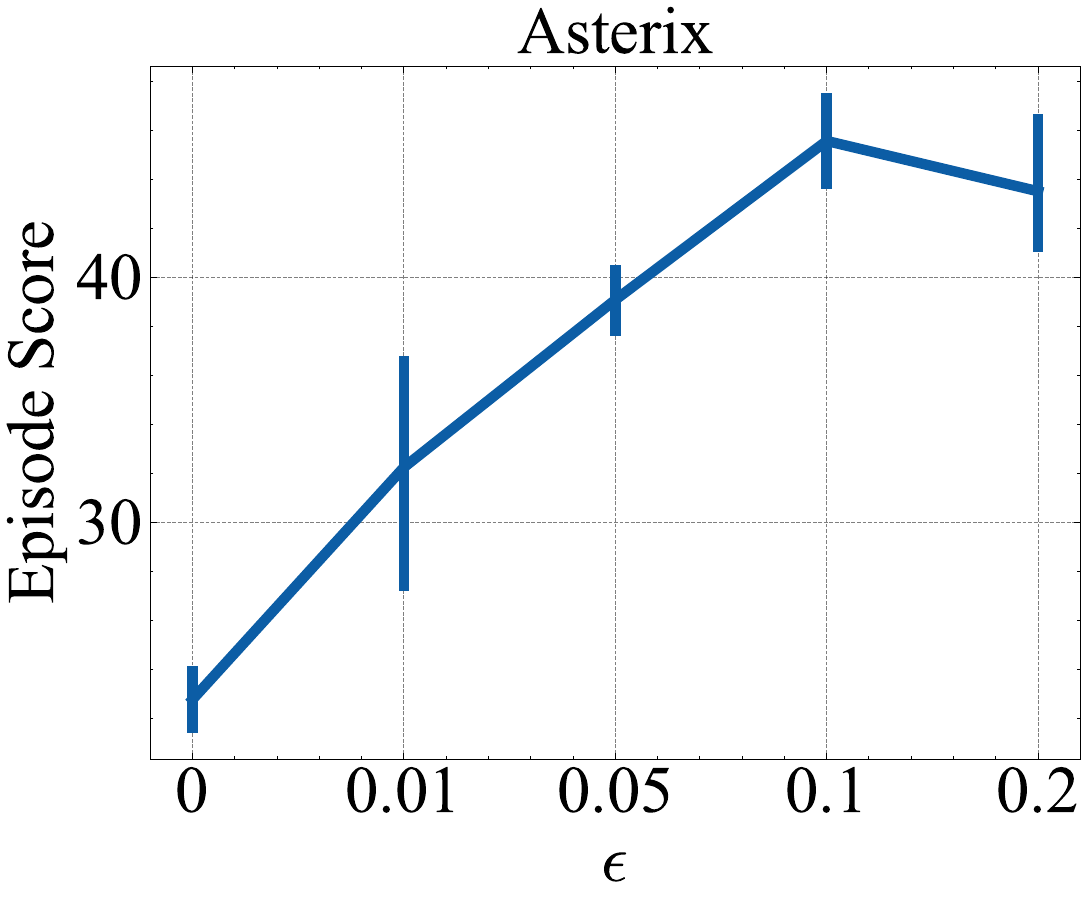}}
    \subfigure{\Description{}\includegraphics[width=0.24\textwidth]{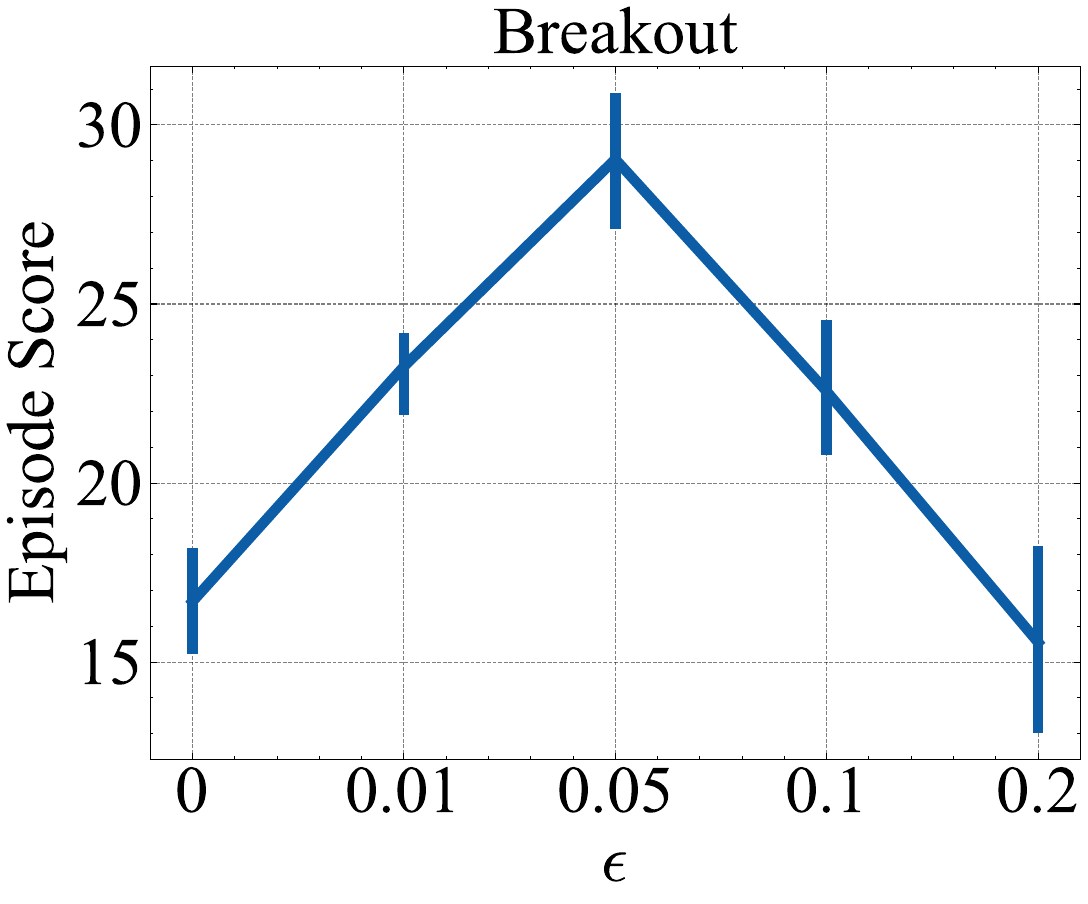}}
    \subfigure{\Description{}\includegraphics[width=0.24\textwidth]{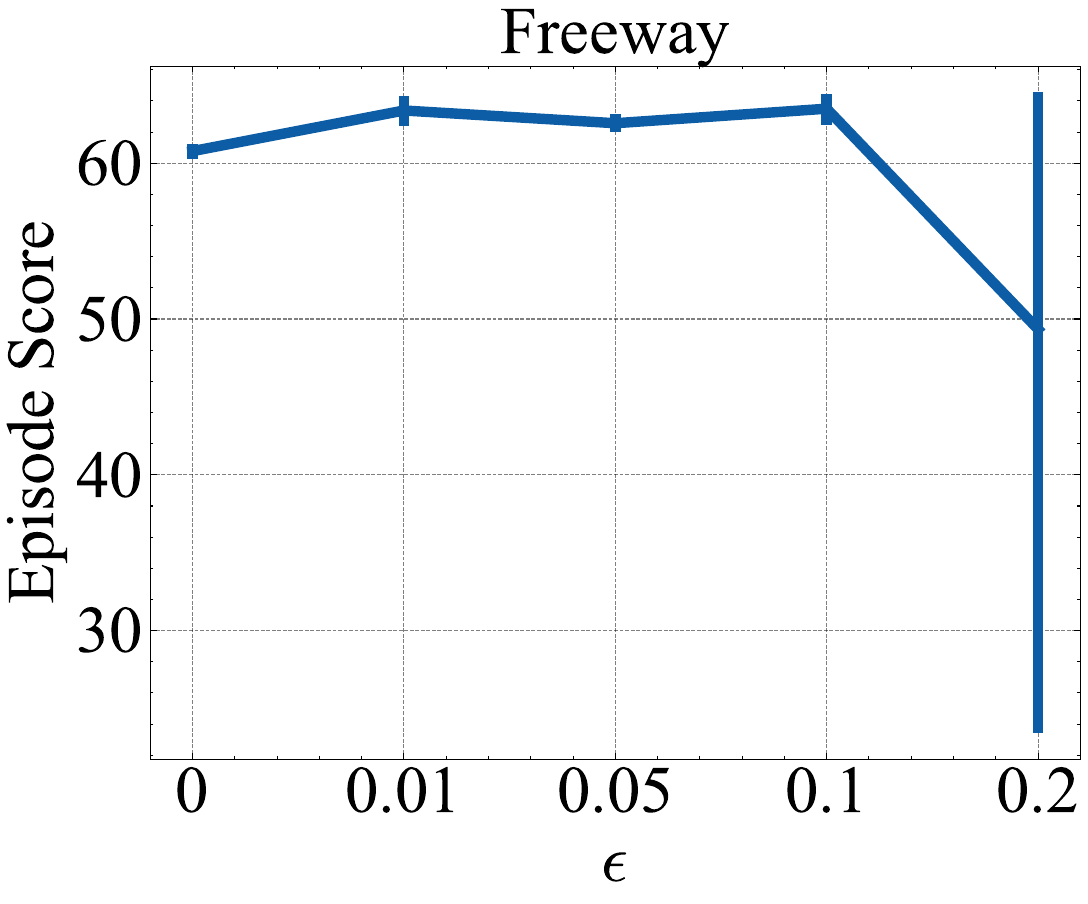}}
    \subfigure{\Description{}\includegraphics[width=0.24\textwidth]{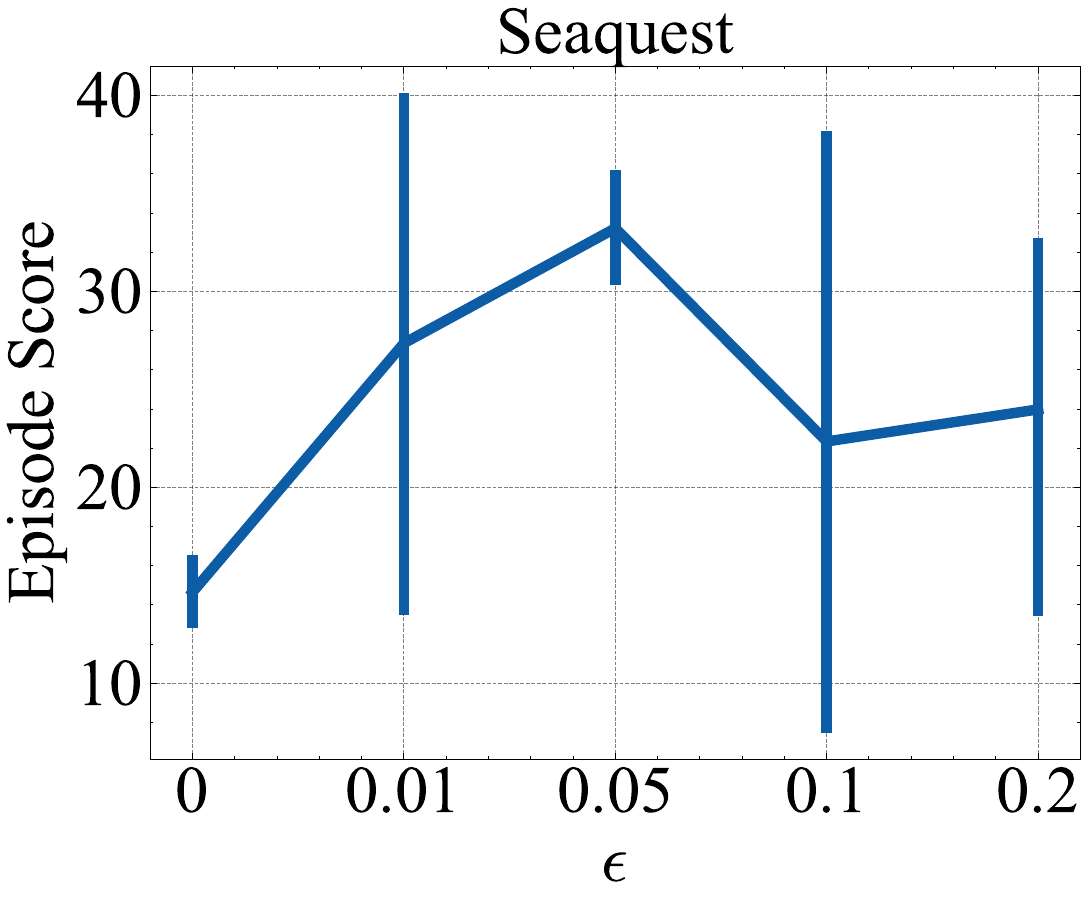}}
    \subfigure{\Description{}\includegraphics[width=0.24\textwidth]{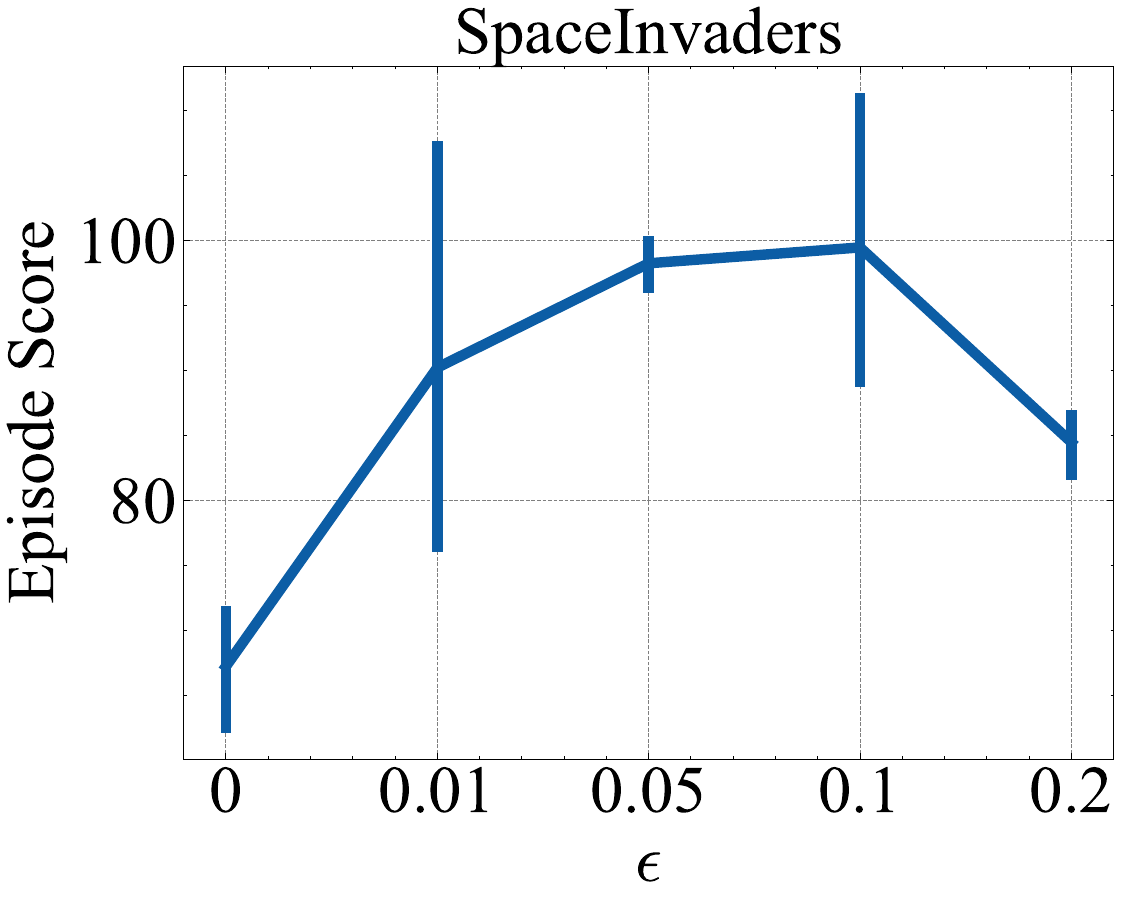}}
    \vskip -0.1in
    \caption{
    Sensitivity of parameter $\epsilon$.
    }
    % \vskip -0.2in
    \label{fig:parameter search epsilon}
\end{center}
\end{figure*}

\end{document}